\documentclass[11pt]{article}
\pdfoutput=1
\usepackage{etoolbox,verbatim}
\usepackage{subcaption}
\newtoggle{arxiv}
\toggletrue{arxiv}
\newtoggle{colt}
\togglefalse{colt}
\newtoggle{iclr}
\togglefalse{iclr}
\newtoggle{customthms}
\toggletrue{customthms}
\iftoggle{iclr}{
\usepackage{iclr2026_conference,times}
}{}

\usepackage{xcolor}
\usepackage{tcolorbox}
\tcbuselibrary{skins,breakable}

\tcbset{
  aibox/.style={
      width=\linewidth,
      top=6pt,
      bottom=0pt,
      left=1.5pt,
      right=1.5pt,
colback=blue!60!gray!5,
colframe=black,
colbacktitle=black,
      enhanced,
      center,
      attach boxed title to top left={yshift=-0.1in,xshift=0.15in},
      boxed title style={boxrule=0pt,colframe=white,},
    }
}
\newtcolorbox{AIbox}[2][]{aibox,title=#2,#1}

\definecolor{primalcolor}{HTML}{A60000}
\definecolor{contrarycolor}{HTML}{00A6A6}
\definecolor{darkcontrarycolor}{HTML}{004C4C}
\definecolor{lightblue}{HTML}{2970CC}
\definecolor{lightpurple}{HTML}{673147}
\definecolor{ForestGreen}{HTML}{FF5733}
\definecolor{myred}{HTML}{AA4A44}
\definecolor{hyppurple}{HTML}{800080}

\newcommand{\linkcolor}{darkcontrarycolor}
\newcommand{\urlcolor}{darkcontrarycolor}
\newcommand{\citecolor}{darkcontrarycolor}

\newcommand{\thmcolordark}{red!30!black}

\definecolor{root}{RGB}{1,115,178}
\definecolor{bimanual}{RGB}{198,102,39}
\definecolor{roothand}{RGB}{72,158,119}
\definecolor{endeffector}{RGB}{193,127,166}
\definecolor{rootendeffector}{RGB}{112,179,228}
\definecolor{upperbody}{RGB}{198,103,39}
\definecolor{rootupperbody}{RGB}{238,228,98}
\definecolor{wholebody}{RGB}{2,2,1}
\usepackage[T1]{fontenc}
\usepackage{capt-of}
\usepackage[font=small,labelfont=bf]{caption}

\usepackage{natbib}
\usepackage{amssymb,upgreek,bm}
\usepackage{mathtools}
\usepackage[english]{babel}  
\usepackage{multirow,tcolorbox,tikz-cd,turnstile,xspace,xparse}
\usepackage{relsize,breakcites,appendix}
\usepackage{longtable,tablefootnote,booktabs,float,makecell}
\usepackage[normalem]{ulem}
\usepackage{tabu}
\usepackage[shortlabels]{enumitem} 
\usepackage{footnote}
\usepackage{boxedminipage}
\usepackage{multirow,nicefrac}
\usepackage{verbatim} % gives the comment environment
\usepackage{wrapfig}

\usepackage[colorlinks=true,linkcolor=\linkcolor,urlcolor=\urlcolor,citecolor=\citecolor,breaklinks]{hyperref}

\usepackage{prettyref}

\usepackage[capitalise,nameinlink]{cleveref}

\iftoggle{colt}{
    \DeclareRobustCommand{\qed}{
        \usepackage{thmtools}
          \ifmmode \mathqed
          \else
            \leavevmode\unskip\penalty9999 \hbox{}\nobreak\hfill
            \quad\hbox{\qedsymbol}%
          \fi
    }
}
{
    \usepackage{amsthm}
     
}
\usepackage{thm-restate}

\iftoggle{arxiv}
{
    \usepackage{mathrsfs}
    \usepackage[margin=2cm]{geometry}
    \usepackage[bitstream-charter]{mathdesign}
     
    \usepackage[scaled=0.92]{PTSans}

    \usepackage{subfig}

    \usepackage{fullpage}
    \usepackage[noend]{algpseudocode}
    \usepackage{algorithm}
   
}
{
    \usepackage{mathrsfs}
}

\DeclareMathAlphabet{\mathbfsf}{\encodingdefault}{\sfdefault}{bx}{n}

\numberwithin{equation}{section}

\iftoggle{arxiv}
{
    }
{
    
}
\iftoggle{arxiv}
{
    
}
{
    
}

\usepackage{thmtools}

\Crefname{equation}{Eq.}{Eqs.}
\Crefname{assumption}{Assumption}{Assumptions}
\Crefname{condition}{Condition}{Conditions}
\Crefname{claim}{Claim}{Claims}
\Crefname{property}{Property}{Properties}
\Crefname{construction}{Construction}{Constructions}

\declaretheoremstyle[
    headformat=\normalfont\textcolor{\thmcolordark}{\bfseries\NAME\,\NUMBER}\NOTE,%
    notefont={\normalfont\textcolor{\thmcolordark}{\bfseries}}, 
    notebraces={}{},
    bodyfont=\normalfont\itshape,
    spaceabove = 6pt,
    spacebelow = 6pt,
    ]{coloredthmversion}

\declaretheoremstyle[
    headformat=\normalfont\textcolor{\thmcolordark}{\bfseries\NAME\,\NUMBER}\NOTE,%
    bodyfont=\normalfont\itshape,
    spaceabove = 6pt,
    spacebelow = 6pt,
    ]{coloredthm}

\declaretheoremstyle[
    headformat=\normalfont\textcolor{\thmcolordark}{\bfseries\NAME\,\NUMBER}\NOTE,%
    bodyfont=\normalfont,
    spaceabove = 6pt,
    spacebelow = 6pt,
    ]{coloreddef}

\iftoggle{customthms}
{
    \theoremstyle{coloredthmversion}
}
{}

\iftoggle{customthms}{
  \theoremstyle{coloredthm}
  \newtheorem{theorem}{Theorem}
  \newtheorem{lemma}{Lemma}[section]
  \newtheorem{corollary}{Corollary}[section]
  \newtheorem{proposition}[lemma]{Proposition}
}
{}

\newtheorem*{thminformal*}{Informal Theorem}

\iftoggle{customthms}{
    \theoremstyle{coloreddef}
    \newtheorem{definition}{Definition}[section]

    \newtheorem{property}{Property}[section]
    
}
{
  
}

\newtheorem{assumption}{Assumption}[section]
\newtheorem{condition}{Condition}[section]

\makeatletter
\newcommand{\neutralize}[1]{\expandafter\let\csname c@#1\endcsname\count@}
\makeatother

\iftoggle{customthms}{
    \newtheoremstyle{named}{}{}{\itshape}{}{\bfseries}{}{.5em}{\Cref{#3} {\normalfont (informal)} }{}
    \theoremstyle{named}
    
    \theoremstyle{plain}
}
{}

\newtheorem*{theorem*}{Theorem}
\newtheorem*{lemma*}{Lemma}
\newtheorem*{corollary*}{Corollary}
\newtheorem*{proposition*}{Proposition}
\newtheorem*{claim*}{Claim}
\newtheorem*{fact*}{Fact}
\newtheorem*{observation*}{Observation}
\newtheorem*{definition*}{Definition}
\newtheorem*{remark*}{Remark}
\newtheorem*{example*}{Example}

\def\ddefloop#1{\ifx\ddefloop#1\else\ddef{#1}\expandafter\ddefloop\fi}
\def\ddef#1{\expandafter\def\csname bb#1\endcsname{\ensuremath{\mathbb{#1}}}}
\ddefloop ABCDEFGHIJKLMNOPQRSTUVWXYZ\ddefloop

\def\ddefloop#1{\ifx\ddefloop#1\else\ddef{#1}\expandafter\ddefloop\fi}
\def\ddef#1{\expandafter\def\csname frak#1\endcsname{\ensuremath{\mathfrak{#1}}}}
\ddefloop ABCDEFGHIJKLMNOPQRSTUVWXYZ\ddefloop

\def\ddefloop#1{\ifx\ddefloop#1\else\ddef{#1}\expandafter\ddefloop\fi}
\def\ddef#1{\expandafter\def\csname fr#1\endcsname{\ensuremath{\mathfrak{#1}}}}
\ddefloop ABCDEFGHIJKLMNOPQRSTUVWXYZ\ddefloop

\def\ddefloop#1{\ifx\ddefloop#1\else\ddef{#1}\expandafter\ddefloop\fi}
\def\ddef#1{\expandafter\def\csname eul#1\endcsname{\ensuremath{\EuScript{#1}}}}
\ddefloop ABCDEFGHIJKLMNOPQRSTUVWXYZ\ddefloop

\def\ddefloop#1{\ifx\ddefloop#1\else\ddef{#1}\expandafter\ddefloop\fi}
\def\ddef#1{\expandafter\def\csname scr#1\endcsname{\ensuremath{\mathscr{#1}}}}
\ddefloop ABCDEFGHIJKLMNOPQRSTUVWXYZ\ddefloop

\def\ddefloop#1{\ifx\ddefloop#1\else\ddef{#1}\expandafter\ddefloop\fi}
\def\ddef#1{\expandafter\def\csname b#1\endcsname{\ensuremath{\mathbf{#1}}}}
\ddefloop ABCDEFGHIJKLMNOPQRSTUVWXYZ\ddefloop

\def\ddefloop#1{\ifx\ddefloop#1\else\ddef{#1}\expandafter\ddefloop\fi}
\def\ddef#1{\expandafter\def\csname bhat#1\endcsname{\ensuremath{\hat{\mathbf{#1}}}}}
\ddefloop ABCDEFGHIJKLMNOPQRSTUVWXYZ\ddefloop

\def\ddefloop#1{\ifx\ddefloop#1\else\ddef{#1}\expandafter\ddefloop\fi}
\def\ddef#1{\expandafter\def\csname btil#1\endcsname{\ensuremath{\tilde{\mathbf{#1}}}}}
\ddefloop ABCDEFGHIJKLMNOPQRSTUVWXYZ\ddefloop

\def\ddefloop#1{\ifx\ddefloop#1\else\ddef{#1}\expandafter\ddefloop\fi}
\def\ddef#1{\expandafter\def\csname bst#1\endcsname{\ensuremath{\mathbf{#1}^\star}}}
\ddefloop ABCDEFGHIJKLMNOPQRSTUVWXYZ\ddefloop

\def\ddefloop#1{\ifx\ddefloop#1\else\ddef{#1}\expandafter\ddefloop\fi}
\def\ddef#1{\expandafter\def\csname bst#1\endcsname{\ensuremath{\mathbf{#1}^\star}}}
\ddefloop abcdeghijklmnopqrstuvwxyz\ddefloop

\def\ddefloop#1{\ifx\ddefloop#1\else\ddef{#1}\expandafter\ddefloop\fi}
\def\ddef#1{\expandafter\def\csname bhat#1\endcsname{\ensuremath{\hat{\mathbf{#1}}}}}
\ddefloop abcdefghijklmnopqrstuvwxyz\ddefloop

\def\ddefloop#1{\ifx\ddefloop#1\else\ddef{#1}\expandafter\ddefloop\fi}
\def\ddef#1{\expandafter\def\csname b#1\endcsname{\ensuremath{\mathbf{#1}}}}
\ddefloop abcdeghijklnopqrstuvwxyz\ddefloop

\def\ddefloop#1{\ifx\ddefloop#1\else\ddef{#1}\expandafter\ddefloop\fi}
\def\ddef#1{\expandafter\def\csname barb#1\endcsname{\ensuremath{\bar{\mathbf{#1}}}}}
\ddefloop abcdefghijklmnopqrstuvwxyz\ddefloop

\def\ddef#1{\expandafter\def\csname c#1\endcsname{\ensuremath{\mathcal{#1}}}}
\ddefloop ABCDEFGHIJKLMNOPQRSTUVWXYZ\ddefloop
\def\ddef#1{\expandafter\def\csname h#1\endcsname{\ensuremath{\widehat{#1}}}}
\ddefloop ABCDEFGHIJKLMNOPQRSTUVWXYZ\ddefloop
\def\ddef#1{\expandafter\def\csname hc#1\endcsname{\ensuremath{\widehat{\mathcal{#1}}}}}
\ddefloop ABCDEFGHIJKLMNOPQRSTUVWXYZ\ddefloop
\def\ddef#1{\expandafter\def\csname t#1\endcsname{\ensuremath{\widetilde{#1}}}}
\ddefloop ABCDEFGHIJKLMNOPQRSTUVWXYZ\ddefloop
\def\ddef#1{\expandafter\def\csname tc#1\endcsname{\ensuremath{\widetilde{\mathcal{#1}}}}}
\ddefloop ABCDEFGHIJKLMNOPQRSTUVWXYZ\ddefloop
\newcommand{\ballkr}[1][r]{\cB_{k}(r)}

\DeclareMathSymbol{\shortminus}{\mathbin}{AMSa}{"39}

\Crefname{component}{Component}{Components}
\Crefname{contribution}{Contribution}{Contributions}
\newcommand{\componentref}[1]{%
  \hyperref[#1]{C\ref*{#1}}%
}
\Crefname{claim}{Claim}{Claims}
\Crefname{property}{Property}{Properties}

\iftoggle{iclr}
{
  
}
{
  
}

\makeatletter
\newcommand\addtometadatalist[5][]{%
  \begingroup
  \if\relax#3\relax\def\sep{}\else\def\sep{#5}\fi
  \let\protect\@unexpandable@protect
  \xdef#3{\expandafter{#3}\sep #4[#1]{#2}}%
  \endgroup
}

\newcommand\metadatalist{}
\newcommand\metadataformat[2][]{{\small \textbf{#1:} #2}}
\newcommand\metadata[2][]{\addtometadatalist[#1]{#2}{\metadatalist}{\metadataformat}{\\}}
\makeatother

\newcommand{\paperwebsite}[1]{\metadata[Website]{\href{https://#1}{\nolinkurl{#1}}}}
\newcommand{\papercode}[1]{\metadata[Code]{\url{#1}}}
\newcommand{\paperdocs}[1]{\metadata[Documentation]{\url{#1}}}
\newcommand{\paperblog}[1]{\metadata[Blog]{\url{#1}}}
\newcommand{\ignore}[1]{}

\iftoggle{arxiv}
{
\input{preamble_new/arxiv_title}
}
{}

\usepackage{adjustbox}
\usetikzlibrary{calc}
\usepackage{abstract}
\makeatletter
\renewcommand{\maketitle}{
    \newpage
    \null
    \begingroup
    \raggedright
    {\LARGE \bfseries \@title \par}
    \vskip 1.5em
    {\large
    \lineskip .5em
    \begin{tabular}[t]{l}
    \@author
    \end{tabular}\par}
    \vskip 1em
    {\large \@date \par}
    \endgroup
    \par
    \vskip 1.5em
}
\makeatother
\usepackage{amsmath,amsfonts,bm}
\usepackage{bbm}

\def\eqref#1{equation~\ref{#1}}
\def\1{\bm{1}}

\DeclareMathAlphabet{\mathsfit}{\encodingdefault}{\sfdefault}{m}{sl}
\SetMathAlphabet{\mathsfit}{bold}{\encodingdefault}{\sfdefault}{bx}{n}

\makeatletter
\@ifundefined{argmax}{}{}
\makeatother
\makeatletter
\@ifundefined{argmin}{}{}
\makeatother

\title{Scaling Behavior Foundation Model for Humanoid Robots}

\author{\small
Weishuai Zeng \textsuperscript{${1,6,7,*}$}~ 
Kangning Yin \textsuperscript{${2,7,*}$}~
Xiaojie Niu \textsuperscript{${7,*}$}~
Shunlin Lu \textsuperscript{${7}$}~
Weixiang Zhong \textsuperscript{${7}$}\\
\small
Jiahe Chen \textsuperscript{${3,7}$}~
Feiyu Jia \textsuperscript{${7}$}~
Xiao Chen \textsuperscript{${7}$}~
Zirui Wang \textsuperscript{${3,7}$}~
Furui Xu \textsuperscript{${7}$}~
Ming Zhou \textsuperscript{${7}$}~
Kailin Li \textsuperscript{${7}$}\\
\small
Weinan Zhang \textsuperscript{${2,7}$}~
He Wang \textsuperscript{${4,6}$}~
Li Yi \textsuperscript{${5,6}$}~
Dahua Lin \textsuperscript{${1,7}$}~
Jiangmiao Pang \textsuperscript{${7}$}~
Jingbo Wang \textsuperscript{${7}$}\\
\small
\vspace{-.3em}
 \rule{.38\textwidth}{.7pt}
\\
\footnotesize
$^{1}$The Chinese University of Hong Kong ~~$^{2}$Shanghai Jiao Tong University ~~ $^{3}$Zhejiang University ~~ $^{4}$ Peking University\\
\footnotesize
$^{5}$ Tsinghua University ~~ $^{6}$ Galbot ~~ $^{7}$ Shanghai Artificial Intelligence Laboratory \\
\footnotesize
$^{*}$Equal Contribution\\
}

\date{\vspace{-0.5cm}}

\usepackage{adjustbox}
\newcommand{\logo}[1]{%
  \adjustbox{
    max width=2.5cm,
    max height=1cm,
    keepaspectratio,
    valign=m
  }{\includegraphics{#1}}%
}

\newcommand{\mediumlogo}[1]{%
  \adjustbox{
    max width=4cm,
    max height=2cm,
    keepaspectratio,
    valign=m
  }{\includegraphics{#1}}%
}

\newcommand{\largelogo}[1]{%
  \adjustbox{
    max width=6cm,
    max height=2.5cm,
    keepaspectratio,
    valign=m
  }{\includegraphics{#1}}%
}

\paperwebsite{scalebfm.github.io}
\usepackage{tikz}
\usepackage{eso-pic} % shipout hooks (place things at fixed page positions)
\usepackage[dvipsnames,table,svgnames]{xcolor}
\usepackage{tcolorbox}
\usepackage{wrapfig}
\makeatletter
\providecommand\sf@counterlist{}
\makeatother

\newcommand{\PlaceFirstPageLogos}{%
  \AddToShipoutPictureFG*{%
    \begin{tikzpicture}[remember picture,overlay]

      \node[anchor=north west,xshift=25mm,yshift=-7mm]
        at (current page.north west){%
        \mediumlogo{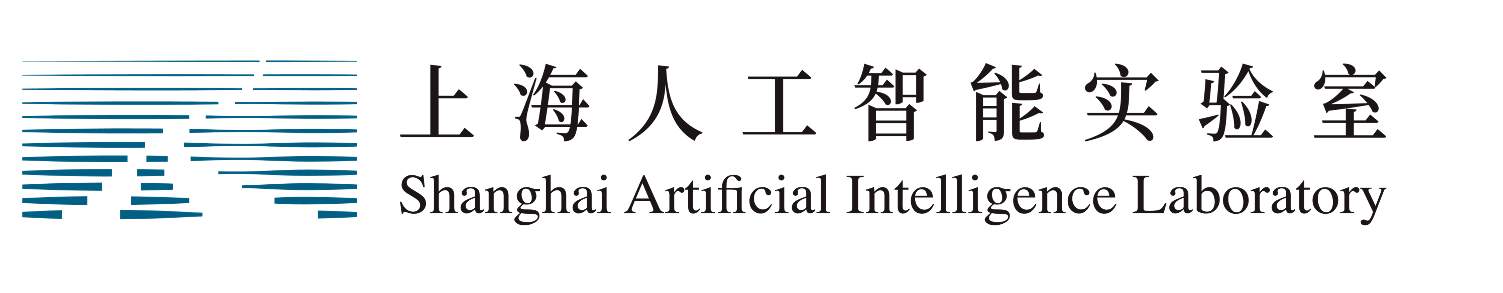}\hspace{0.4em}%
        \largelogo{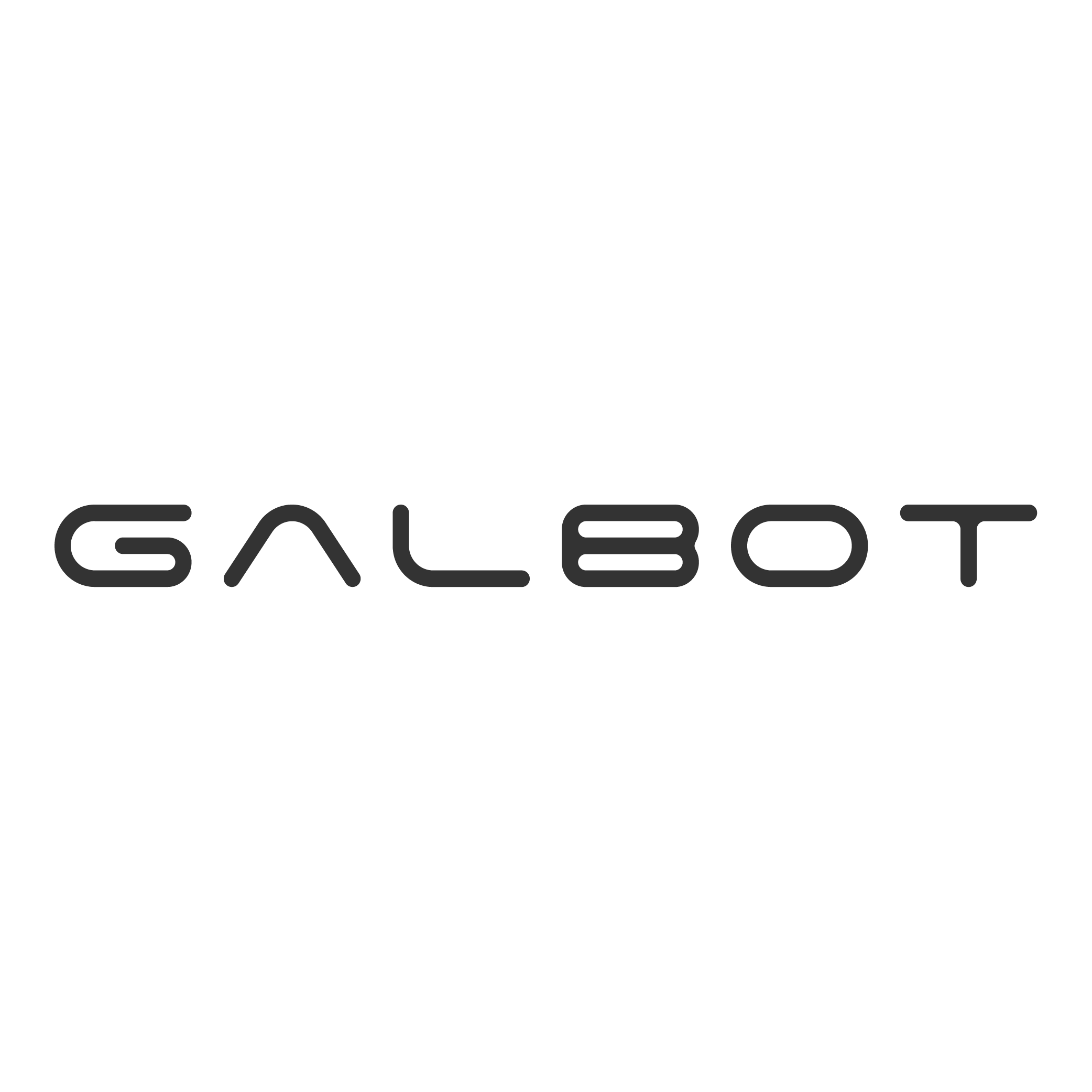}%
      };

      \node[anchor=north east,xshift=-25mm,yshift=-13mm]
        at (current page.north east){%
        \logo{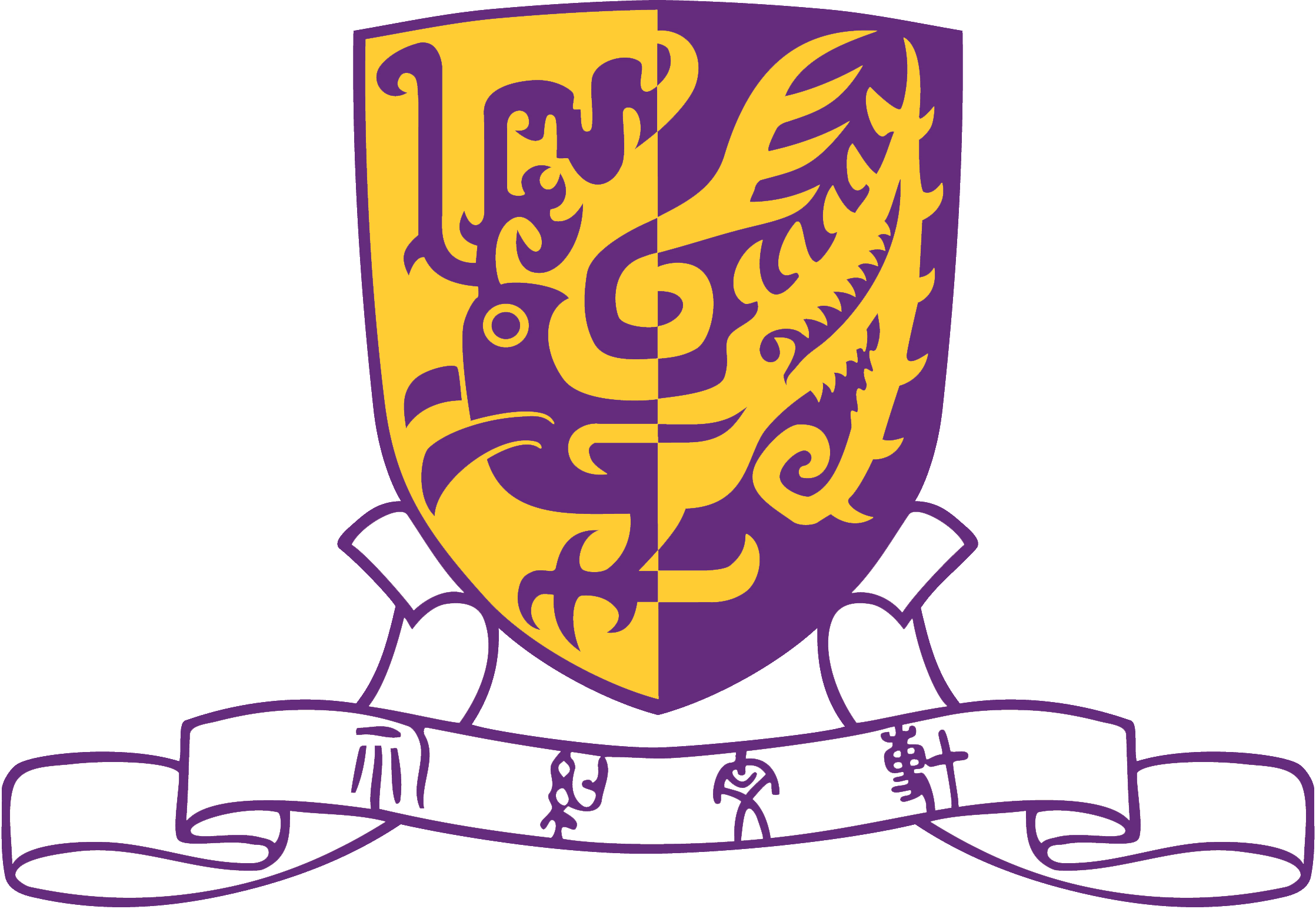}\hspace{0.01em}%
        \logo{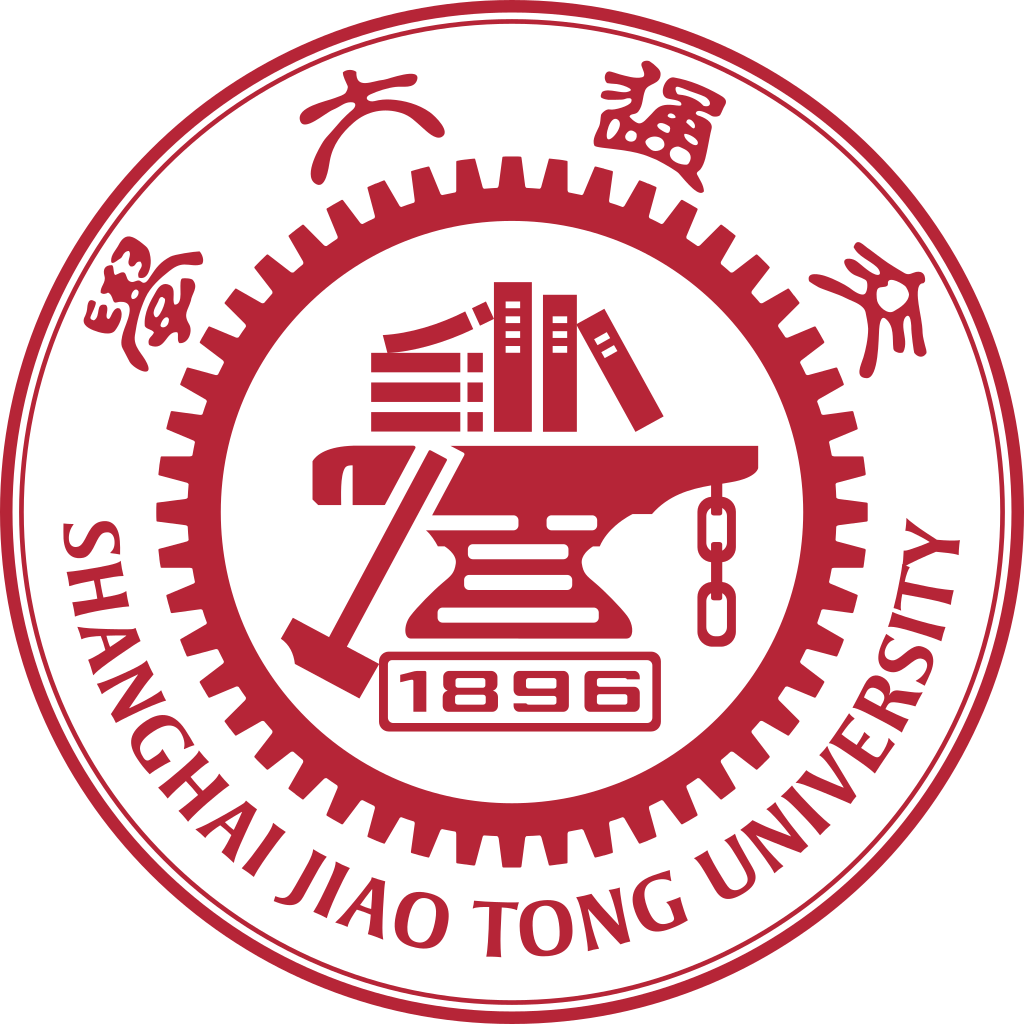}\hspace{0.35em}%
        \logo{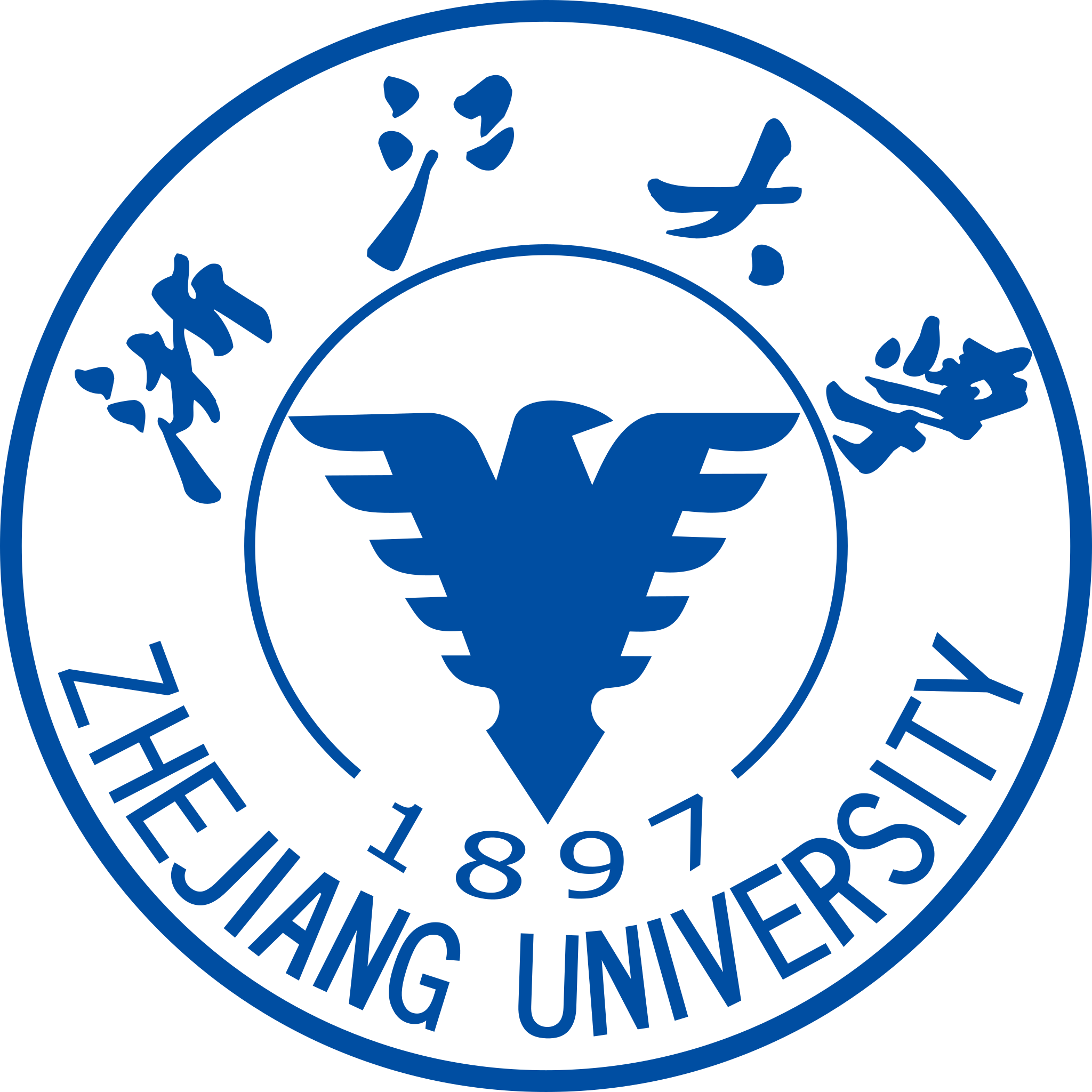}\hspace{0.35em}%
        \logo{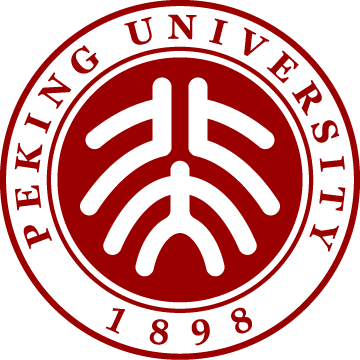}\hspace{0.35em}%
        \logo{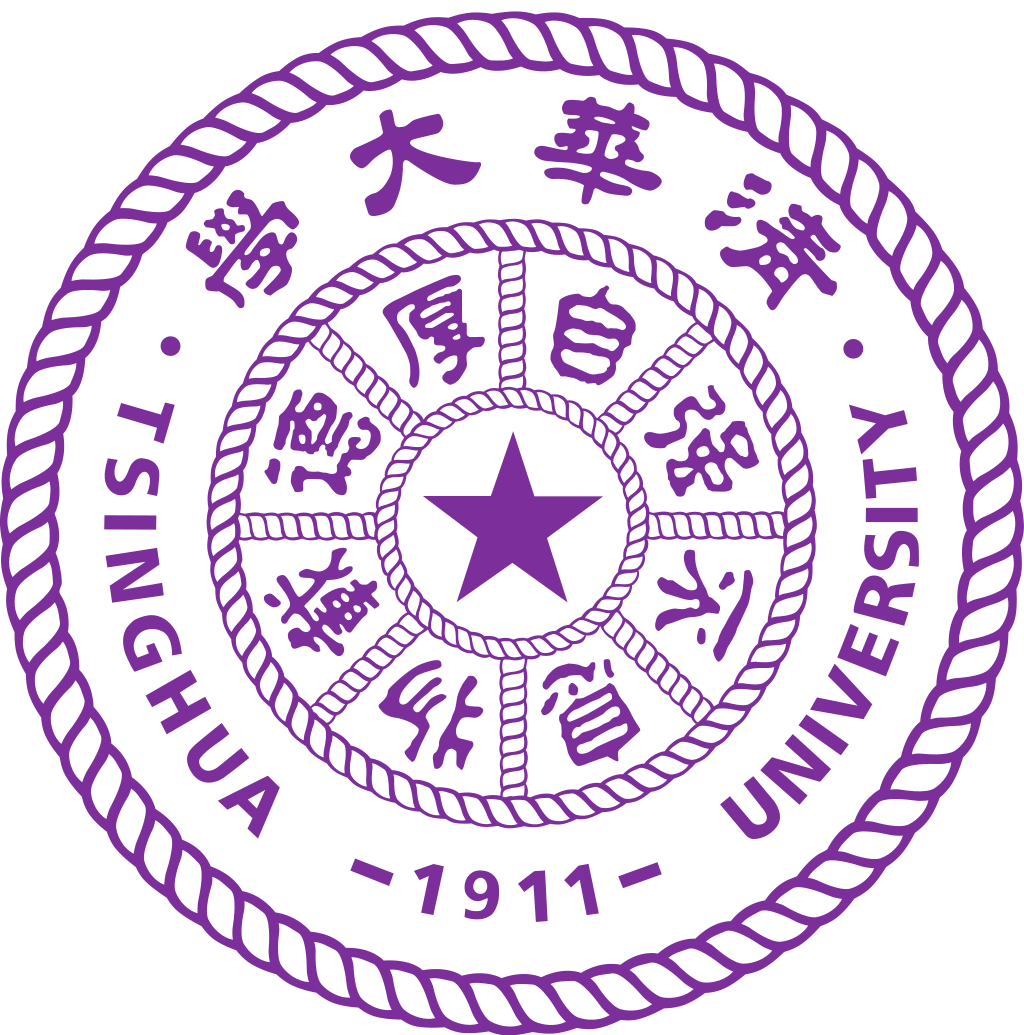}%
      };

    \end{tikzpicture}%
  }%
}

\begin{document}
\PlaceFirstPageLogos
\begin{tcolorbox}[    colback=cyan!4, colframe=cyan!25,
    boxrule=0pt,
    arc=2mm%title=\bfseries Document Information
  ]
  \maketitle
  \vspace{-1em}
  \tcbline

  \begin{minipage}[t]{1.0\linewidth}
    \centering
    \includegraphics[width=1.0\textwidth]{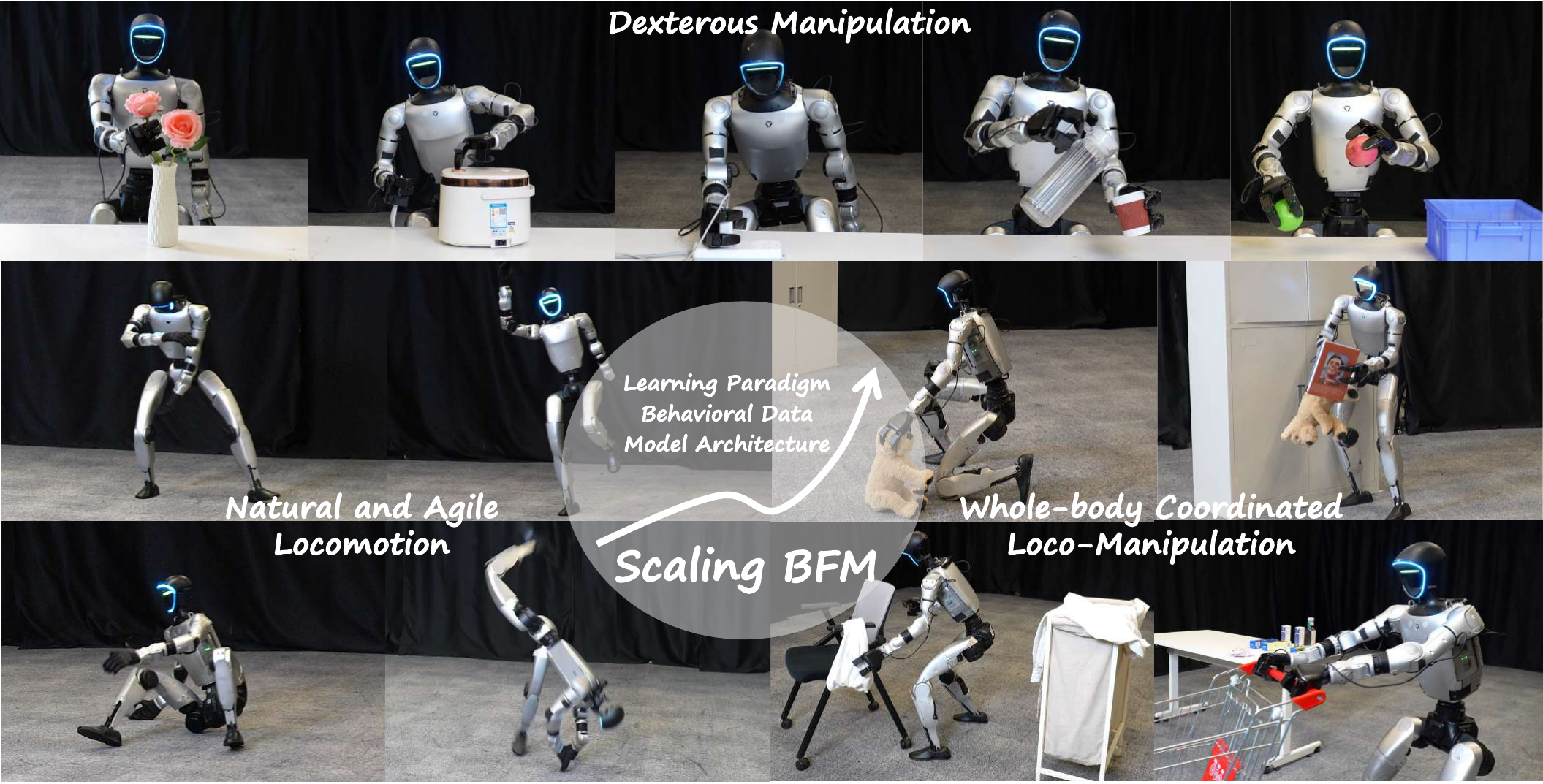}
    \captionof{figure}{ \small Our \textbf{Behavior Foundation Model} enables versatile humanoid behaviors across a broad spectrum of tasks, including dexterous manipulation, natural and agile locomotion and whole-body coordinated loco-manipulation, which highlights its potential as a foundation for general-purpose humanoid control.}
    \label{figure_1_teaser}
\end{minipage}
  \tcbline
  \vspace{0.3em}
  % Metadata section (after abstract, inside tcolorbox)
  % \vskip 0.5cm
  \makeatletter
  \ifdefempty{\metadatalist}{}{\metadatalist\par}
  % \begin{center}
  %   \logo{logos/cuhk.png}\hfill
  %   \logo{logos/sjtu.png}\hfill
  %   \logo{logos/zju.jpg}\hfill
  %   \logo{logos/pku.png}\hfill
  %   \logo{logos/thu.png}\hfill
  %   \largelogo{logos/galbot.png}\hfill
  %   \mediumlogo{logos/ailab.png}
  % \end{center} 
  \makeatother
  \vspace{0.3em}
\end{tcolorbox}

\renewcommand{\abstractnamefont}{\fontsize{13}{15}\selectfont\bfseries}

\vspace{3mm}
\begin{abstract}
\vspace{-3mm}
\noindent Humanoid control requires natural whole-body coordination, precise real-time responses to control signals, and robust generalization across diverse environmental contexts, making it a cornerstone for generalist embodied agents. 
Behavior Foundation Models (BFMs) have recently emerged as a promising solution to address these challenges by leveraging large-scale behavioral data to achieve superior expressiveness, versatility and generalization. 
However, despite growing interest in scaling BFMs to further improve their capabilities, it remains unclear how key factors, including the learning paradigm, behavioral data and model architecture should be coordinated to enable effective scaling. 
In this work, we revisit the scaling recipe for BFMs and demonstrate that substantial performance gains can be achieved through the coordination of three core components: 
1) the learning paradigm of motion tracking that reformulates diverse humanoid control problems as the reproduction of integrated whole-body behaviors in the global frame;
2) the strategic synergy between on-policy rollout quantity and reference motion diversity;
and 3) the expressive and scalable model architecture termed Humanoid Transformer that facilitates the natural emergence of structured behavioral representations. 
Through extensive experiments in both simulation and real-world deployment, we demonstrate that our approach yields significant improvements in control fidelity and task generalization, reducing Mean Per-Keypoint Position Error (MPKPE) on the test set by over 10\% in local mode and 82\% in global mode compared with existing humanoid controllers. These results establish BFM as a principled and effective foundation for scalable and general-purpose humanoid control.
\end{abstract}

\vspace{-3mm}
\section{Introduction}
\vspace{-2mm}

Humanoid robots are increasingly viewed as a promising general-purpose embodiment for operating in human-centric environments, owing to their rich actuation, human-like morphology and natural compatibility with daily infrastructures designed for humans. Realizing this potential, however, depends on effective whole-body control that requires the coordinated regulation of numerous degrees of freedom to achieve natural, responsive and robust behaviors across multiple tasks and environmental contexts. 

Recent years have witnessed substantial progress in humanoid whole-body control, enabling impressive capabilities across diverse motor skills~\citep{huang2025learning,he2025learning,xue2025unified,ben2025homie,he2025asap}. Nevertheless, these systems remain largely task-specific, often relying on extensive reward engineering tailored to individual tasks and contextual settings. This specialization limits their versatility and hinders generalization across diverse scenarios. 

To address these limitations, recent research has identified behavior as a shared objective across multiple whole-body control tasks~\cite{zeng2025behavior} and introduced \textbf{Behavior Foundation Models (BFMs)}, large-scale pretrained humanoid foundation controllers capable of robust and versatile humanoid control in response to diverse behavior specifications~\cite{tessler2024maskedmimic,he2025hover,liao2025beyondmimic,li2025bfm,luo2025sonic}. By decoupling underlying behaviors from task-specific control modes, BFMs represent a paradigm shift from isolated task learning toward holistic behavior learning. In this work, we focus on a special line of BFMs that incorporate an explicit control interface~\cite{zeng2025behavior,tessler2024maskedmimic,luo2025sonic}. This design not only enables direct responses to behavioral specifications at deployment, but also promotes the emergence of structured latent representations that organize explicit behavioral intentions within a coherent latent space. Such reusable behavioral representations facilitate strong generalization across diverse specifications and efficient adaptation to downstream applications.

Inspired by the success of foundation models in other domains~\cite{kaplan2020scaling,wei2022emergent,blattmann2023stable}, a natural question arises: \textit{Can behavior foundation models for humanoid robots be effectively scaled to further enhance their capabilities?} Recent studies have taken initial steps toward scaling BFMs~\cite{luo2025sonic}. However, these efforts remain fragmented, offering only preliminary conclusions while leaving concrete design principles for effective scaling largely unexplored. To bridge this gap, we revisit current BFM pretraining pipelines and develop a principled scaling recipe that coordinates multiple factors, including a unified and efficient learning paradigm, large-scale and diverse behavioral data, and an expressive yet scalable model architecture.

Our framework identifies motion tracking~\cite{peng2018deepmimic} as a unified and scalable paradigm which reformulates diverse humanoid control problems as the imitation of reference motions, thereby providing dense whole-body guidance for efficient behavior learning. While sharing a similar vision with prior work~\cite{liao2025beyondmimic,luo2025sonic,chen2026holomotion}, we adopt a distinct training recipe that requires the robot to reproduce reference motions as \emph{integrated whole-body trajectories in the global frame}. This design not only mitigates the behavioral ambiguity caused by ignoring root translation, under which behaviors such as walking forward and marching in place may become indistinguishable despite reflecting different behavioral intentions. It also avoids decoupling global root evolution from local pose tracking, allowing deviations in root motion to propagate to the entire body and hence providing stronger guidance for behavior learning. Notably, this formulation preserves flexibility in behavioral specification: the same BFM can support both global control with root localization and local control by re-anchoring the control signals to the robot's current root state.

Under this learning paradigm, we argue that scaling BFMs is fundamentally driven by the non-trivial interplay between the quantity and diversity of training data. We instantiate motion tracking with Proximal Policy Optimization (PPO)~\cite{schulman2017proximal}, given its mature ecosystem in humanoid control, while noting that alternative algorithms have also shown promising results~\cite{seo2025fasttd3,yi2026flow}. Existing studies often equate training data scaling with increasing the number of reference motions. However, under the PPO framework, effective training data are the on-policy rollouts collected through environment interaction. Therefore, data quantity is more directly governed by factors such as the number of parallel environments and the rollout horizon, whereas increasing the number of reference motions primarily shapes the behavioral distribution and enriches its diversity when properly curated. We thus posit that data scaling for BFM pretraining emerges from the synergy between quantity and diversity: quantity is obtained by scaling on-policy data collection, while diversity is achieved by expanding the reference motion corpus.

Finally, we emphasize that an expressive and scalable model architecture is essential for scaling BFMs. While existing works primarily adopt MLPs as the backbone, and recent studies have shown that scaling MLPs can improve performance~\cite{luo2025sonic}, we investigate whether more expressive architectures exhibit stronger scaling behavior. Beyond architectural design, we further examine whether structured latent representations of behavioral intentions can naturally emerge without relying on auxiliary objectives that may encode human priors. Such representations could facilitate the flexible utilization and efficient transfer of behavioral knowledge learned by BFMs, thereby supporting diverse downstream applications.

Overall, this paper presents a systematic investigation into the scaling behavior of BFMs by revisiting three key ingredients of the scaling recipe: the learning paradigm, behavioral data, and model architecture. Through comprehensive experiments in both simulation and real-world deployment, we validate our findings and demonstrate that our approach significantly improves control fidelity, execution robustness, and generalization capability, reducing Mean Per-Keypoint Position Error (MPKPE) on the test set by over 10\% in local mode and 82\% in global mode compared with existing humanoid controllers. These results establish our BFM as a principled foundation for scalable, general-purpose humanoid control.
\vspace{-3mm}
\section{Related Work}
\vspace{-1mm}
\label{sec:related-work}

\subsection{Humanoid Whole-body Control}
% \vspace{-1mm}

Humanoid control has recently undergone a paradigm shift from task-specific reward engineering toward whole-body reference imitation. Early approaches achieved impressive performance across diverse motor skills through carefully designed reward functions and regularization terms~\cite{huang2025learning,xue2025unified,ben2025homie}. However, their reliance on task-specific objectives often constrains generalization across heterogeneous control settings. In contrast, mimic-based approaches formulate humanoid control as reference motion tracking, providing a unified framework for learning natural, expressive, and coordinated whole-body behaviors~\cite{chen2025gmt,yin2026unitracker,wang2026omnixtreme}. This paradigm has enabled wide applications, including human teleoperation~\cite{li2025clone,ze2025twist,ze2025twist2}, humanoid-scene interaction~\cite{wang2025physhsi,yang2025omniretarget,zhuang2026deep}, and humanoid-object interaction~\cite{weng2025hdmi,wang2026humanx,he2026ultra}.

Despite these advances, conventional motion-tracking-based approaches remain constrained by their dependence on explicit whole-body reference motions as the primary control interface. To address this limitation, recent research has explored two complementary directions. The first extends humanoid control to multimodal commands, either preserving whole-body motions as an intermediate representation and then training high-level motion generators for translating multimodal inputs into trackable reference trajectories~\cite{jiang2025uniact,xie2026textop,tao2026heracles}, or developing more flexible control interfaces that can directly respond to diverse command forms without requiring explicit whole-body references during inference~\cite{zeng2025behavior,he2025hover}. The second direction focuses on learning reusable behavioral representations, where structured latent spaces are acquired through motion tracking to capture diverse motor skills and behavioral priors in a compact and transferable form~\cite{yin2026robostriker,zhang2026learning}. Such representations can subsequently be leveraged across downstream tasks, facilitating efficient skill acquisition and rapid adaptation to varying contexts.

\vspace{-2mm}
\subsection{Behavior Foundation Models for Humanoid Robots}
% \vspace{-1mm}

Behavior Foundation Models (BFMs) have emerged as a promising paradigm for general-purpose humanoid control, owing to their broad versatility and strong generalization. Recent studies have explored diverse pretraining strategies and control mechanisms for constructing such models. BFM4Humanoid~\cite{zeng2025behavior} pretrains BFMs within the DAgger framework~\cite{ross2011reduction}, where actions are labeled by an expert policy trained via motion tracking. It further introduces an explicit control interface, enabling the pretrained model to directly respond to diverse behavior specifications at deployment. Based on a similar distillation framework, BeyondMimic~\cite{liao2025beyondmimic} first learns structured behavioral representations as a proxy action space and then performs pretraining through supervised learning on offline behavioral trajectories. At inference time, it applies classifier guidance~\cite{dhariwal2021diffusion} to the diffusion-based BFM to generate task-conditioned behavior trajectories for humanoid control~\cite{huang2025diffuse}. BFM-Zero~\cite{li2025bfm} pretrains BFMs via unsupervised reinforcement learning based on forward-backward representations~\cite{touati2021learning,tirinzoni2025zero}. With the promptable latent space learned through pretraining, it enables zero-shot generalization and few-shot adaptation through sampling-based inference at test time. More recently, SONIC~\cite{luo2025sonic} explores the direct usage of motion tracking for BFM pretraining. By learning discrete behavioral representations and enforcing alignment across multiple command forms, it enables versatile control through the resulting shared latent space.

As BFMs continue to demonstrate strong performance across diverse humanoid control settings, scaling has naturally emerged as a key direction for further improving their capabilities. However, their concrete scaling behavior has yet to be systematically identified. While SONIC offers preliminary evidence by scaling the number of reference motions, model capacity, and GPU hours within the motion-tracking framework, it fails to elucidate the coupling among key factors that govern effective scaling, including the learning paradigm, training data, and model architecture. In this work, we revisit the design of motion tracking, clarify the notion of training data in PPO-based pretraining, and introduce a more expressive architecture termed the Humanoid Transformer. These explorations aim to offer a more principled understanding of BFM scaling and lay a stronger foundation for general humanoid control.
\section{Method}
\label{sec:method}
\vspace{-2mm}

In this section, we first present our revised formulation and learning objective for BFM pretraining and establish motion tracking as its principled and scalable instantiation. We then detail our motion-tracking recipe, including the state representation, control interface, reward design, and domain randomization strategy. Next, we formalize the notion of training data within our framework and elucidate the corresponding dataset construction pipeline to achieve a synergy between data quantity and diversity. Finally, we introduce a more expressive and scalable architecture termed the \emph{Humanoid Transformer}, which facilitates the natural emergence of structured behavioral representations. Collectively, these components constitute the methodological foundation of our scaling recipe for BFM pretraining.

\vspace{-3mm}
\subsection{Problem Formulation}
\label{sec:formulation}
\vspace{-2mm}
    
We formulate humanoid control as a Markov Decision Process (MDP) ${M}=\langle{S}, {A}, {T}, {R}, \gamma\rangle$, where ${S}$ denotes the state space, ${A}$ the action space, ${T}$ the transition dynamics, ${R}$ the reward function, and $\gamma$ the discount factor. At each timestep $t$, the state $\boldsymbol{s}_t \in {S}$ comprises the agent’s proprioceptive state $\boldsymbol{s}^{\mathrm{p}}_t$ and a goal state $\boldsymbol{s}^{\mathrm{g}}_t$ that specifies the control objective. The action $\boldsymbol{a}_t \in {A}$ represents desired joint angles, which are then executed by a low-level proportional-derivative (PD) controller to actuate the robot.

Following prior work~\cite{zeng2025behavior}, we identify humanoid behavior $\boldsymbol{B}$ as the common objective underlying existing WBC systems and formally define it as a \emph{trajectory over the humanoid proprioceptive states $\boldsymbol{s}^{\mathrm{p}}_t$ and actions $\boldsymbol{a}_t$}. Notably, goal states $\boldsymbol{s}^{\mathrm{g}}_t$ are excluded from this definition, as we interpret them as external specifications that guide or constrain behavior rather than the intrinsic components of behavior itself. 
\begin{equation}
    \boldsymbol{B} \triangleq (\boldsymbol{s}^{\mathrm{p}}_0,\boldsymbol{a}_0,\boldsymbol{s}^{\mathrm{p}}_1,\boldsymbol{a}_1,\cdots,\boldsymbol{s}^{\mathrm{p}}_{T-1},\boldsymbol{a}_{T-1},\boldsymbol{s}^{\mathrm{p}}_T)
\end{equation}
Building upon the formulation above, BFMs intend to decouple behavior learning from task-specific control modes, thereby establishing a unified and scalable objective for learning various behaviors while enabling a versatile control interface that accommodates diverse behavioral specifications of the same underlying behavior. Formally, we structure the pretraining of BFMs as a goal-conditioned reinforcement learning (GCRL) problem, the learning objective of which is to maximize the expected return:
\begin{equation}
    \max_\theta E_{\boldsymbol{\tau}\sim p_\theta(\boldsymbol{\tau})}[\sum_{t=0}^T \gamma^tr_t],
\end{equation}
where $p_\theta(\boldsymbol{\tau})=p(\boldsymbol{s}_0)\prod_{t=0}^{T}p(\boldsymbol{s}_{t+1}|\boldsymbol{s}_{t},\boldsymbol{a}_t)\pi_\theta (\boldsymbol{a}_t|\boldsymbol{s}_t)$ is the distribution over trajectories induced by the model $\pi_\theta$. Existing WBC systems typically tailor goal states $\boldsymbol{s}_t^g$ and reward functions $r_t$ to individual task requirements, failing to establish a unified abstraction across diverse scenarios. In contrast, BFMs aim to 1) identify a proxy task coupled with a general reward formulation that can be shared across a wide range of behavior patterns and control settings; 2) construct a versatile control interface of goal states $\boldsymbol{s}_t^g$ from diverse forms of behavioral specifications that describe the same underlying behavior. 

Intuitively, imitation naturally serves as a general learning paradigm for behavior acquisition, as it aims to reproduce demonstrated trajectories without relying on task-specific requirements or manually designed evaluation protocols. Therefore, we instantiate the proxy task for BFM pretraining as humanoid whole-body motion tracking, where the reward function evaluates how faithfully the humanoid follows the reference motions. This formulation casts behavior learning as a well-defined and measurable optimization problem, thereby establishing a unified and scalable objective across diverse scenarios.

Despite the shared usage of reference motions, the formulation above differs fundamentally from general motion trackers. BFMs employ motion tracking solely as a proxy task for behavior learning rather than as the intended control objective. As a consequence, they are not constrained to a specific form of behavioral specification, whereas conventional motion trackers typically treat whole-body reference states as the only way of applying control and therefore remain specialized for a specific WBC task.

In practice, we optimize the objective of BFM pretraining using the Proximal Policy Optimization (PPO) algorithm~\cite{schulman2017proximal} for its stability and robustness. Nevertheless, the proposed formulation is algorithm-agnostic and can be readily integrated with a wide range of reinforcement learning algorithms.

\vspace{-3mm}
\subsection{Learning Paradigm}
\label{sec:tracking_recipe}
\vspace{-2mm}

Motion tracking has been widely adopted as an effective learning paradigm for humanoid control. Existing approaches, however, differ substantially in their detailed designs, including state representations, control interfaces, and reward formulations. To enable unified and scalable behavior learning together with versatile humanoid control through diverse forms of behavioral specification, we carefully curate a motion-tracking recipe for BFM pretraining, the key components of which are described below.

\textbf{Proprioceptive State Design.} We adopt the asymmetric actor-critic design in PPO to jointly account for efficient policy learning and practical real-world deployment. The proprioceptive state of actor is defined as $\boldsymbol{s}^{\mathrm{p}}_t\triangleq (\omega_t^{\mathrm{root}},{g}_t,{q}_t,\dot{{q}}_t)$ where $\omega_t$ is the base angular velocity, ${g}_t$ the projected gravity, ${q}_t$ current joint position and $\dot{{q}}_t$ current joint velocity. In contrast, critic's proprioception may leverage privileged information in simulators, which is defined as $\hat{\boldsymbol{s}}^{\mathrm{p}}_t\triangleq (h_t,p_t,\theta_t,v_t,\omega_t,q_t,\dot{q}_t)$ where $h_t$ is the root height, $p_t$ the link positions, $\theta_t$ the link orientations, $v_t$ the link linear velocities and $\omega_t$ the link angular velocities. All these whole-body states of the critic are expressed in humanoid's heading frame.

\textbf{Control Interface Design.} To enable versatile humanoid control from diverse forms of behavioral specifications, we design a simple yet expressive control interface based on \emph{masked whole-body target poses in the root-relative Cartesian space}. Specifically, the whole-body target representation is defined as:
$\boldsymbol{x}_t \triangleq
    \left(
    {p}^{\mathrm{ref}}_{t+1} - {p}^{\mathrm{root}}_t,\,
    {p}^{\mathrm{ref}}_{t+1} - {p}_t,\,
    {\theta}^{\mathrm{ref}}_{t+1} \ominus {\theta}^{\mathrm{root}}_t,\,
    {\theta}^{\mathrm{ref}}_{t+1} \ominus {\theta}_t
    \right)
$, which contains the target link poses and their deviations from the current poses, all expressed in the humanoid root frame. We then randomly sample a link-wise mask from a predefined mask set $\mathcal{M}=\{\boldsymbol{m}_0,\boldsymbol{m}_1,\dots,\boldsymbol{m}_7\}$ and construct the actor goal state as $\boldsymbol{s}^{\mathrm{g}}_t
    \triangleq
    \left(
    \boldsymbol{x}_t \odot \boldsymbol{m}_i,\,
    \boldsymbol{m}_i
    \right),\boldsymbol{m}_i \sim \mathcal{M}
$. The masks specify eight control modes, the details of which are provided in the Appendix~\ref{app:control_mode}. In contrast, the critic is provided with complete, unmasked target information to provide more accurate value estimation. Accordingly, we define the critic goal state in the humanoid root frame as
$
    \hat{\boldsymbol{s}}^{\mathrm{g}}_t
    \triangleq
    \left(
    {p}^{\mathrm{ref}}_{t+1} - {p}^{\mathrm{root}}_t,\,
    {p}^{\mathrm{ref}}_{t+1} - {p}_t,\,
    {\theta}^{\mathrm{ref}}_{t+1} \ominus {\theta}^{\mathrm{root}}_t,\,
    {\theta}^{\mathrm{ref}}_{t+1} \ominus {\theta}_t,\,
    {v}^{\mathrm{ref}}_{t+1} - {v}_t,\,
    {\omega}^{\mathrm{ref}}_{t+1} - {\omega}_t
    \right)
$. This asymmetric design is consistent with BFM objective of decoupling behavior learning from concrete specifications, allowing the critic to focus exclusively on motion tracking as the proxy task for BFM pretraining. 

\textbf{Reward Design.} We formulate the reward as a weighted sum of tracking rewards, penalty terms and survival terms. Unlike prior works that either remove root-position tracking or decouple root following from whole-body pose tracking~\cite{liao2025beyondmimic, luo2025sonic}, we require the robot to reproduce reference motions as integrated whole-body trajectories in the global frame. Removing root-position tracking may assign nearly identical learning guidance to behaviors with distinct global semantics, while decoupling root and pose tracking may compromise the coordination of root and pose evolution. Our design instead provides coherent guidance over whole-body behavior, reducing guidance ambiguity and encouraging more consistent behavior acquisition. Details on reward components are provided in Appendix~\ref{app:reward_design}.

\textbf{Domain Randomization.} To enhance the robustness and generalization of our BFMs, we apply domain randomization during training. We randomize physical parameters including friction coefficient, joint default positions, base center-of-mass positions and mass of hands particularly to accommodate potential variations in end-effector configurations. We also periodically apply random perturbations to robot velocities to simulate external pushes. Detailed components are provided in Appendix~\ref{app:dr}.

\vspace{-3mm}
\subsection{Training Data}
\vspace{-2mm}
\label{sec:training_data}

\begin{figure}[t]
    \centering
    \includegraphics[width=\linewidth]{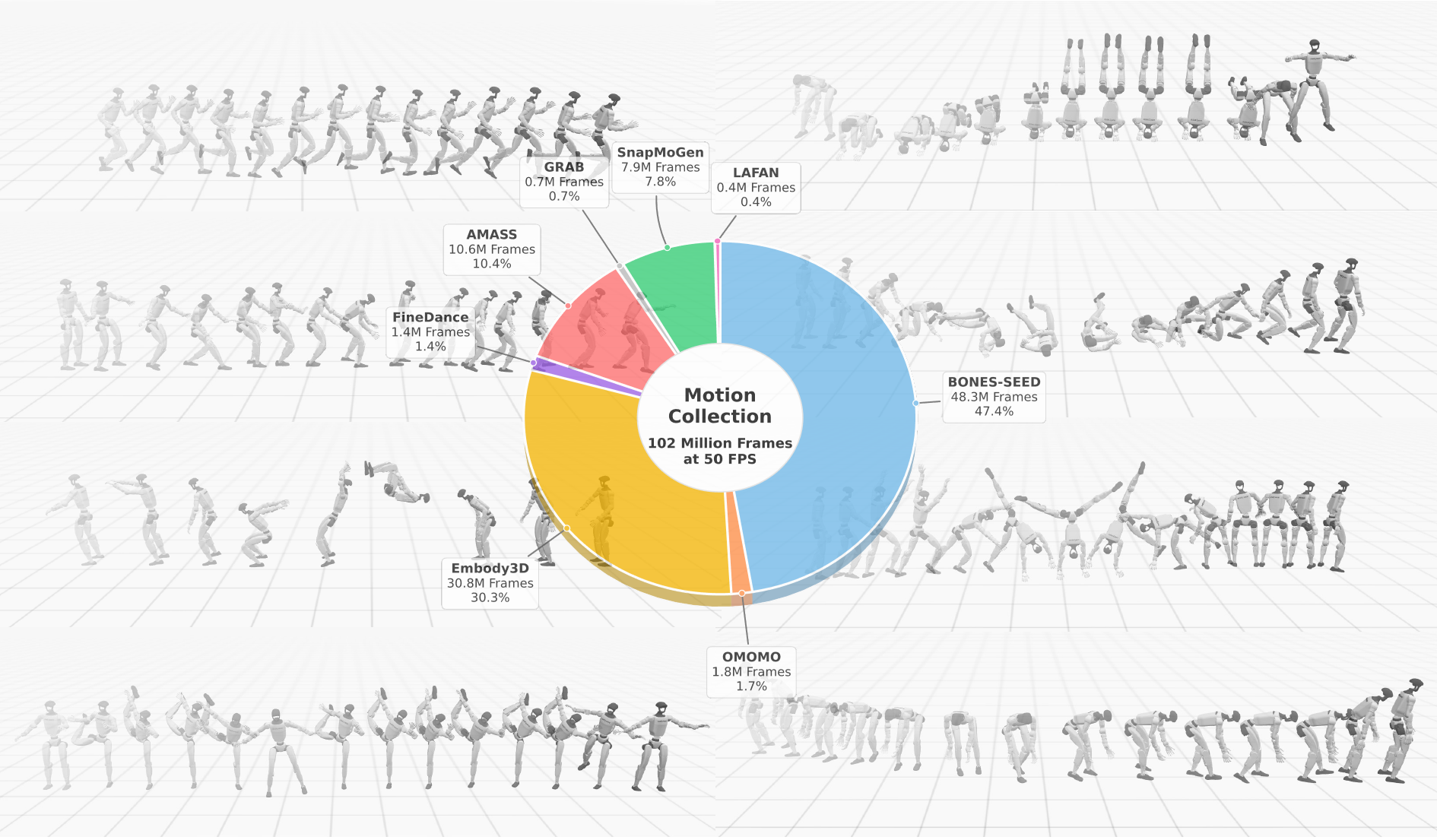}
    \caption{\textbf{Reference Motion Dataset Composition.}  We construct a large-scale human motion dataset by aggregating over 102 million frames at 50 FPS from multiple sources, which are subsequently retargeted to the target humanoid. Random samples are visualized to demonstrate its broad coverage of diverse behavior patterns.}
    \label{fig:dataset}
    \vspace{-2mm}
\end{figure}

After establishing motion tracking with versatile control interface as the learning paradigm, we next describe how the training data is collected and organized for scalable BFM pretraining. Existing studies often equate scaling training data with simply increasing the number of reference motions. Under the PPO framework, however, the effective training data are the on-policy rollouts collected through environment interaction, the scale of which is primarily determined by the degree of environment parallelism and the rollout horizon. Increasing the number of reference motions instead influences the distribution of these rollouts by introducing a wider range of behavior patterns for imitation. This benefit, however, depends critically on the diversity of the reference motion corpus rather than its size alone. For instance, a dataset containing an arbitrarily large number of walking forward motions still represents only a single behavior pattern and therefore provides limited behavioral coverage. Consequently, scalable BFM pretraining requires the synergy of quantity and diversity of the training data. In practice, we increase the number of parallel environments and rollout horizon to collect more on-policy data, while curating reference motions from diverse sources to ensure broad behavioral coverage.

\textbf{Human Motion Collection.} To expand the behavioral coverage of reference motions, we first construct a large-scale human motion collection with fully open-source datasets, including LAFAN~\cite{harvey2020robust}, AMASS~\cite{mahmood2019amass}, OMOMO~\cite{li2023object}, GRAB~\cite{taheri2020grab}, SnapMoGen~\cite{guo2025snapmogen}, FineDance~\cite{li2023finedance}, BONES-SEED~\cite{bones_seed_2026}, and Embody3D~\cite{mclean2025embody}. This collection contains 102M frames at 50 FPS, which is expected to contain rich and coordinated behavior patterns for BFM pretraining. Detailed components may refer to figure~\ref{fig:dataset}.

\textbf{Motion Retargeting.} We adopt a two-stage retargeting pipeline to bridge the embodiment gap between human and humanoid. In the first stage, we align the human and humanoid skeleton by optimizing the human shape coefficient $\beta$ to minimize the difference of corresponding links at the same rest pose. Notably, our reference motion corpus comprises heterogeneous human motion representations, including both SMPL and BVH formats. For SMPL motions, we optimize the built-in shape coefficients, whereas for BVH motions, we optimize the local skeletal offsets. Human skeletons from diverse sources are thereby aligned with the target humanoid morphology. In the second stage, we follow prior work and retarget human motions to the humanoid in a sequential frame-wise manner~\cite{araujo2025retargeting}. For each frame $t$, the humanoid root pose and joint configurations are initialized from the solution of the previous frame and subsequently optimized by solving an inverse kinematics problem that minimizes the weighted pose discrepancy between selected links. Visualization of retargeted motions is also presented in figure~\ref{fig:dataset}.

\textbf{Early Termination and Reference State Initialization.} Early termination and reference state initialization (RSI) jointly shape the on-policy data distribution for BFM pretraining. 
We terminate an episode whenever any humanoid link deviates by more than $0.5\,\mathrm{m}$ from the reference motion in the global frame, thereby preventing the collection of low-quality rollout data. After termination, we adopt the RSI strategy~\cite{peng2018deepmimic}, where the initial states are directly derived from the reference motions. Rather than randomly selecting an arbitrary reference frame, we reinitialize the humanoid with the reference state next to the termination point, bringing it back to the desired trajectory manifold from which it departs. 

\textbf{Adaptive Sampling.} Adaptive sampling intends to bias the data distribution toward challenging motion sequences to improve sample efficiency while maintaining sufficient behavioral coverage to avoid catastrophic forgetting. Inspired by the implementation of ProtoMotions~\cite{ProtoMotions}, we assign a uniform sampling weight $w_0=1$ to each motion sequence at the beginning of training and then periodically evaluate the model on the full training set every $T_{\text{eval}}$ epochs. 
A trajectory is marked as failed if any humanoid link at any timestep deviates by more than $0.5\,\mathrm{m}$ from the reference in the global frame.
Given a discount factor $\beta$ where $0<\beta<1$, the sampling weight of each trajectory after the $t$-th evaluation is updated as
\begin{equation}
    w_t =
    \begin{cases}
    w_{t-1} / \beta^{T_{\text{eval}}}, & \text{if the trajectory fails},\\
    w_{t-1} \cdot \beta^{T_{\text{eval}}}, & \text{if the trajectory succeeds}.
    \end{cases}
\end{equation}
To avoid excessive concentration of the sampling distribution, we clamp the updated sampling weights:
\begin{equation}
    w_t \leftarrow \mathrm{clip}(w_t, w_{\min}, w_{\max}).
\end{equation}
In practice, we set $T_{\mathrm{eval}}=200$, $\beta=0.999$, $w_{\min}=0.03$, and $w_{\max}=1.0$ and normalize the weights to obtain the sampling distribution. This forms a closed loop of evaluation, feedback, and resampling, allowing training to focus more on difficult motions while retaining coverage of the reference corpus.

\vspace{-3mm}
\subsection{Model Architecture}
\vspace{-2mm}
\label{sec:model}

\begin{figure}[t]
    \centering
    \includegraphics[width=\linewidth]{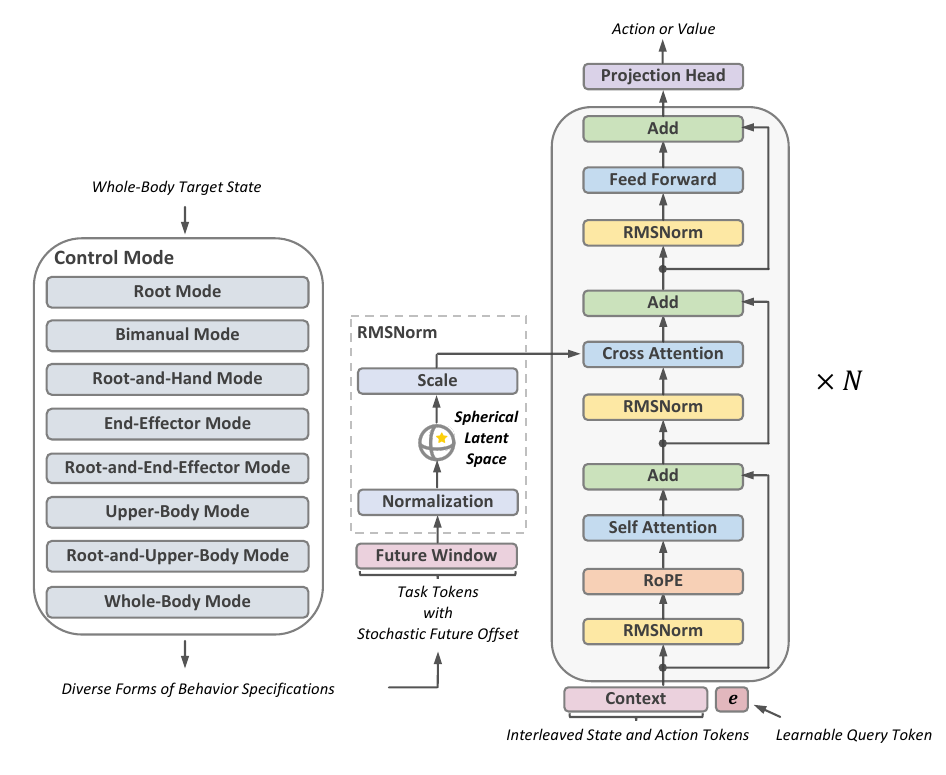}
    \caption{\textbf{Overview of Model Architecture.} We introduce an expressive and scalable architecture termed the \emph{Humanoid Transformer}, which facilitates the natural emergence of structured behavioral representations.}
    \label{fig:model_arch}
    \vspace{-2mm}
\end{figure}

We finally introduce the \emph{Humanoid Transformer}, an expressive and scalable architecture tailored for BFM pretraining. To provide richer temporal information for behavior learning, we first extend the proprioceptive states, goal states, and actions into finite temporal windows. These inputs are then encoded by modality-specific tokenizers, the parameters of which are shared across all inputs of the same modality. The resulting proprioception and action tokens are interleaved to form the context sequence, followed by a learnable query token for action or value prediction. During self-attention, the query token is prevented from being attended to by the context tokens, while retaining full attention over the entire context sequence to aggregate historical information. Goal states, together with their temporal offsets relative to the current timestep, are concatenated along the temporal dimension, tokenized, and injected into the backbone through cross-attention, providing a flexible mechanism for conditioning the model on behavioral specifications. An overview of the architecture is presented in figure~\ref{fig:model_arch}.

Notably, we employ RMSNorm~\cite{zhang2019root} to normalize the goal embeddings onto a continuous and bounded hypersphere, thereby naturally inducing a structured latent representation space for behavioral intentions. Rather than explicitly regularizing this latent space with auxiliary objectives, it is shaped solely through the objective of motion tracking during pretraining. Interestingly, a similar structural design of the latent space has also been explored in generative frameworks for image generation~\cite{yue2026image}.

We employ the same backbone architecture for both the actor and the critic, while adopting different temporal offset schemes. Let index 0 denote the immediate next target frame. The future window of actor contains the next five consecutive frames with indices $\{0,1,2,3,4\}$, together with an additional frame randomly sampled from the range $[5,32]$. This \emph{stochastic future offset} provides a practical mechanism for mitigating system delays during real-time deployment, as the last future frame can be dynamically adjusted according to the timestamp of the received control signals. In contrast, the critic adopts fixed exponentially spaced indices $\{0,1,2,4,8,16,32\}$, providing access to both fine-grained short-term dynamics and long-horizon outcomes. This design improves value estimation while maintaining a comparable token budget. More details on the model architecture are provided in Appendix~\ref{app:model}.
\section{Experiments}\label{sec:experiments}
\vspace{-2mm}

In this section, we first investigate the scaling behavior of BFMs from the perspectives of training data and model architecture. Building on these findings, we then analyze the learned latent space to show that structured representations of behavioral intentions emerge naturally without auxiliary objectives and are further refined through scaling. Finally, we benchmark the resulting BFMs against existing humanoid controllers to evaluate their effectiveness as humanoid foundation controllers.
    
\vspace{-2mm}
\subsection{Experiment Setup}
\vspace{-2mm}

\textbf{Environments.} We conduct experiments on Unitree G1~\cite{unitreeg1}, which stands 1.3 meters tall and has 29 degrees of freedom. All models are trained in IsaacLab~\cite{mittal2025isaac} and evaluated in MuJoCo~\cite{todorov2012mujoco} to assess robustness under dynamic transfer. Details on real-world deployment are provided in Appendix \ref{app:real_robot}.

\textbf{Evaluation Protocol.} BFM pretraining is instantiated as motion tracking, which naturally provides direct evaluation protocols by measuring how accurately the humanoid tracks the control signals under diverse control modes. Accordingly, we report the success rate (Succ) together with global and local tracking errors. These metrics are computed only over the activated links specified by each control mode, as the remaining links are left unspecified and are therefore naturally inpainted. A motion sequence is deemed unsuccessful if any selected link deviates from its desired state in the global frame by more than a predefined threshold. G-MPKPE and G-MPKRE denote the mean per-keypoint position and rotation errors of the selected links in the global frame, whereas L-MPKPE and L-MPKRE measure the corresponding root-relative errors after removing the horizontal translation offsets and heading differences~\cite{liao2025beyondmimic}. Formal definitions of all the evaluation metrics are provided in Appendix~\ref{app:evaluation_metrics}.

\textbf{Evaluation Dataset.} We evaluate all the models on two test sets. The first ({BONES Test Set}) consists of 10,000 held-out motion sequences sampled from the BONES dataset, forming a source-aligned benchmark of approximately 3.6 million frames at 50 FPS. The second ({Ours Test Set}) comprises high-quality, previously unseen motions, including 839 sequences captured using Xsens~\cite{schepers2018xsens} and 810 sequences from 100Style~\cite{mason2022local}. Together, these cross-source sequences comprise around 4.5 million frames at 50 FPS.

\vspace{-2mm}
\subsection{Scaling with Training data}
\vspace{-2mm}

We first study how training data affects the scaling behavior of BFMs from two complementary aspects: scaling on-policy data collection and scaling reference motions under the PPO-based motion-tracking framework. Specifically, scaling on-policy data collection primarily increases the \emph{quantity} of training data per update, whereas scaling reference motions is expected to enrich the \emph{diversity} by introducing a broader range of behavioral patterns, thereby shaping the data distribution encountered during training.

\vspace{-5mm}
\subsubsection{Scaling On-policy Data Collection}
\vspace{-2mm}

\begin{figure}[t!]
    \centering
    \includegraphics[width=\linewidth]{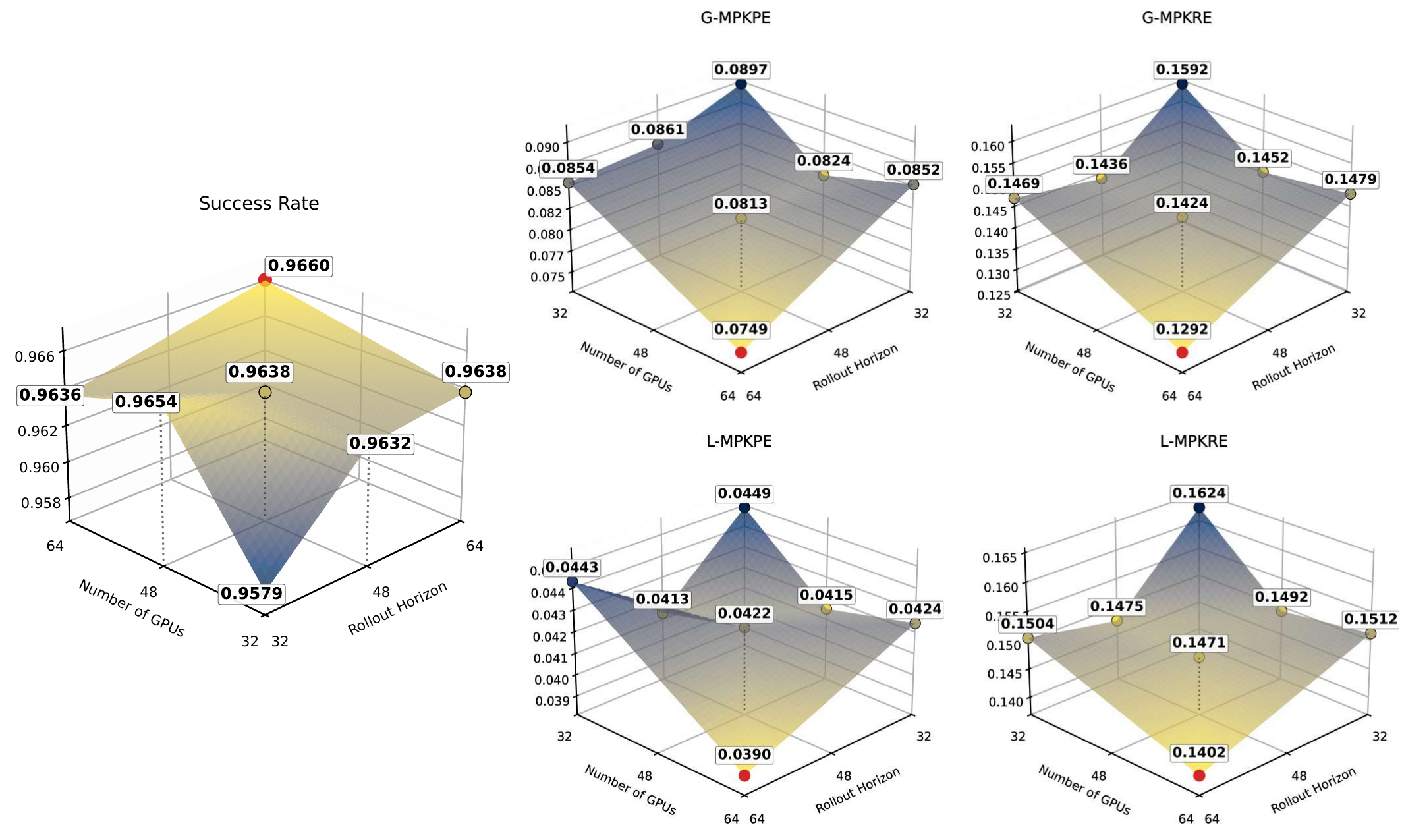}
    \caption{Results of scaling \textbf{on-policy data collection} on the \textbf{BONES Test Set} under \textbf{whole-body control mode}. Best outcomes with the highest success rate and lowest tracking errors are highlighted with red circles.}
    \label{fig:scale_on_policy_bones_whole_body}
    \vspace{-2mm}
\end{figure}

\begin{figure}[t!]
    \centering
    \includegraphics[width=\linewidth]{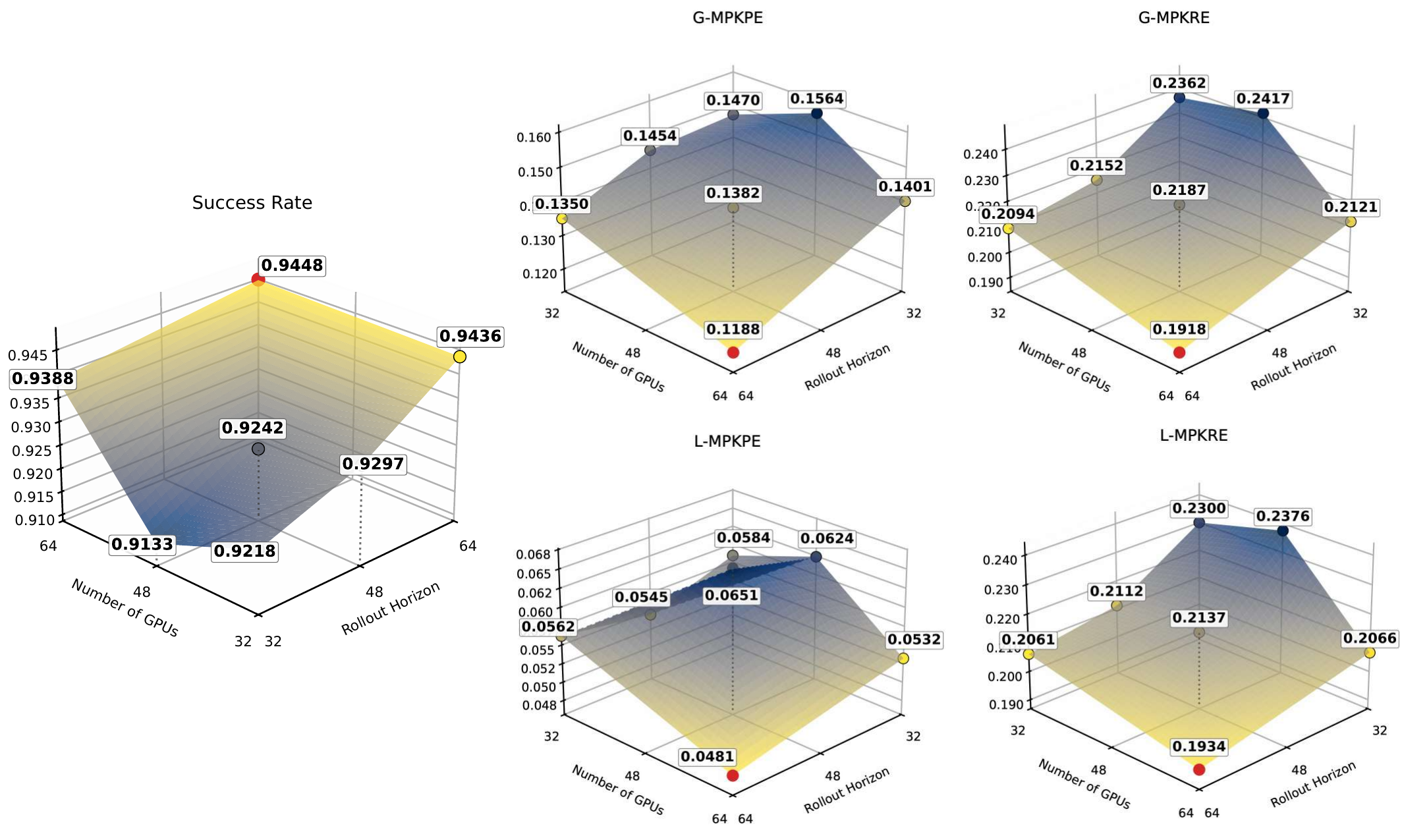}
    \caption{Results of scaling \textbf{on-policy data collection} on the \textbf{Ours Test Set} under \textbf{whole-body control mode}. Best outcomes with the highest success rate and lowest tracking errors are highlighted with red circles.}
    \label{fig:scale_on_policy_ours_whole_body}
    \vspace{-2mm}
\end{figure}

We scale on-policy data collection by increasing both the \emph{width} and \emph{depth} of environmental interaction. With the number of environments per GPU held constant, the \emph{width} is increased by scaling the number of parallel GPUs, whereas the \emph{depth} is increased by extending the rollout horizon of each environment.

To evaluate the effect of scaling on-policy data collection, we keep all other training configurations fixed across experiments while varying the rollout horizon and the number of GPUs across three levels. We report results on two test sets under four representative control modes. The whole-body control results are presented in figure~\ref{fig:scale_on_policy_bones_whole_body} and~\ref{fig:scale_on_policy_ours_whole_body}, whereas the results for the other three control modes, together with the experimental designs adopted to ensure a fair comparison, are further provided in Appendix~\ref{app:scaling_on_policy_data_collection}.

As shown by the results, jointly scaling the \emph{width} and \emph{depth} dimensions substantially improves performance across diverse control modes, with the largest configuration (64 GPUs and a rollout horizon of 64) achieving the best performance in nearly all experiments. We attribute these gains primarily to the increased amount of on-policy experience collected for each policy update, which provides more reliable estimates of quantities involved in PPO optimization, thereby leading to more effective policy updates.

However, scaling either dimension alone does not always produce a consistent trend of improvement. This observation suggests that the benefits of scaling on-policy data collection depend not only on the total quantity of collected experience but also on how it is accumulated. In other words, effectively scaling on-policy data collection may require an appropriate balance between the \emph{width} and \emph{depth} dimensions, rather than increasing either one in isolation. Understanding the intricate interplay between these two complementary factors remains an interesting direction for future investigation.

\vspace{-5mm}
\subsubsection{Scaling Reference Motions}
\vspace{-2mm}

\begin{figure}[t!]
    \centering
    \includegraphics[width=\linewidth]{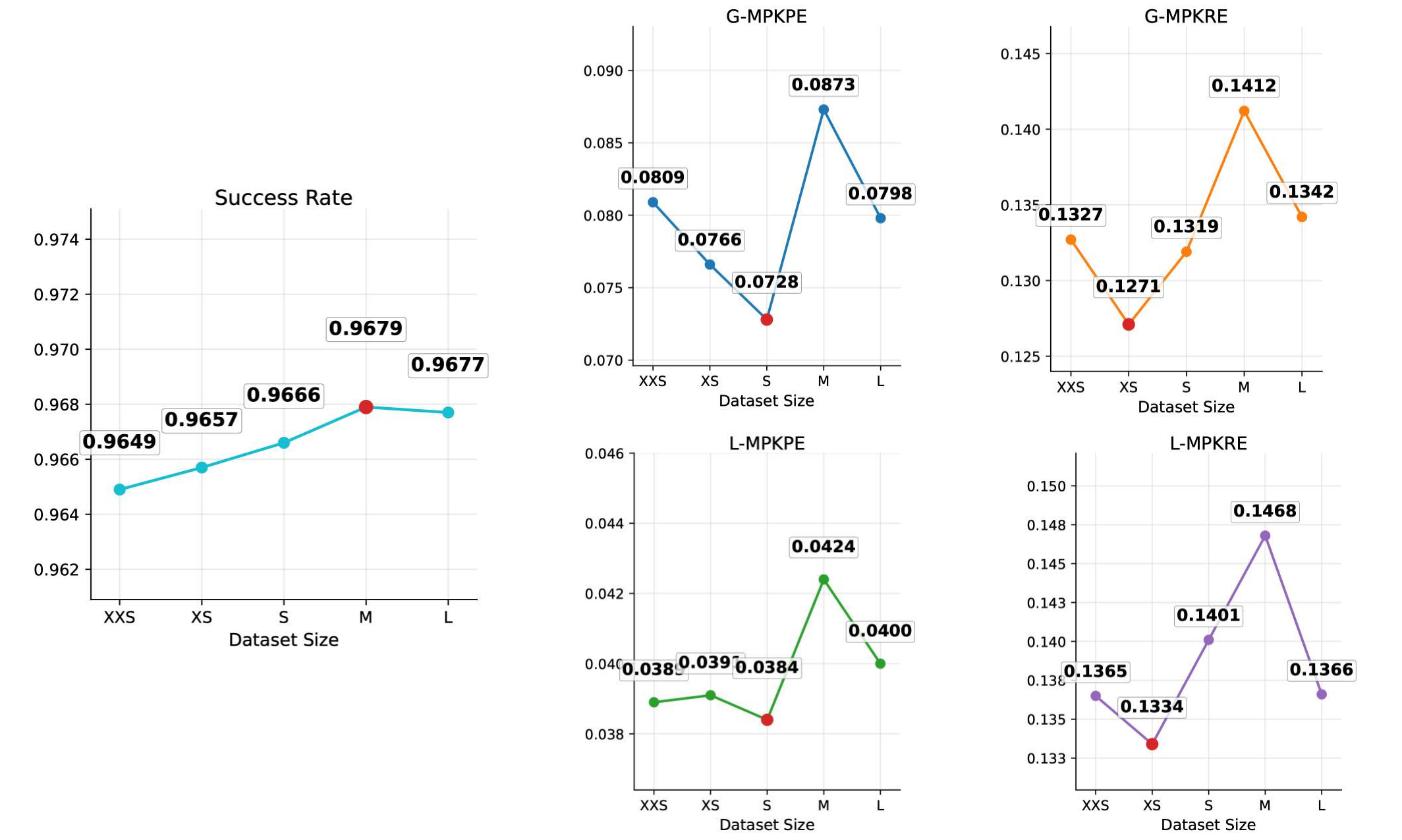}
    \caption{Results of scaling \textbf{reference motions} on the \textbf{BONES Test Set} under \textbf{whole-body mode}. Best outcomes corresponding to the highest success rate and lowest tracking errors are highlighted with red circles.}
    \label{fig:scale_ref_whole_body_bones}
    \vspace{-4mm}
\end{figure}

\begin{figure}[t!]
    \centering
    \includegraphics[width=\linewidth]{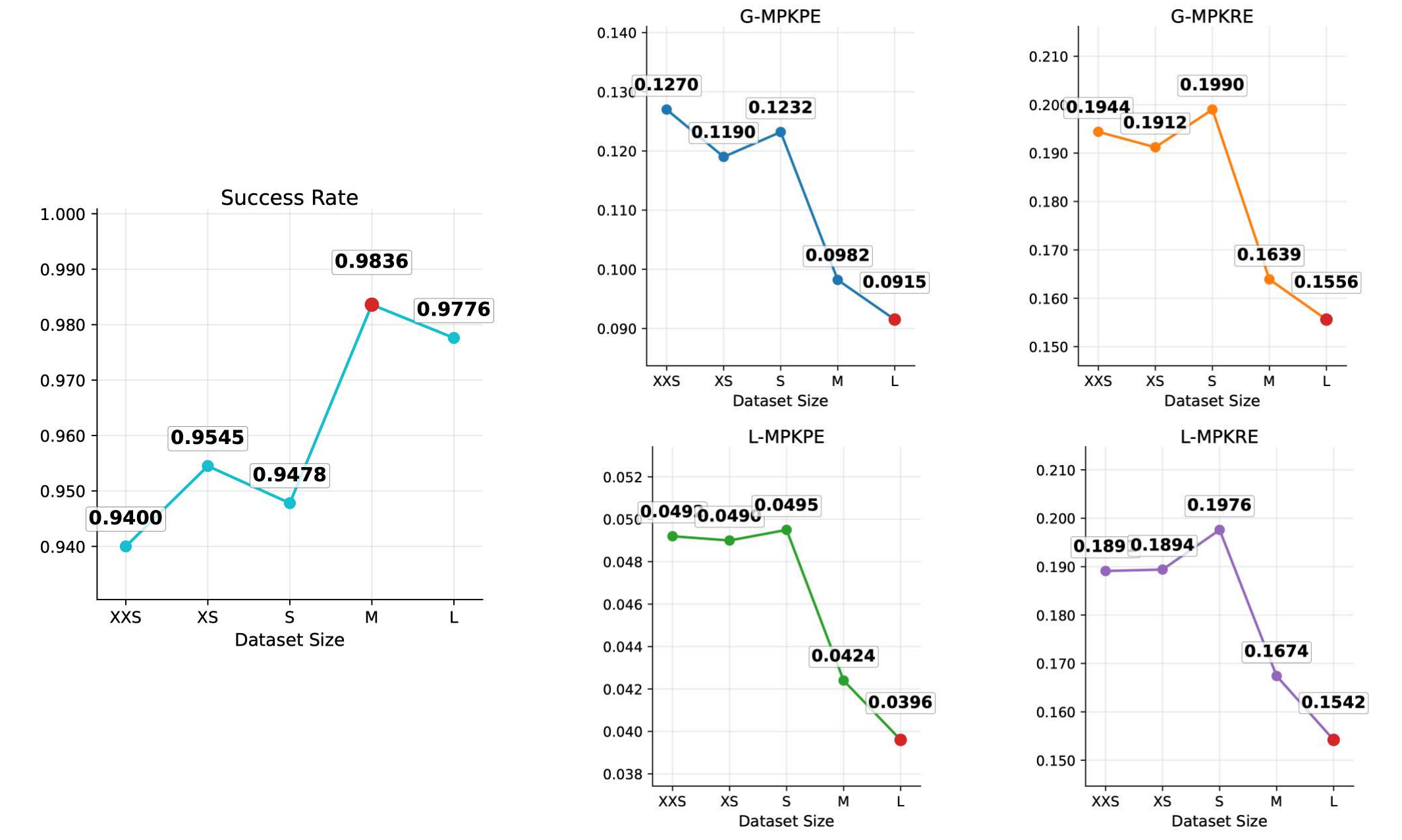}
    \caption{Results of scaling \textbf{reference motions} on the \textbf{Ours Test Set} under \textbf{whole-body mode}. Best outcomes corresponding to the highest success rate and lowest tracking errors are highlighted with red circles.}
    \label{fig:scale_ref_whole_body_ours}
    \vspace{-4mm}
\end{figure}

We evaluate the effect of scaling reference motions by partitioning the full dataset into five nested subsets of increasing size, such that each larger subset fully contains all smaller ones. These subsets are denoted as {XXS}, {XS}, {S}, {M}, and {L}. Specifically, {XXS}, {XS}, and {S} correspond to one-third, two-thirds, and the complete BONES-SEED dataset excluding the held-out test split, respectively. The {M} subset comprises all reference motion datasets except Embody3D, whereas {L} contains the complete collection.

All other training configurations are kept fixed across experiments, and the optimal on-policy data configuration is adopted throughout. We still report results of two test sets under four representative control modes. The whole-body control results are presented in figure~\ref{fig:scale_ref_whole_body_bones} and~\ref{fig:scale_ref_whole_body_ours}, whereas those for the remaining three control modes, together with the detailed experimental designs are provided in Appendix~\ref{app:scaling_reference_motions}.

\begin{wraptable}{r}{0.44\textwidth}
\vspace{-2mm}
\centering
\begin{tabular}{cc}
\toprule
Partition & Occupancy Rate \\
\midrule
XXS & 0.9365 \\
XS & 0.9350 \\
S & 0.9365 \\
M & 0.9825 \\
L & 0.9995 \\
\bottomrule
\end{tabular}
\vspace{2mm}
\caption{Occupancy Rates of Training Partitions}
\label{tab:occupancy}
\vspace{-2mm}
\end{wraptable}

Before interpreting the detailed evaluation results, we first quantitatively analyze the behavioral characteristics of the training partitions. Specifically, we extract behavioral features from all motion frames and fit a clustering model using features randomly sampled from the complete L partition. An equal number of features are subsequently sampled from each partition and assigned to the nearest cluster defined by the shared clustering model. The \textbf{occupancy rate} is therefore defined as the ratio of occupied clusters to the total number of clusters.

As shown in table~\ref{tab:occupancy}, the occupancy rate remains nearly constant from XXS to S, but increases substantially from S to L. This observation indicates that scaling from XXS to S may primarily increase the amount of reference motions while yielding little measurable expansion in the behavioral coverage. In contrast, scaling from S to L not only increases the amount of reference motions but also broadens the range of represented behaviors. Additional details of this analysis are provided in the Appendix~\ref{app:scaling_reference_motions}.

The behavioral coverage analysis naturally divides reference-motion scaling into two distinct regimes: \emph{homogeneous scaling} increases the amount of reference motions while leaving the measured behavioral coverage nearly unchanged (XXS$\rightarrow$S), whereas \emph{heterogeneous scaling} increases both the amount of reference motions and the measured behavioral coverage (S$\rightarrow$L). As shown in the experiments, homogeneous scaling yields only marginal improvements on both the BONES and Ours Test Set, despite increasing the amount of reference motions from the same domain. In contrast, heterogeneous scaling produces little additional benefit on the BONES Test Set but leads to substantial performance gains on the Ours Test Set. Notably, significant improvements are observed only when scaling markedly expands measured behavioral coverage, and only on the benchmark containing motions from novel data sources.

These observations suggest that the effectiveness of scaling reference motions depends not only on the amount of reference motions, but also on whether scaling expands the behavioral coverage and whether the newly introduced behaviors are relevant to the target evaluation benchmark. Intuitively, a BFM trained exclusively on walking motions is unlikely to perform a cartwheel without prior exposure to similar behaviors. Expanding the training set to include cartwheel-like motions can potentially improve performance on such behaviors. Conversely, if the current objective is to improve walking performance, introducing cartwheel motions into the training set is unlikely to provide significant additional benefit.

The findings above provide several practical insights for future research: 1) If the objective is to improve performance within a specific domain, simply increasing the amount of in-domain reference motions is likely to yield diminishing returns. Once the dominant behavioral patterns of the target domain are well represented, further gains are more likely to be achieved through a more carefully curated training set with fine-grained coverage of domain-specific behaviors. 2) Since no finite dataset can exhaust the space of possible behaviors, rare and long-tail behavioral patterns will inevitably remain absent. A more meaningful question is therefore not how to achieve a fine-grained behavioral coverage but how to curate a finite training set that achieves an effective balance between data efficiency and behavioral coverage. 3) If the objective is to improve the general capabilities of BFMs, incorporating diverse data sources appears to be a practical strategy. Although additional data sources do not inherently guarantee broader behavioral coverage, they often contribute complementary behavioral patterns---for example, LAFAN primarily provides locomotion behaviors, whereas OMOMO introduces human-object interaction behaviors. Combining such complementary sources can therefore provide a stronger behavioral foundation for BFM pretraining by exposing the model to a broader range of behavioral patterns.

\vspace{-2mm}
\subsection{Scaling with Model Architecture}
\vspace{-2mm}
\label{sec:scale_models}

\begin{figure}[t!]
    \centering
    \includegraphics[width=\linewidth]{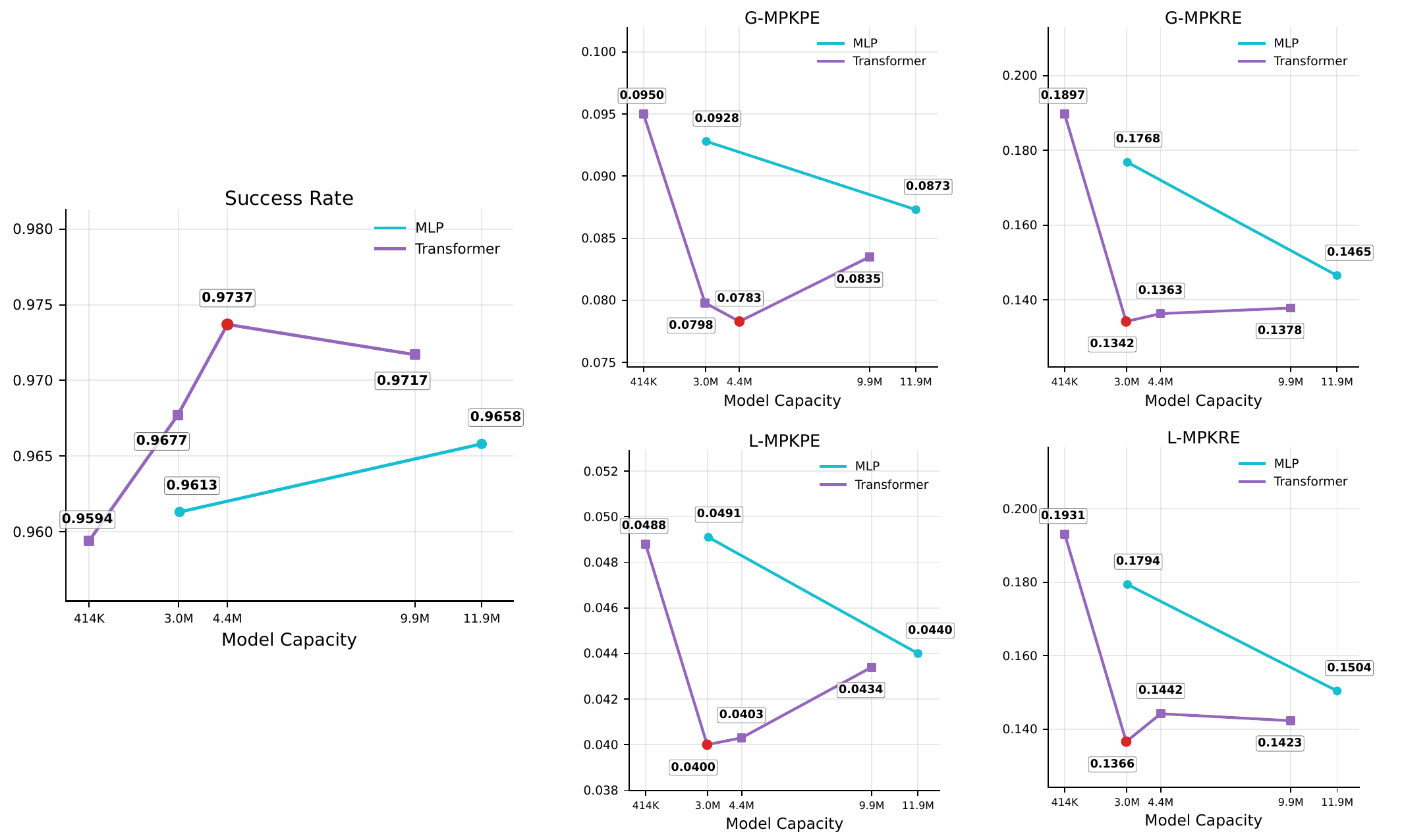}
    \caption{Results of scaling \textbf{model architectures} on the \textbf{BONES Test Set} under \textbf{whole-body control mode}. Best outcomes with the highest success rate and lowest tracking errors are highlighted with red circles.}
    \label{fig:scale_model_bones_whole_body}
    \vspace{-2mm}
\end{figure}

\begin{figure}[t!]
    \centering
    \includegraphics[width=\linewidth]{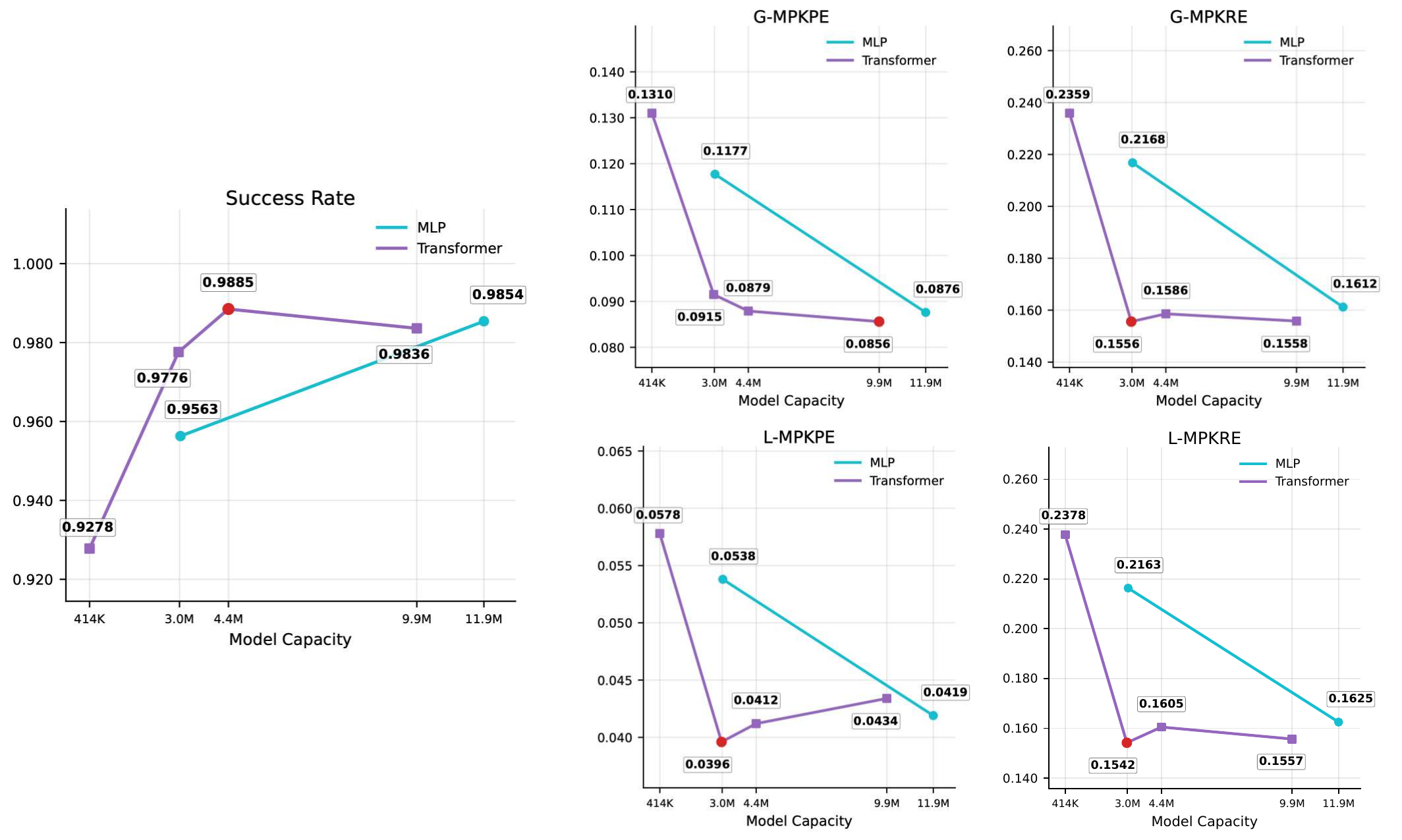}
    \caption{Results of scaling \textbf{model architectures} on the \textbf{Ours Test Set} under \textbf{whole-body control mode}. Best outcomes with the highest success rate and lowest tracking errors are highlighted with red circles.}
    \label{fig:scale_model_ours_whole_body}
    \vspace{-3mm}
\end{figure}

We next investigate the effect of model architectures on scaling BFMs. Specifically, we intend to examine whether the proposed Humanoid Transformer achieves superior performance compared with previously adopted MLP architecture, and whether increasing its capacity leads to further performance gains. 

All other training configurations are kept fixed across experiments, including the optimal on-policy data and reference motion configurations. We report results on both test sets under four representative control modes. The whole-body control results are presented in figures~\ref{fig:scale_model_bones_whole_body} and~\ref{fig:scale_model_ours_whole_body}, while those for the remaining three modes, together with the detailed experimental setup are provided in Appendix~\ref{app:scaling_models}. 

As shown in the results, the Humanoid Transformer generally achieves higher success rates and lower tracking errors than the MLP across diverse experiments. Notably, the medium-sized Humanoid Transformer already achieves performance comparable to, or even better than, the substantially larger MLP, while further increasing the Transformer size yields only diminishing returns. These results suggest that the proposed architecture serves as a stronger backbone for BFM pretraining, achieving superior performance through more effective utilization of model capacity under the same optimization budget.

However, we also observe that scaling Humanoid Transformer does not consistently improve performance across all the control modes. While some modes continue to benefit from increased model capacities, others exhibit early saturation at relatively modest scales. We hypothesize that this behavior arises from the imbalance in learning difficulty among different control modes. For example, the whole-body mode may be easier to learn than the root-and-end-effector mode, as the latter requires the model to infer natural whole-body behaviors from only sparse constraints. As training progresses, optimization trade-offs between these modes may become increasingly pronounced, allowing the model to further improve certain modes at the expense of slight performance degradation in others. To better understand this phenomenon, we provide a clearer analysis of the underlying learning dynamics in Section~\ref{sec:latent_analysis}.

\vspace{-3mm}
\subsection{Latent Analysis}
\vspace{-2mm}
\label{sec:latent_analysis}

\begin{table}[t]
\renewcommand{\arraystretch}{1.0}
    \centering
    \begin{tabular}{ccccccc}
        \toprule
        \textbf{Test Set} & \textbf{Noise Level} & \textbf{Success Rate} $\uparrow$ & \textbf{G-MPKPE} $\downarrow$ & \textbf{G-MPKRE} $\downarrow$ & \textbf{L-MPKPE} $\downarrow$ & \textbf{L-MPKRE} $\downarrow$ \\
        \midrule
        \multirow{5}{*}{\textbf{BONES}}
        & 0 & 0.9717 & 0.0835 & 0.1378 & 0.0434 & 0.1423 \\
        & 5 & 0.9720 & 0.0846 & 0.1389 & 0.0437 & 0.1438 \\
        & 10 & 0.9710 & 0.0883 & 0.1450 & 0.0447 & 0.1501 \\
        & 15 & 0.9701 & 0.0956 & 0.1565 & 0.0468 & 0.1617 \\
        & 20 & 0.9646 & 0.1067 & 0.1736 & 0.0505 & 0.1784 \\
        \midrule
        \multirow{5}{*}{\textbf{Ours}}
        & 0 & 0.9836 & 0.0856 & 0.1558 & 0.0434 & 0.1557 \\
        & 5 & 0.9830 & 0.0877 & 0.1567 & 0.0435 & 0.1565 \\
        & 10 & 0.9776 & 0.0970 & 0.1640 & 0.0450 & 0.1631 \\
        & 15 & 0.9715 & 0.1100 & 0.1744 & 0.0470 & 0.1724 \\
        & 20 & 0.9400 & 0.1387 & 0.2033 & 0.0543 & 0.1991 \\
        \bottomrule
    \end{tabular}
    \caption{\textbf{Evaluation of Latent Space Robustness.} Latent representations on the hypersphere are perturbed by rotating each latent vector along a randomly sampled direction through an angle specified by the noise level.}
    \label{tab:robustness}
    \vspace{-4mm}
\end{table}

Beyond evaluating the performance of the proposed Humanoid Transformer, we further investigate whether structured latent representations can emerge without introducing auxiliary objectives. In this work, a latent space is considered structured if it satisfies: 1) Locality. Temporally close commands should be mapped to nearby latent representations, preserving local neighborhood relationships and forming a locally continuous latent manifold. 2) Global Organization. Beyond preserving local neighborhood, the latent space should exhibit a coherent global structure, in which distinct behavioral intentions are organized into distinguishable regions rather than being arbitrarily intermixed. 3) Robustness. The latent representations should be resilient to perturbations, such that moderate noise on the latent representations does not substantially affect the model's interpretation of the intended behavior.

\begin{wrapfigure}{r}{0.5\textwidth}
\centering
\vspace{-8mm}
\includegraphics[width=0.43\textwidth]{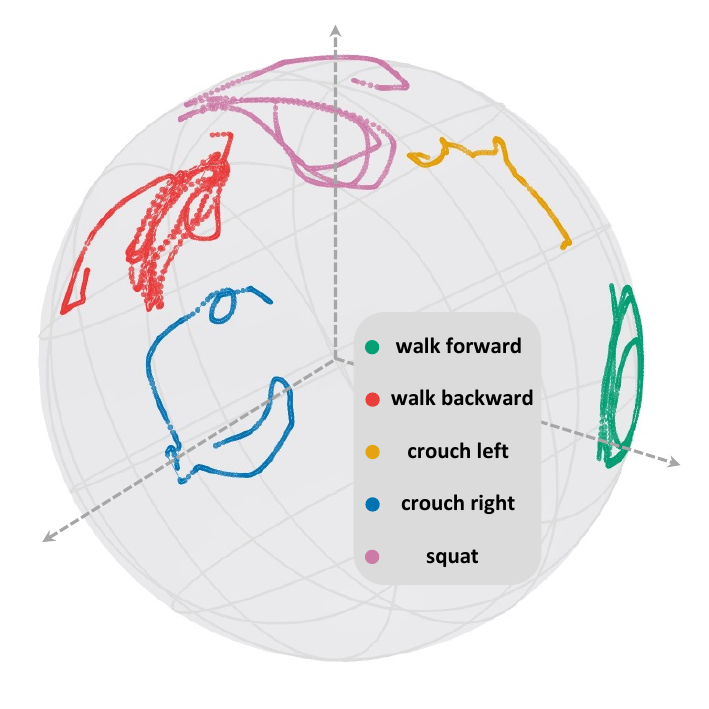}
\vspace{-5mm}
\caption{Visualization of the Latent Space.}
\label{fig:sphere}
\vspace{-8mm}
\end{wrapfigure}
As introduced in Section~\ref{sec:model}, the latent space of the Humanoid Transformer is designed as a continuous manifold on the hypersphere. Throughout this paper, we focus on the unit hypersphere, which is obtained by removing the multiplicative scaling factor in the RMSNorm module and therefore remains consistent across different layers. To examine the locality and global organization of this latent space, we extract the normalized latent representation corresponding to the first future target at each timestep of a motion sequence. The resulting latent representations are then projected into a three-dimensional space using a random Gaussian projection matrix and subsequently normalized to ensure unit length for visualization.

According to the visualization results in figure~\ref{fig:sphere}, the latent space exhibits both locality and global organization across five selected motion sequences from the complete motion collection. From a local perspective, each motion sequence forms a continuous and smooth trajectory on the latent manifold, without abrupt transitions or high-frequency fluctuations. From a global perspective, trajectories corresponding to behavioral intentions in different directions occupy distinguishable regions of the latent space rather than being arbitrarily intermixed. Furthermore, the latent representations preserve the underlying directional relationships among behaviors: walking forward and backward are distributed on opposite sides of one latent dimension, while crouching left and right are separated along another. The visualization above provides qualitative evidence of the locality and global organization of the learned latent space, while the quantitative results in table~\ref{tab:robustness} show that the model maintains strong performance even when relatively large perturbations are introduced into the latent representations. This observation provides empirical evidence for the robustness of the learned latent space, suggesting that the model can preserve its interpretation of the underlying behavioral intentions despite moderate latent noise.

\begin{figure}[t!]
    \centering
    \includegraphics[width=0.75\linewidth]{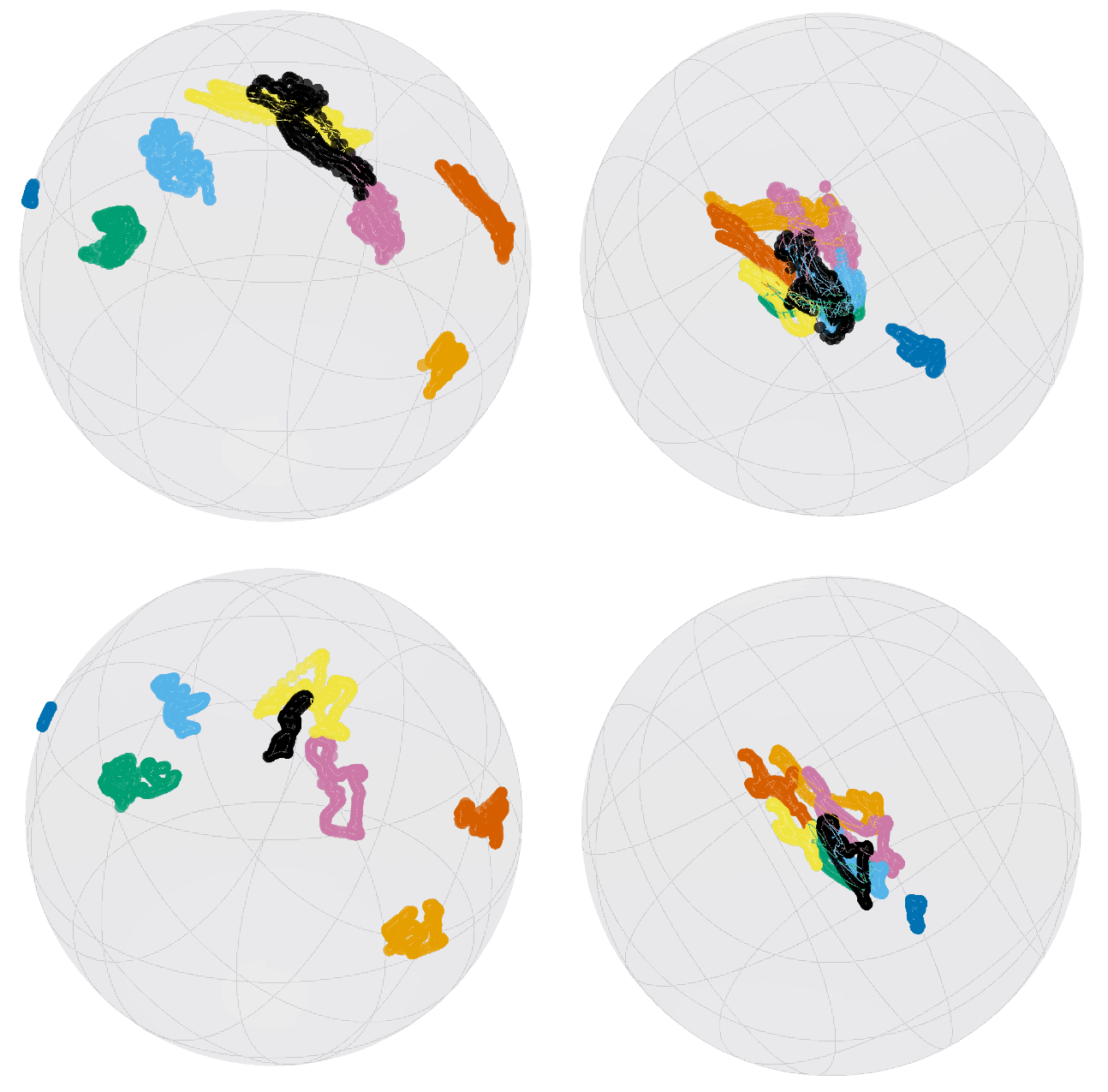}
    \caption{\textbf{Comparison of Latent-space Visualizations across Multiple Control Modes for Models of Different Sizes.} Each row corresponds to the same motion sequence, and each column corresponds to a different model checkpoint. The left and right columns show the relatively small and relatively large models, respectively. Colors indicate different control modes, with the same color representing the same mode across all visualizations.}
    \label{fig:convergence}
    \vspace{-3mm}
\end{figure}

In addition to the structural analysis, we also examine how diverse behavioral specifications are organized within this shared representation space and how this organization evolves as BFMs scale with stronger backbones. Specifically, we compare latent-space visualizations of multiple control modes over randomly sampled motion sequences between a relatively small model and a relatively large model. As shown in figure~\ref{fig:convergence}, {increasing model capacity leads to convergence of the latent representations across different control modes}. This observation may help explain the scaling behavior reported in Section~\ref{sec:scale_models}, where improved performance on certain control modes is accompanied by slight degradation on others, potentially reflecting a strong coupling among control modes in the shared latent space.

% \vspace{-1mm}
\subsection{Model Performance Benchmarking}
\vspace{-2mm}

\begin{table}[t]
\renewcommand{\arraystretch}{1.0}
    \centering
    \begin{tabular}{ccccccc}
        \toprule
        \textbf{Test Set} & \textbf{Method} & \textbf{Success Rate} $\uparrow$ & \textbf{G-MPKPE} $\downarrow$ & \textbf{G-MPKRE} $\downarrow$ & \textbf{L-MPKPE} $\downarrow$ & \textbf{L-MPKRE} $\downarrow$ \\
        \midrule
        \multirow{6}{*}{\textbf{BONES}}
        & GMT & 0.4407 & 0.5674 & 0.4789 & 0.0761 & 0.3177 \\
        & TWIST & 0.4226 & 0.6212 & 0.5487 & 0.0704 & 0.3337 \\
        & SONIC$^\dagger$ & 0.9239 & 0.1740 & 0.1810 & 0.0436 & 0.1762 \\
        & BFM-Bym & 0.9644 & 0.1005 & 0.1406 & 0.0406 & 0.1360 \\
        & BFM-Local & 0.7286 & 0.2281 & \textbf{0.1279} & \textbf{0.0398} & \textbf{0.1312} \\
        & BFM-Global & \textbf{0.9677} & \textbf{0.0798} & 0.1342 & 0.0400 & 0.1366 \\
        \midrule
        \multirow{6}{*}{\textbf{Ours}}
        & GMT & 0.0855 & 1.8693 & 0.9829 & 0.0896 & 0.3158 \\
        & TWIST & 0.0831 & 2.2744 & 1.0705 & 0.0596 & 0.3002 \\
        & SONIC & 0.5937 & 0.5035 & 0.2098 & 0.0430 & 0.1974 \\
        & BFM-Bym & 0.9709 & 0.1224 & 0.1648 & 0.0403 & 0.1592 \\
        & BFM-Local & 0.2553 & 0.4731 & \textbf{0.1473} & \textbf{0.0388} & \textbf{0.1474} \\
        & BFM-Global & \textbf{0.9776} & \textbf{0.0915} & 0.1556 & 0.0396 & 0.1542 \\
        \bottomrule
    \end{tabular}
    \caption{\textbf{Performance Benchmark on Whole-body Control.} We compare the 3M-parameter variant of the proposed BFM with off-the-shelf whole-body controllers and a self-curated ablation baseline (BFM-Bym), trained using the reward design of BeyondMimic and evaluated under the global control mode. The best result for each metric is highlighted in bold. $\dagger$: The randomly sampled BONES Test Set may overlap with SONIC's training set.}
    \label{tab:benchmark}
    \vspace{-3mm}
\end{table}

To benchmark the performance of the proposed BFM, we select whole-body control as a common evaluation task and compare our model with existing whole-body controllers. The baselines include several off-the-shelf models, namely GMT~\cite{chen2025gmt}, TWIST~\cite{ze2025twist,ze2025twist2}, and SONIC~\cite{luo2025sonic}. We further include an ablation baseline trained using the same pipeline but with the reward design of BeyondMimic~\cite{liao2025beyondmimic}, allowing us to evaluate the contribution of the proposed reward formulation. As shown in table~\ref{tab:benchmark}, the proposed BFM consistently achieves higher success rates and lower tracking errors across both local and global control modes, outperforming all competing methods by a clear margin. Notably, the proposed reward formulation consistently yields lower global tracking errors than the ablation baseline, suggesting that tracking integrated whole-body trajectories in the global frame may provide more coherent behavioral guidance without compromising the coordination between root motion and whole-body pose evolution. Detailed descriptions of the local and global control modes are provided in Appendix~\ref{app:real_robot}.
\vspace{-7mm}
\section{Discussion}\label{sec:discussion}
\vspace{-4mm}
In this paper, we present a systematic investigation of the scaling behavior of Behavior Foundation Models for humanoid robots by revisiting three fundamental aspects: the learning paradigm, behavioral data, and model architecture. Through extensive empirical studies, we derive an effective scaling recipe that yields substantial improvements in control fidelity, execution robustness, and generalization capability. Nevertheless, the proposed framework has its limitations. Although we curate a unified control interface with diverse control modes, it remains unclear whether this interface is the most appropriate abstraction and how these modes should be integrated with future high-level policies. Consequently, the design of BFMs should continue to evolve alongside advances in high-level policy learning. Moreover, the current infrastructure for BFM pretraining remains relatively preliminary and constrained, limiting its scalability to broader settings. We hope that this work provides a foundation for systematic development of BFMs, ultimately contributing to the realization of general-purpose humanoid intelligence.
{

\bibliography{iclr2026_conference}

\begin{thebibliography}{10}

\bibitem{huang2025learning}
Tao Huang, Junli Ren, Huayi Wang, Zirui Wang, Qingwei Ben, Muning Wen, Xiao Chen, Jianan Li, and Jiangmiao Pang.
\newblock Learning humanoid standing-up control across diverse postures.
\newblock {\em arXiv preprint arXiv:2502.08378}, 2025.

\bibitem{he2025learning}
Xialin He, Runpei Dong, Zixuan Chen, and Saurabh Gupta.
\newblock Learning getting-up policies for real-world humanoid robots.
\newblock {\em arXiv preprint arXiv:2502.12152}, 2025.

\bibitem{xue2025unified}
Yufei Xue, Wentao Dong, Minghuan Liu, Weinan Zhang, and Jiangmiao Pang.
\newblock A unified and general humanoid whole-body controller for versatile locomotion.
\newblock {\em arXiv preprint arXiv:2502.03206}, 2025.

\bibitem{ben2025homie}
Qingwei Ben, Feiyu Jia, Jia Zeng, Junting Dong, Dahua Lin, and Jiangmiao Pang.
\newblock Homie: Humanoid loco-manipulation with isomorphic exoskeleton cockpit.
\newblock {\em arXiv preprint arXiv:2502.13013}, 2025.

\bibitem{he2025asap}
Tairan He, Jiawei Gao, Wenli Xiao, Yuanhang Zhang, Zi~Wang, Jiashun Wang, Zhengyi Luo, Guanqi He, Nikhil Sobanbab, Chaoyi Pan, et~al.
\newblock Asap: Aligning simulation and real-world physics for learning agile humanoid whole-body skills.
\newblock {\em arXiv preprint arXiv:2502.01143}, 2025.

\bibitem{zeng2025behavior}
Weishuai Zeng, Shunlin Lu, Kangning Yin, Xiaojie Niu, Minyue Dai, Jingbo Wang, and Jiangmiao Pang.
\newblock Behavior foundation model for humanoid robots.
\newblock {\em arXiv preprint arXiv:2509.13780}, 2025.

\bibitem{tessler2024maskedmimic}
Chen Tessler, Yunrong Guo, Ofir Nabati, Gal Chechik, and Xue~Bin Peng.
\newblock Maskedmimic: Unified physics-based character control through masked motion inpainting.
\newblock {\em ACM Transactions On Graphics (TOG)}, 43(6):1--21, 2024.

\bibitem{he2025hover}
Tairan He, Wenli Xiao, Toru Lin, Zhengyi Luo, Zhenjia Xu, Zhenyu Jiang, Jan Kautz, Changliu Liu, Guanya Shi, Xiaolong Wang, et~al.
\newblock Hover: Versatile neural whole-body controller for humanoid robots.
\newblock In {\em 2025 IEEE International Conference on Robotics and Automation (ICRA)}, pages 9989--9996. IEEE, 2025.

\bibitem{liao2025beyondmimic}
Qiayuan Liao, Takara~E Truong, Xiaoyu Huang, Yuman Gao, Guy Tevet, Koushil Sreenath, and C~Karen Liu.
\newblock Beyondmimic: From motion tracking to versatile humanoid control via guided diffusion.
\newblock {\em arXiv preprint arXiv:2508.08241}, 2025.

\bibitem{li2025bfm}
Yitang Li, Zhengyi Luo, Tonghe Zhang, Cunxi Dai, Anssi Kanervisto, Andrea Tirinzoni, Haoyang Weng, Kris Kitani, Mateusz Guzek, Ahmed Touati, et~al.
\newblock Bfm-zero: A promptable behavioral foundation model for humanoid control using unsupervised reinforcement learning.
\newblock {\em arXiv preprint arXiv:2511.04131}, 2025.

\bibitem{luo2025sonic}
Zhengyi Luo, Ye~Yuan, Tingwu Wang, Chenran Li, Sirui Chen, Fernando Castaneda, Zi-Ang Cao, Jiefeng Li, David Minor, Qingwei Ben, et~al.
\newblock Sonic: Supersizing motion tracking for natural humanoid whole-body control.
\newblock {\em arXiv preprint arXiv:2511.07820}, 2025.

\bibitem{kaplan2020scaling}
Jared Kaplan, Sam McCandlish, Tom Henighan, Tom~B Brown, Benjamin Chess, Rewon Child, Scott Gray, Alec Radford, Jeffrey Wu, and Dario Amodei.
\newblock Scaling laws for neural language models.
\newblock {\em arXiv preprint arXiv:2001.08361}, 2020.

\bibitem{wei2022emergent}
Jason Wei, Yi~Tay, Rishi Bommasani, Colin Raffel, Barret Zoph, Sebastian Borgeaud, Dani Yogatama, Maarten Bosma, Denny Zhou, Donald Metzler, et~al.
\newblock Emergent abilities of large language models.
\newblock {\em arXiv preprint arXiv:2206.07682}, 2022.

\bibitem{blattmann2023stable}
Andreas Blattmann, Tim Dockhorn, Sumith Kulal, Daniel Mendelevitch, Maciej Kilian, Dominik Lorenz, Yam Levi, Zion English, Vikram Voleti, Adam Letts, et~al.
\newblock Stable video diffusion: Scaling latent video diffusion models to large datasets.
\newblock {\em arXiv preprint arXiv:2311.15127}, 2023.

\bibitem{peng2018deepmimic}
Xue~Bin Peng, Pieter Abbeel, Sergey Levine, and Michiel Van~de Panne.
\newblock Deepmimic: Example-guided deep reinforcement learning of physics-based character skills.
\newblock {\em ACM Transactions On Graphics (TOG)}, 37(4):1--14, 2018.

\bibitem{chen2026holomotion}
Maiyue Chen, Kaihui Wang, Bo~Zhang, Xihan Ma, Zhiyuan Yang, Yi~Ren, Qijun Huang, Zihao Zhu, Yucheng Wang, and Zhizhong Su.
\newblock Holomotion-1 technical report.
\newblock {\em arXiv preprint arXiv:2605.15336}, 2026.

\bibitem{schulman2017proximal}
John Schulman, Filip Wolski, Prafulla Dhariwal, Alec Radford, and Oleg Klimov.
\newblock Proximal policy optimization algorithms.
\newblock {\em arXiv preprint arXiv:1707.06347}, 2017.

\bibitem{seo2025fasttd3}
Younggyo Seo, Carmelo Sferrazza, Haoran Geng, Michal Nauman, Zhao-Heng Yin, and Pieter Abbeel.
\newblock Fasttd3: Simple, fast, and capable reinforcement learning for humanoid control.
\newblock {\em arXiv preprint arXiv:2505.22642}, 2025.

\bibitem{yi2026flow}
Brent Yi, Hongsuk Choi, Himanshu~Gaurav Singh, Xiaoyu Huang, Takara~E Truong, Carmelo Sferrazza, Yi~Ma, Rocky Duan, Pieter Abbeel, Guanya Shi, et~al.
\newblock Flow policy gradients for robot control.
\newblock {\em arXiv preprint arXiv:2602.02481}, 2026.

\bibitem{chen2025gmt}
Zixuan Chen, Mazeyu Ji, Xuxin Cheng, Xuanbin Peng, Xue~Bin Peng, and Xiaolong Wang.
\newblock Gmt: General motion tracking for humanoid whole-body control.
\newblock {\em arXiv preprint arXiv:2506.14770}, 2025.

\bibitem{yin2026unitracker}
Kangning Yin, Weishuai Zeng, Ke~Fan, Minyue Dai, Zirui Wang, Qiang Zhang, Zheng Tian, Jingbo Wang, Jiangmiao Pang, and Weinan Zhang.
\newblock Unitracker: Learning universal whole-body motion tracker for humanoid robots.
\newblock {\em IEEE Robotics and Automation Letters}, 2026.

\bibitem{wang2026omnixtreme}
Yunshen Wang, Shaohang Zhu, Peiyuan Zhi, Yuhan Li, Jiaxin Li, Yong-Lu Li, Yuchen Xiao, Xingxing Wang, Baoxiong Jia, and Siyuan Huang.
\newblock Omnixtreme: Breaking the generality barrier in high-dynamic humanoid control.
\newblock {\em arXiv preprint arXiv:2602.23843}, 2026.

\bibitem{li2025clone}
Yixuan Li, Yutang Lin, Jieming Cui, Tengyu Liu, Wei Liang, Yixin Zhu, and Siyuan Huang.
\newblock Clone: Closed-loop whole-body humanoid teleoperation for long-horizon tasks.
\newblock In {\em 9th Annual Conference on Robot Learning}, 2025.

\bibitem{ze2025twist}
Yanjie Ze, Zixuan Chen, Joao~Pedro Ara{\'u}jo, Zi-ang Cao, Xue~Bin Peng, Jiajun Wu, and C~Karen Liu.
\newblock Twist: Teleoperated whole-body imitation system.
\newblock {\em arXiv preprint arXiv:2505.02833}, 2025.

\bibitem{ze2025twist2}
Yanjie Ze, Siheng Zhao, Weizhuo Wang, Angjoo Kanazawa, Rocky Duan, Pieter Abbeel, Guanya Shi, Jiajun Wu, and C~Karen Liu.
\newblock Twist2: Scalable, portable, and holistic humanoid data collection system.
\newblock {\em arXiv preprint arXiv:2511.02832}, 2025.

\bibitem{wang2025physhsi}
Huayi Wang, Wentao Zhang, Runyi Yu, Tao Huang, Junli Ren, Feiyu Jia, Zirui Wang, Xiaojie Niu, Xiao Chen, Jiahe Chen, et~al.
\newblock Physhsi: Towards a real-world generalizable and natural humanoid-scene interaction system.
\newblock {\em arXiv preprint arXiv:2510.11072}, 2025.

\bibitem{yang2025omniretarget}
Lujie Yang, Xiaoyu Huang, Zhen Wu, Angjoo Kanazawa, Pieter Abbeel, Carmelo Sferrazza, C~Karen Liu, Rocky Duan, and Guanya Shi.
\newblock Omniretarget: Interaction-preserving data generation for humanoid whole-body loco-manipulation and scene interaction.
\newblock {\em arXiv preprint arXiv:2509.26633}, 2025.

\bibitem{zhuang2026deep}
Ziwen Zhuang, Shaoting Zhu, Mengjie Zhao, and Hang Zhao.
\newblock Deep whole-body parkour.
\newblock {\em arXiv preprint arXiv:2601.07701}, 2026.

\bibitem{weng2025hdmi}
Haoyang Weng, Yitang Li, Nikhil Sobanbabu, Zihan Wang, Zhengyi Luo, Tairan He, Deva Ramanan, and Guanya Shi.
\newblock Hdmi: Learning interactive humanoid whole-body control from human videos.
\newblock {\em arXiv preprint arXiv:2509.16757}, 2025.

\bibitem{wang2026humanx}
Yinhuai Wang, Qihan Zhao, Yuen~Fui Lau, Runyi Yu, Hok~Wai Tsui, Qifeng Chen, Jingbo Wang, Jiangmiao Pang, and Ping Tan.
\newblock Humanx: Toward agile and generalizable humanoid interaction skills from human videos.
\newblock {\em arXiv preprint arXiv:2602.02473}, 2026.

\bibitem{he2026ultra}
Xialin He, Sirui Xu, Xinyao Li, Runpei Dong, Liuyu Bian, Yu-Xiong Wang, and Liang-Yan Gui.
\newblock Ultra: Unified multimodal control for autonomous humanoid whole-body loco-manipulation.
\newblock {\em arXiv preprint arXiv:2603.03279}, 2026.

\bibitem{jiang2025uniact}
Nan Jiang, Zimo He, Wanhe Yu, Lexi Pang, Yunhao Li, Hongjie Li, Jieming Cui, Yuhan Li, Yizhou Wang, Yixin Zhu, et~al.
\newblock Uniact: Unified motion generation and action streaming for humanoid robots.
\newblock {\em arXiv preprint arXiv:2512.24321}, 2025.

\bibitem{xie2026textop}
Weiji Xie, Jiakun Zheng, Jinrui Han, Jiyuan Shi, Weinan Zhang, Chenjia Bai, and Xuelong Li.
\newblock Textop: Real-time interactive text-driven humanoid robot motion generation and control.
\newblock {\em arXiv preprint arXiv:2602.07439}, 2026.

\bibitem{tao2026heracles}
Zelin Tao, Zeran Su, Peiran Liu, Jingkai Sun, Wenqiang Que, Jiahao Ma, Jialin Yu, Jiahang Cao, Pihai Sun, Hao Liang, et~al.
\newblock Heracles: Bridging precise tracking and generative synthesis for general humanoid control.
\newblock {\em arXiv preprint arXiv:2603.27756}, 2026.

\bibitem{yin2026robostriker}
Kangning Yin, Zhe Cao, Wentao Dong, Weishuai Zeng, Tianyi Zhang, Qiang Zhang, Jingbo Wang, Jiangmiao Pang, Ming Zhou, and Weinan Zhang.
\newblock Robostriker: Hierarchical decision-making for autonomous humanoid boxing.
\newblock {\em arXiv preprint arXiv:2601.22517}, 2026.

\bibitem{zhang2026learning}
Zhikai Zhang, Haofei Lu, Yunrui Lian, Ziqing Chen, Yun Liu, Chenghuai Lin, Han Xue, Zicheng Zeng, Zekun Qi, Shaolin Zheng, et~al.
\newblock Learning athletic humanoid tennis skills from imperfect human motion data.
\newblock {\em arXiv preprint arXiv:2603.12686}, 2026.

\bibitem{ross2011reduction}
St{\'e}phane Ross, Geoffrey Gordon, and Drew Bagnell.
\newblock A reduction of imitation learning and structured prediction to no-regret online learning.
\newblock In {\em Proceedings of the fourteenth international conference on artificial intelligence and statistics}, pages 627--635. JMLR Workshop and Conference Proceedings, 2011.

\bibitem{dhariwal2021diffusion}
Prafulla Dhariwal and Alexander Nichol.
\newblock Diffusion models beat gans on image synthesis.
\newblock {\em Advances in neural information processing systems}, 34:8780--8794, 2021.

\bibitem{huang2025diffuse}
Xiaoyu Huang, Takara Truong, Yunbo Zhang, Fangzhou Yu, Jean~Pierre Sleiman, Jessica Hodgins, Koushil Sreenath, and Farbod Farshidian.
\newblock Diffuse-cloc: Guided diffusion for physics-based character look-ahead control.
\newblock {\em ACM Transactions on Graphics (TOG)}, 44(4):1--12, 2025.

\bibitem{touati2021learning}
Ahmed Touati and Yann Ollivier.
\newblock Learning one representation to optimize all rewards.
\newblock {\em Advances in Neural Information Processing Systems}, 34:13--23, 2021.

\bibitem{tirinzoni2025zero}
Andrea Tirinzoni, Ahmed Touati, Jesse Farebrother, Mateusz Guzek, Anssi Kanervisto, Yingchen Xu, Alessandro Lazaric, and Matteo Pirotta.
\newblock Zero-shot whole-body humanoid control via behavioral foundation models.
\newblock {\em arXiv preprint arXiv:2504.11054}, 2025.

\bibitem{harvey2020robust}
F{\'e}lix~G Harvey, Mike Yurick, Derek Nowrouzezahrai, and Christopher Pal.
\newblock Robust motion in-betweening.
\newblock {\em ACM Transactions on Graphics (TOG)}, 39(4):60--1, 2020.

\bibitem{mahmood2019amass}
Naureen Mahmood, Nima Ghorbani, Nikolaus~F Troje, Gerard Pons-Moll, and Michael~J Black.
\newblock Amass: Archive of motion capture as surface shapes.
\newblock In {\em Proceedings of the IEEE/CVF international conference on computer vision}, pages 5442--5451, 2019.

\bibitem{li2023object}
Jiaman Li, Jiajun Wu, and C~Karen Liu.
\newblock Object motion guided human motion synthesis.
\newblock {\em ACM Transactions on Graphics (TOG)}, 42(6):1--11, 2023.

\bibitem{taheri2020grab}
Omid Taheri, Nima Ghorbani, Michael~J Black, and Dimitrios Tzionas.
\newblock Grab: A dataset of whole-body human grasping of objects.
\newblock In {\em European conference on computer vision}, pages 581--600. Springer, 2020.

\bibitem{guo2025snapmogen}
Chuan Guo, Inwoo Hwang, Jian Wang, and Bing Zhou.
\newblock Snapmogen: Human motion generation from expressive texts.
\newblock {\em arXiv preprint arXiv:2507.09122}, 2025.

\bibitem{li2023finedance}
Ronghui Li, Junfan Zhao, Yachao Zhang, Mingyang Su, Zeping Ren, Han Zhang, Yansong Tang, and Xiu Li.
\newblock Finedance: A fine-grained choreography dataset for 3d full body dance generation.
\newblock In {\em Proceedings of the IEEE/CVF International Conference on Computer Vision}, pages 10234--10243, 2023.

\bibitem{bones_seed_2026}
{Bones Studio}.
\newblock Bones-seed: Skeletal everyday embodiment dataset.
\newblock \url{https://huggingface.co/datasets/bones-studio/seed}, 2026.

\bibitem{mclean2025embody}
Claire McLean, Makenzie Meendering, Tristan Swartz, Orri Gabbay, Alexandra Olsen, Rachel Jacobs, Nicholas Rosen, Philippe de~Bree, Tony Garcia, Gadsden Merrill, et~al.
\newblock Embody 3d: A large-scale multimodal motion and behavior dataset.
\newblock {\em arXiv preprint arXiv:2510.16258}, 2025.

\bibitem{araujo2025retargeting}
Joao~Pedro Araujo, Yanjie Ze, Pei Xu, Jiajun Wu, and C~Karen Liu.
\newblock Retargeting matters: General motion retargeting for humanoid motion tracking.
\newblock {\em arXiv preprint arXiv:2510.02252}, 2025.

\bibitem{ProtoMotions}
Chen Tessler*, Yifeng Jiang*, Xue~Bin Peng, Erwin Coumans, Yi~Shi, Haotian Zhang, Davis Rempe, Gal Chechik†, and Sanja Fidler†.
\newblock Protomotions3: An open-source framework for humanoid simulation and control.
\newblock \url{https://github.com/NVLabs/ProtoMotions/}, 2025.

\bibitem{zhang2019root}
Biao Zhang and Rico Sennrich.
\newblock Root mean square layer normalization.
\newblock {\em Advances in neural information processing systems}, 32, 2019.

\bibitem{yue2026image}
Kaiyu Yue, Menglin Jia, Ji~Hou, and Tom Goldstein.
\newblock Image generation with a sphere encoder.
\newblock {\em arXiv preprint arXiv:2602.15030}, 2026.

\bibitem{unitreeg1}
Unitree.
\newblock Unitree g1 humanoid agent ai avatar, 2024.

\bibitem{mittal2025isaac}
Mayank Mittal, Pascal Roth, James Tigue, Antoine Richard, Octi Zhang, Peter Du, Antonio Serrano-Munoz, Xinjie Yao, Ren{\'e} Zurbr{\"u}gg, Nikita Rudin, et~al.
\newblock Isaac lab: A gpu-accelerated simulation framework for multi-modal robot learning.
\newblock {\em arXiv preprint arXiv:2511.04831}, 2025.

\bibitem{todorov2012mujoco}
Emanuel Todorov, Tom Erez, and Yuval Tassa.
\newblock Mujoco: A physics engine for model-based control.
\newblock In {\em 2012 IEEE/RSJ International Conference on Intelligent Robots and Systems}, pages 5026--5033. IEEE, 2012.

\bibitem{schepers2018xsens}
Martin Schepers, Matteo Giuberti, Giovanni Bellusci, et~al.
\newblock Xsens mvn: Consistent tracking of human motion using inertial sensing.
\newblock {\em Xsens Technol}, 1(8):1--8, 2018.

\bibitem{mason2022local}
Ian Mason, Sebastian Starke, and Taku Komura.
\newblock Real-time style modelling of human locomotion via feature-wise transformations and local motion phases.
\newblock {\em Proceedings of the ACM on Computer Graphics and Interactive Techniques}, 5(1), may 2022.

\bibitem{shazeer2020glu}
Noam Shazeer.
\newblock Glu variants improve transformer.
\newblock {\em arXiv preprint arXiv:2002.05202}, 2020.

\bibitem{LCM}
Albert~S. Huang, Edwin Olson, and David~C. Moore.
\newblock Lcm: Lightweight communications and marshalling.
\newblock In {\em 2010 IEEE/RSJ International Conference on Intelligent Robots and Systems}, pages 4057--4062, 2010.

\bibitem{HTC2024VIVEUltimateTracker}
{HTC VIVE}.
\newblock Vive ultimate tracker -- full-body tracking for standalone vr, 2024.
\newblock Accessed: 2026-01-18.

\end{thebibliography}
\bibliographystyle{unsrt}
}
\newpage

% \tableofcontents
\newpage
\appendix

\section{Method Details}
\vspace{-3mm}
\subsection{Control Modes}
\vspace{-3mm}
\label{app:control_mode}

\begin{table}[H]
\centering
\renewcommand{\arraystretch}{1.0}
\begin{tabular}{c c c}

\hline
\textbf{Control Mode}  & \textbf{\# Links} &  \textbf{Activated Links}                              \\ 
\hline
Root Mode & 1 & \small{Pelvis} \\
\hline
Bimanual Mode & 2 & \small{Left Hand, Right Hand} \\
\hline
Root-and-Hand Mode & 3 & \makecell{\small{Pelvis, Left Hand, Right Hand}} \\
\hline
End-Effector Mode & 4 & \makecell{\small{Left Hand, Right Hand, Left Foot, Right Foot}} \\
\hline
Root-and-End-Effector Mode & 5 & \makecell{\small{Pelvis, Left Hand, Right Hand, Left Foot, Right Foot}} \\
\hline
Upper-Body Mode & 6 & \small{\makecell{Left Shoulder, Right Shoulder, Left Elbow \\ Right Elbow, Left Hand, Right Hand}} \\
\hline
Root-and-Upper-Body Mode & 7 & \small{\makecell{Pelvis, Left Shoulder, Right Shoulder\\Left Elbow, Right Elbow, Left Hand, Right Hand}} \\
\hline
Whole-Body Mode & 14 & \small{\makecell{Pelvis, Left Shoulder, Right Shoulder, Left Elbow \\ Right Elbow, Left Hand, Right Hand, Torso,  Left Hip\\ Right Hip, Left Knee, Right Knee, Left Foot, Right Foot}} \\
\hline
\end{tabular}
% \vspace{3mm}
\caption{\textbf{Curated control modes featuring different activated target links.} Instead of constructing arbitrary link-wise masks by independently activating each link according to a Bernoulli distribution with probability $0.5$~\cite{zeng2025behavior}, we curate a set of eight representative control modes spanning different levels of specification granularity. By randomly sampling a control mode at each environment reset, the same underlying behavior can be represented by diverse forms of behavioral specifications. Such design enables versatile humanoid control across multiple control modes while naturally supporting behavior inpainting under sparse control signals. }
\label{tab:appendix_control_mode}
\end{table}

\vspace{-3mm}
\subsection{Reward Design}
\vspace{-3mm}
\label{app:reward_design}
\begin{table}[H]
\centering
\resizebox{0.9\linewidth}{!}{
\renewcommand{\arraystretch}{1.1}

\begin{tabular}{  c  c  c  c }
\hline
\textbf{Term} & \textbf{Expression} & \textbf{Weight} & \textbf{Notes}    \\ 
\hline
    & Tracking Reward &  & \\
\hline
Global Root Height & $\exp (-\lVert {h}_t-{h}_t^{\text{ref}}\rVert_2^2 / \sigma^2_\text{height})$ & 0.5 & $\sigma_{\text{height}}=0.3$ \\
Global Body Position & $\exp (-\frac1{\mid {B}\mid}\sum_{b\in {B}}\lVert {p}_{t,b}-{p}_{t,b}^{\text{ref}}\rVert_2^2 / \sigma^2_\text{pos})$ & 1.0 & $\sigma_{\text{pos}}=0.3$ \\
Global Body Rotation & $\exp (-\frac{1}{\mid {B}\mid}\sum_{b\in {B}}\lVert {\theta}_{t,b}\boxminus{\theta}_{t,b}^{\text{ref}}\rVert_2^2 / \sigma^2_\text{rot})$ & 1.0 & $\sigma_{\text{rot}}=0.4$ \\
Global Body Linear Velocity & $\exp (-\frac{1}{\mid {B}\mid}\sum_{b\in {B}}\lVert {v}_{t,b}-{v}_{t,b}^{\text{ref}}\rVert_2^2 / \sigma^2_\text{lin})$ & 1.0 & $\sigma_{\text{lin}}=1.0$ \\
Global Body Angular Velocity & $\exp (-\frac{1}{\mid {B}\mid}\sum_{b\in {B}}\lVert {\omega}_{t,b}-{\omega}_{t,b}^{\text{ref}}\rVert_2^2 / \sigma^2_\text{ang})$ & 1.0 & $\sigma_{\text{ang}}=3.14$ \\
\hline
 & Penalty Term &  & \\
\hline
Action Rate & $ \lVert \boldsymbol{a}_t - \boldsymbol{a}_{t-1} \rVert_2^2  $ & -0.1 & \\
Joint Position Limits      &   $\sum_{j\in {J}}\boldsymbol{1}\left[q_{t,j}\notin \left[q_{j}^{\text{min}},q_{j}^{\text{max}}\right]\right]$      & $-10.0$ &   \\
\hline
& Survival Reward& & \\
\hline
Survival & $\boldsymbol{1}\left[\text{is\_alive}\right]$ & 1.0 & \\
\hline
\end{tabular}
}
% \vspace{2mm}
\caption{\textbf{Reward components and weights.} Variables annotated with the superscript $\mathrm{ref}$ are derived from the reference motion, whereas those without this annotation correspond to the humanoid's current state. Reward weights are automatically scaled by the environment according to the simulation timestep $dt$.
}
\label{tab:appendix_reward}
\end{table}

\vspace{-3mm}
\subsection{Domain Randomization}
\vspace{-3mm}
\label{app:dr}
\begin{table}[H]
\centering
\footnotesize
\setlength{\tabcolsep}{6pt}
\begin{tabular}{l l}
\toprule
\textbf{Domain Randomization} & \textbf{Sampling Distribution} \\
\midrule
\multicolumn{2}{l}{\textit{Physical parameters}} \\
\quad Static friction coefficients & $\mu_s \sim {U}[0.3,1.6]$ \\
\quad Dynamic friction coefficients & $\mu_d \sim {U}[0.3,1.2]$ \\
\quad Default joint positions & ${q}_0 \sim {q}_0 + {U}[-0.01, 0.01]$ \\
\quad Base COM offset (x, y, z) & $\Delta x \sim {U}[-0.025,0.025],\ \Delta y \sim {U}[-0.05,0.05],\ \Delta z \sim {U}[-0.05,0.05]$ \\
\quad Hand mass & ${m}_0 \sim m_0 + {U}[0.0, 1.0]$ \\
\multicolumn{2}{l}{\textit{Root velocity perturbations}} \\
\quad Root linear vel (x, y, z) & $v_x \sim {U}[-0.5,0.5],\ v_y \sim {U}[-0.5,0.5],\ v_z \sim {U}[-0.2,0.2]$ \\
\quad Root angular vel & $\omega_{\text{roll}} \sim {U}[-0.52,0.52],\ \omega_{\text{pitch}} \sim {U}[-0.52,0.52],\ \omega_{\text{yaw}} \sim {U}[-0.78,0.78]$ \\
\quad Push duration & $\Delta t \sim {U}[1,3]\text{s}$ \\
\bottomrule
\end{tabular}
\caption{\textbf{Domain randomization parameters.} Randomization of physical parameters is applied at the start of training. Root velocity perturbations are introduced periodically during training. ${U}[\cdot]$: uniform distribution.}
\label{tab:rand_runtime_eq}
\end{table}

\vspace{-3mm}
\subsection{Model Architecture}
\vspace{-3mm}
\label{app:model}

We provide the implementation details of the proposed \emph{Humanoid Transformer} for BFM pretraining. 

Let $L$ denotes the history length. Proprioceptions and actions are first extended to temporal windows: 
\begin{equation}
    {H}_t \triangleq (\boldsymbol{s}_{t-L+1}^{\text{p}},\boldsymbol{a}_{t-L+1},\cdots, \boldsymbol{s}_{t-1}^{\text{p}},\boldsymbol{a}_{t-1},\boldsymbol{s}_t^{\text{p}}).
\end{equation}
Given the historical information, we encode proprioceptive states and actions with separate tokenizers:
\begin{equation}
    z_i^{\boldsymbol{s}^{\text{p}}}=E_{\boldsymbol{s}^{\text{p}}}(\boldsymbol{s}_i^{\text{p}})\quad z_i^{\boldsymbol{a}}=E_{\boldsymbol{a}}(\boldsymbol{a}_i).
\end{equation}
The historical context is then represented as a sequence of interleaved proprioceptive and action tokens:
\begin{equation}
    \left(z_{t-L+1}^{\boldsymbol{s}^{\text{p}}},z_{t-L+1}^{\boldsymbol{a}},\cdots,z_{t-1}^{\boldsymbol{s}^{\text{p}}},z_{t-1}^{\boldsymbol{a}},z_t^{\boldsymbol{s}^p}\right).
\end{equation}
We additionally append a learnable query token $e$ at the end and obtain the complete context:
\begin{equation}
    {Z}_t^{\text{p}}\triangleq\left(z_{t-L+1}^{\boldsymbol{s}^{\text{p}}},z_{t-L+1}^{\boldsymbol{a}},\cdots,z_{t-1}^{\boldsymbol{s}^{\text{p}}},z_{t-1}^{\boldsymbol{a}},z_t^{\boldsymbol{s}^p},e\right).
\end{equation}

Let $N$ denote the number of future frames. The goal states are also extended to a temporal window:
\begin{equation}
    {G}_t \triangleq  \left(\boldsymbol{s}_{t+\delta_1}^{\text{g}},\boldsymbol{s}_{t+\delta_2}^{\text{g}},\cdots, \boldsymbol{s}_{t+\delta_N}^{\text{g}}\right),
\end{equation}
where $\delta_i$ is the temporal offset of future frame $i$ relative to current frame and $\boldsymbol{s}_{t+\delta_i}^{\text{g}}$ is defined as:
\begin{equation}
    \begin{aligned}
        \boldsymbol{x}_{t+\delta_i} \triangleq (p_{t+\delta_i}^{\text{ref}}-&p_t^{\text{root}},p_{t+\delta_i}^{\text{ref}}-p_t,\theta_{t+\delta_i}^{\text{ref}}\ominus \theta_t^{\text{root}},\theta_{t+\delta_i}^{\text{ref}}\ominus \theta_t) \\
        \boldsymbol{s}_{t+\delta_i}^{\text{g}} &\triangleq (\boldsymbol{x}_{t+\delta_i}\odot \boldsymbol{m}_i, \boldsymbol{m}_i),\boldsymbol{m}_i\sim {M}
    \end{aligned}
\end{equation}
We concatenate goal states with the temporal offsets and encode them with a shared tokenizer:
\begin{equation}
z_{i}^{\boldsymbol{s}^{\text{g}}}=E_{\boldsymbol{s}^{\text{g}}}(\boldsymbol{s}_{t+\delta_i}^{\text{g}}).
\end{equation}
The future window then becomes a sequence of task tokens, each corresponding to a future frame:
\begin{equation}
    {Z}_{t}^{\text{g}}\triangleq (z_{1}^{\boldsymbol{s}^{\text{g}}},z_{2}^{\boldsymbol{s}^{\text{g}}},\cdots,z_{N}^{\boldsymbol{s}^{\text{g}}})
\end{equation}

Given the historical context ${Z}_t^{\text{p}}$ and future window ${Z}_t^{\text{g}}$, we employ self-attention to capture the temporal dependencies and dynamic evolution in historical context and cross-attention to condition the model on diverse behavioral specifications. This design decouples the modeling of the behavioral manifold from the injection of control conditions, naturally aligning with our objective of building BFMs.

Each layer of the transformer consists of the self-attention, the cross-attention, the feed-forward and the RMSNorm module. Let ${{Z}_{t,(i-1)}^{\text{p}}}$ denote the context after layer ${i-1}$. The updating rule in layer ${i}$ is:
\begin{equation}
    \begin{aligned}
        {Z}_{t,(i-1)}^{\text{p}}&={Z}_{t,(i-1)}^{\text{p}}+\text{SelfAttn}(\text{RoPE}(\text{RMSNorm}({Z}_{t,(i-1)}^{\text{p}}))) \\
        {Z}_{t,(i-1)}^{\text{p}}&={Z}_{t,(i-1)}^{\text{p}}+\text{CrossAttn}(\text{RMSNorm}({Z}_{t,(i-1)}^{\text{p}}),\text{RMSNorm}({Z}_t^{g}))\\
        {Z}_{t,(i)}^{\text{p}}&=Z_{t,(i-1)}^{\text{p}}+\text{FFN}(\text{RMSNorm}({Z}_{t,(i-1)}^{\text{p}}))
    \end{aligned} 
\end{equation}
where the learnable query token is masked from being attended to by the other context tokens while it can attend to all context tokens to aggregate historical information in the self-attention module.

The RMSNorm module projects goal embeddings onto a continuous and bounded hypersphere via:
\begin{equation}
    \mathrm{RMSNorm}(\boldsymbol{x})
    =
    \boldsymbol{\gamma}
    \cdot
    \frac{\boldsymbol{x}}
    {
        \sqrt{\frac{1}{d}\sum_{i=1}^{d} x_i^2}
    },
\end{equation}
where $\boldsymbol{\gamma}$ is a learnable scalable parameter. Throughout this paper, we mainly focus on the unit hypersphere, regardless of any multiplicative scaling coefficient. We do not employ auxiliary objectives to regularize the latent space; instead, it is shaped solely through the motion-tracking objective.

Given the output of the backbone, we denote the latent embedding corresponding to the learnable query token as $\hat{e}$ and use a projection head $f$ to predict the action or value, which is expressed as $f\!\left(\hat{e}\right)$.

In practice, we implement the proprioception tokenizer $E_{\boldsymbol{s}^{\text{p}}}$, action tokenizer $E_{\boldsymbol{a}}$, goal tokenizer $E_{\boldsymbol{s}^{\text{g}}}$, projection head $f$ all as single linear layers, and the feed-forward module as SwiGLU~\cite{shazeer2020glu}.

\section{Experiment Details}
\vspace{-3mm}
\subsection{Real-World Deployment}
\vspace{-3mm}
\label{app:real_robot}

All real-world experiments are conducted on the Unitree G1 humanoid robot with 29 degrees of freedom. We decouple high-level BFM inference from the low-level joint-space PD controller and employ LCM~\cite{LCM} for inter-process communication. This architecture provides direct access to the low-level control loop, thereby facilitating the implementation of safety mechanisms, such as binding emergency termination to a dedicated button on the remote controller for direct human interference.

The low-level controller is implemented upon Unitree SDK2 and operates at 200~Hz, whereas high-level BFM inference is accelerated with TensorRT and runs at 50~Hz. Both processes can be executed either onboard with the assistance of the Jetson Orin GPU, or on a local PC connected to the robot via a network cable. For scenarios involving online streaming of control signals, we transmit data over Wi-Fi using TCP and buffer the received signals with their timestamps. During inference, the index of the final future reference frame is dynamically adjusted according to the recorded timestamps, providing an effective mechanism for compensating communication and system latency during real-time deployment.

Although BFMs are pretrained with objectives to reproduce reference motions as integrated whole-body trajectories in the global frame, they naturally support both global and local control during real-world deployment. In the global control mode, we adopt HTC VIVE Ultimate Trackers~\cite{HTC2024VIVEUltimateTracker} for root localization. The root position on the $xy$-plane and the heading direction of the control signals are aligned with those of the robot through first-frame calibration. In the local control mode, the current root position specified by the control signal is treated as the robot's actual root position. While the reference heading could likewise be regarded as the robot's current heading, we instead preserve the heading derived from the IMU and align the direction of control signals through the same first-frame calibration procedure. This design avoids modifying the robot's raw sensory inputs while still enabling accurate tracking of the desired root orientation. The overall procedure can be interpreted as a reference-forcing mechanism: it assumes that the robot has already reached the desired root position and therefore only needs to track the remaining root orientation and whole-body pose, a strategy commonly adopted in prior work.

Since BFMs support eight curated control modes, any of them can be activated and switched online during real-world deployment. Notably, some control modes do not provide an explicit specification of the root target state; consequently, local control is not directly applicable under these modes.

\vspace{-3mm}
\subsection{Evaluation Protocol}
\vspace{-2mm}
\label{app:evaluation_metrics}

We report the success rate together with global and local tracking errors across diverse control modes. A motion sequence is considered unsuccessful if any selected link deviates from its desired position by more than $0.5\text{m}$ in the global frame. G-MPKPE and G-MPKRE denote the mean per-keypoint position and rotation errors of the selected links in the global frame, whereas L-MPKPE and L-MPKRE measure the corresponding root-relative errors after removing horizontal translation offsets and heading differences. Position errors (G-MPKPE and L-MPKPE) are reported in meters, whereas rotation errors (G-MPKRE and L-MPKRE) are reported in radians. The metrics are defined below, where $T$ is the number of frames and ${B}$ the selected link set.
\begin{align}
   \text{G-MPKPE} &= \frac{1}{T\cdot |{B}|}\sum_{t=1}^T\sum_{b\in {B}}||p_{t,b}-p_{t,b}^{\text{ref}}||_2  \\
   \text{G-MPKRE} &= \frac{1}{T\cdot |{B}|}\sum_{t=1}^T\sum_{b\in {B}}||\theta_{t,b}\boxminus \theta_{t,b}^{\text{ref}}||_2 \\
   \text{L-MPKPE}&=\frac{1}{T\cdot |{B}|}\sum_{t=1}^T\sum_{b\in {B}}||p_{t,b}^{\text{root-relative}}-p_{t,b}^{\text{root-relative,ref}}||_2 \\
   \text{L-MPKRE} &= \frac{1}{T\cdot |{B}|}\sum_{t=1}^T\sum_{b\in {B}}||\theta_{t,b}^{\text{root-relative}}\boxminus \theta_{t,b}^{\text{root-relative,ref}}||_2
\end{align}

\vspace{-3mm}
\subsection{Training Hyperparameters}
\vspace{-3mm}
\label{app:training_hyperparams}

\begin{table}[H]
\renewcommand{\arraystretch}{1.25}
    \centering
    \begin{tabular}{lc}
        \toprule 
        \textbf{Term} & \textbf{Value} \\
        \midrule 
        Optimizer & Adam \\
        Value coefficient & 1.0 \\
        PPO Clipping & 0.2 \\
        Entropy coefficient & 5.0e-3 \\
        Learning epochs & 2 \\
        Actor learning rate & 2.0e-5 \\
        Critic learning rate & 1.0e-3 \\
        Discount factor $\gamma$ & 0.99 \\ 
        GAE $\lambda$ & 0.95  \\ 
        Desired KL value & 0.01 \\
        Maximum gradient norm & 1.0 \\
        Training environments per GPU & 8192 \\
        Initial noise std & 0.8 \\
        GPU Type & NVIDIA RTX 4090 \\
        \bottomrule
    \end{tabular}
    \caption{\textbf{Hyperparameter Details.} These terms remain fixed across experiments to ensure a fair comparison.}
    \label{tab:hyperparams}
\end{table}

\vspace{-3mm}
\subsection{Scaling On-Policy Data Collection}
\vspace{-3mm}
\label{app:scaling_on_policy_data_collection}

We scale on-policy data collection along two complementary dimensions: the \emph{width} and the \emph{depth} of environmental interaction. With the number of environments per GPU held constant, the \emph{width} is increased by scaling the number of GPUs, while the \emph{depth} is increased by extending the rollout horizon. 

To ensure a fair comparison, we keep all the training hyperparameters fixed across experiments, as detailed in table~\ref{tab:hyperparams}. In particular, the number of mini-batches is adjusted according to the rollout horizon so that only the total amount of collected on-policy data varies across experiments, while the mini-batch size remains constant. We further adopt the reference motion dataset of medium size, which consists solely of the BONES-SEED dataset excluding the held-out test split. This removes potential benefits arising from heterogeneous data sources in the full reference motion dataset, allowing us to isolate the impact of scaling on-policy data collection. Finally, all experiments employ the same Humanoid Transformer model architecture of medium size, which provides a favorable balance between control performance and real-time onboard inference efficiency. Collectively, these design choices ensure that the observed differences can be primarily attributed to the scale of on-policy data collection.

\begin{figure}[h!]
    \centering
    \includegraphics[width=0.95\linewidth]{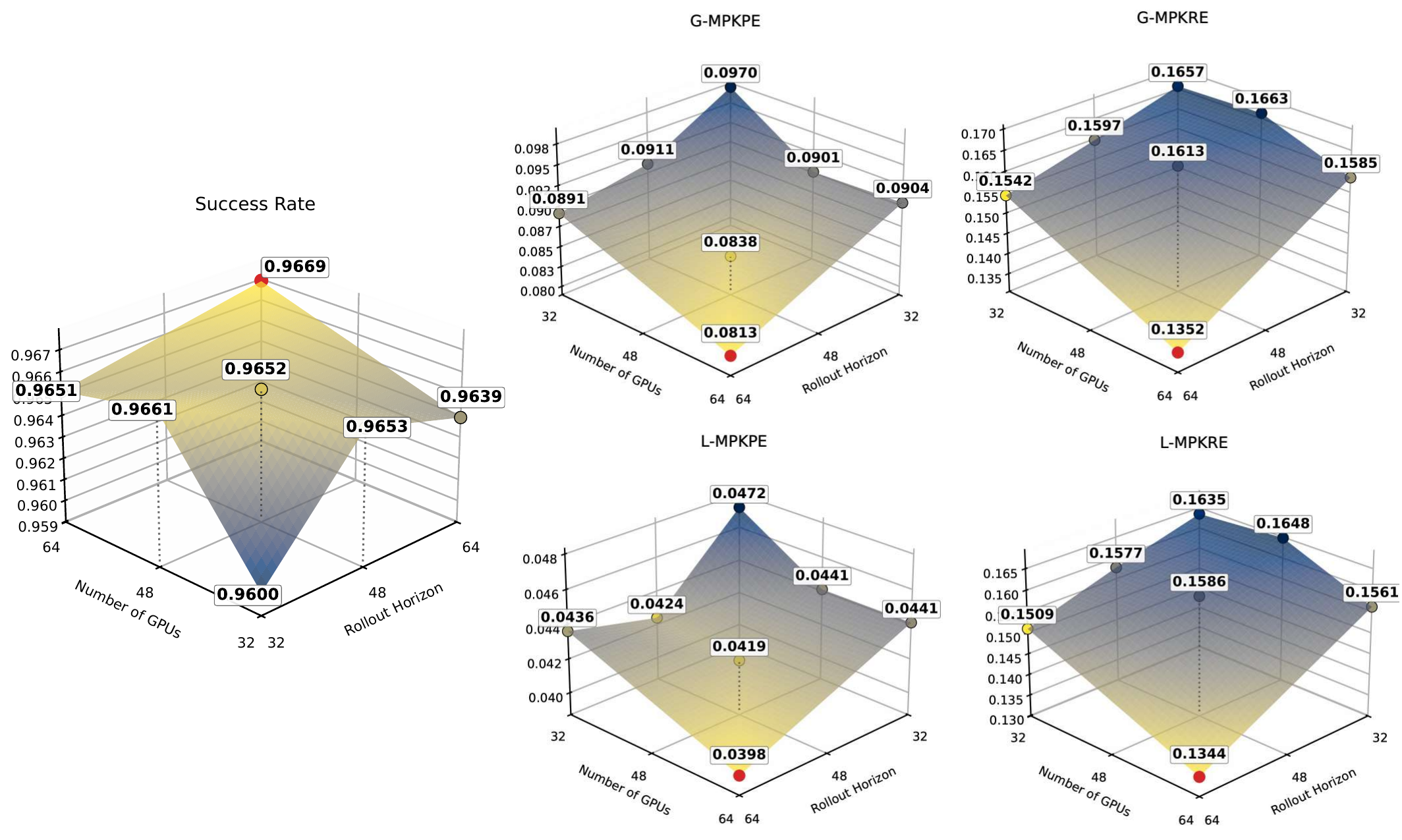}
    \caption{Results of scaling \textbf{on-policy data collection} on the \textbf{BONES Test Set} under \textbf{root-and-hand control mode}. Best results with the highest success rate and lowest tracking errors are highlighted with red circles.}
    \label{fig:scale_on_policy_bones_three_point}
    \vspace{-4mm}
\end{figure}

\begin{figure}[h!]
    \centering
    \includegraphics[width=0.95\linewidth]{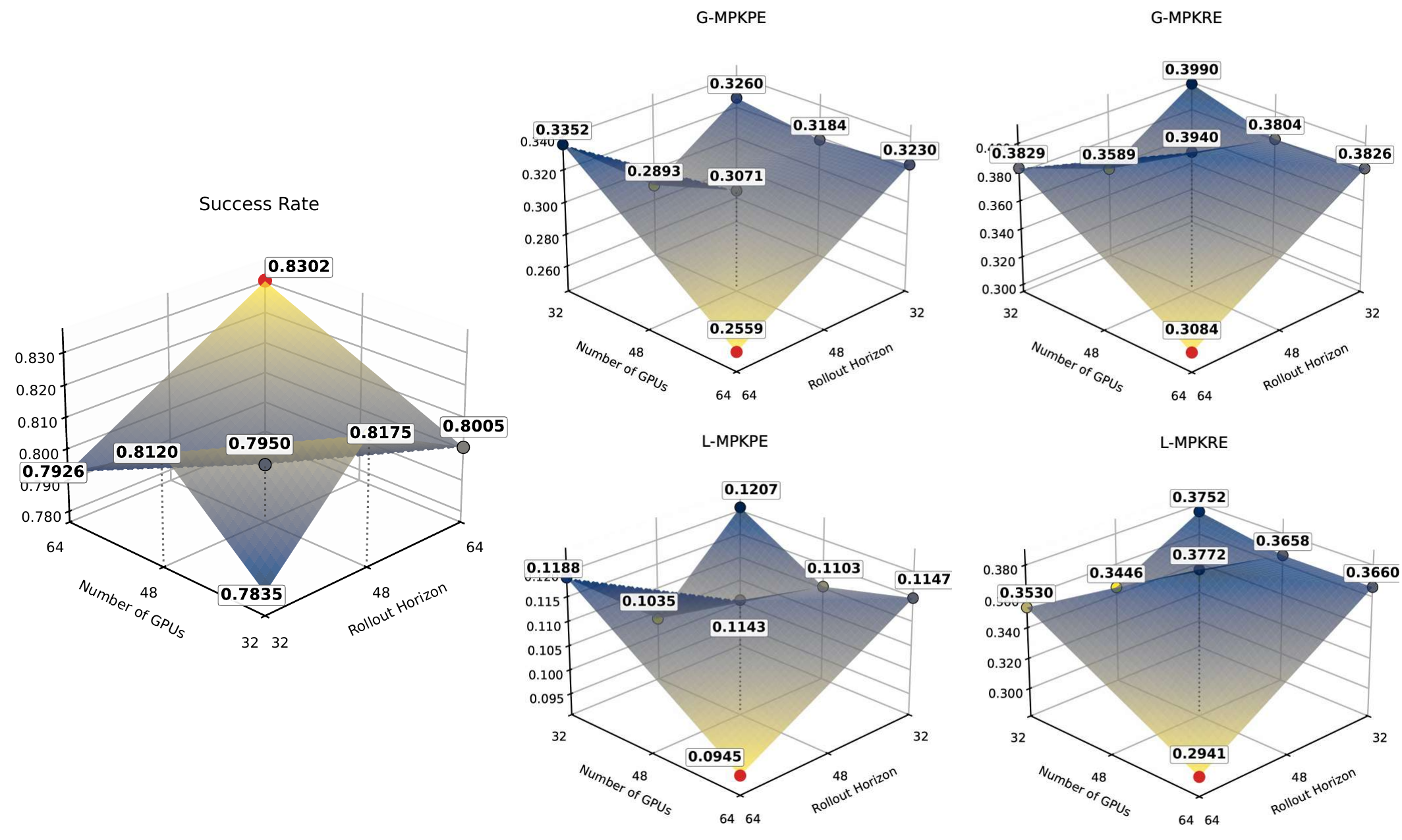}
    \caption{Results of scaling \textbf{on-policy data collection} on the \textbf{Ours Test Set} under \textbf{root-and-hand control mode}. Best results with the highest success rate and lowest tracking errors are highlighted with red circles.}
    \label{fig:scale_on_policy_ours_three_point}
    \vspace{-4mm}
\end{figure}

\begin{figure}[h!]
    \centering
    \includegraphics[width=0.95\linewidth]{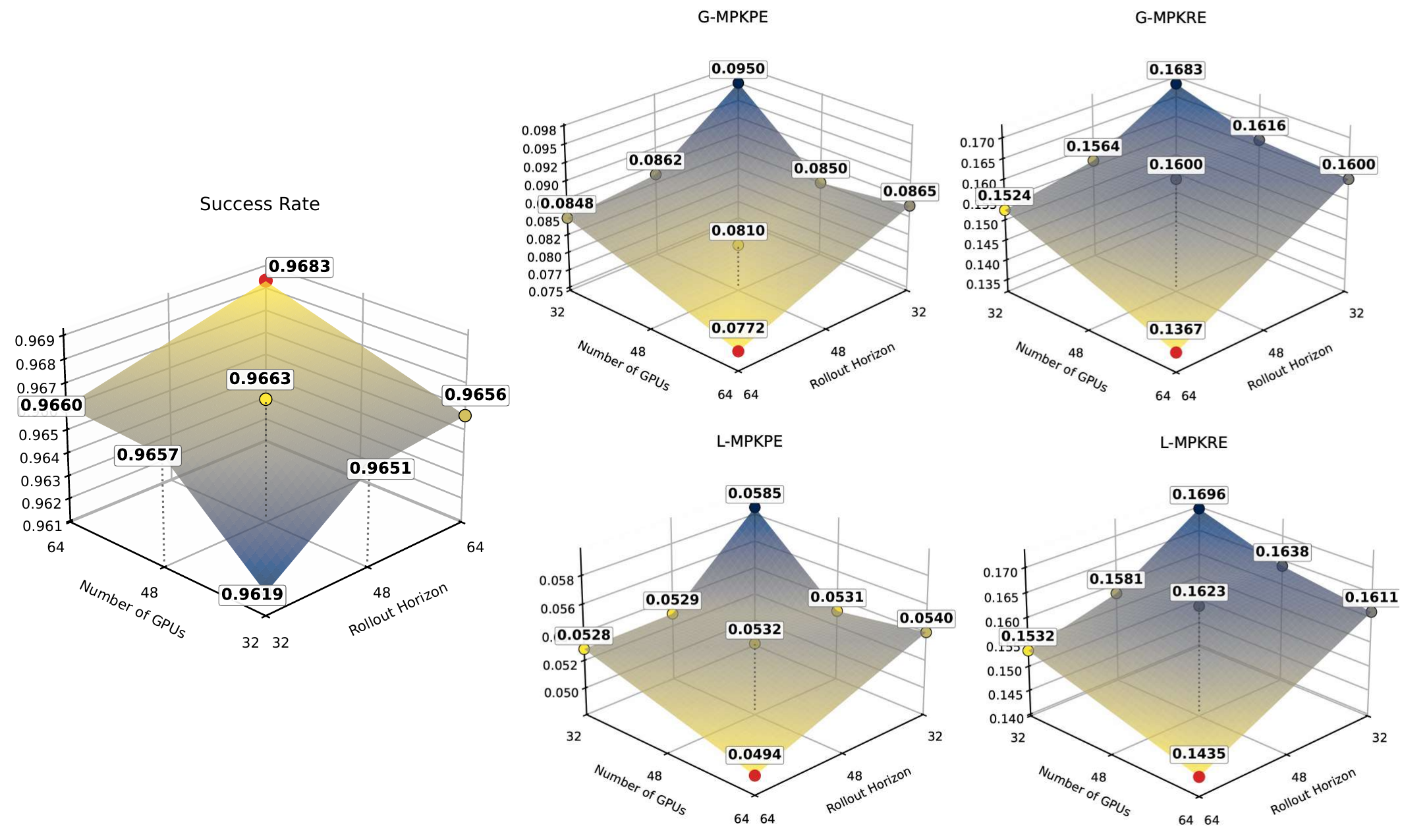}
    \caption{Results of scaling \textbf{on-policy data collection} on the \textbf{BONES Test Set} under \textbf{root-and-end-effector mode}. Best results with the highest success rate and lowest tracking errors are highlighted with red circles.}
    \label{fig:scale_on_policy_bones_five_point}
    \vspace{-4mm}
\end{figure}

\begin{figure}[h!]
    \centering
    \includegraphics[width=0.95\linewidth]{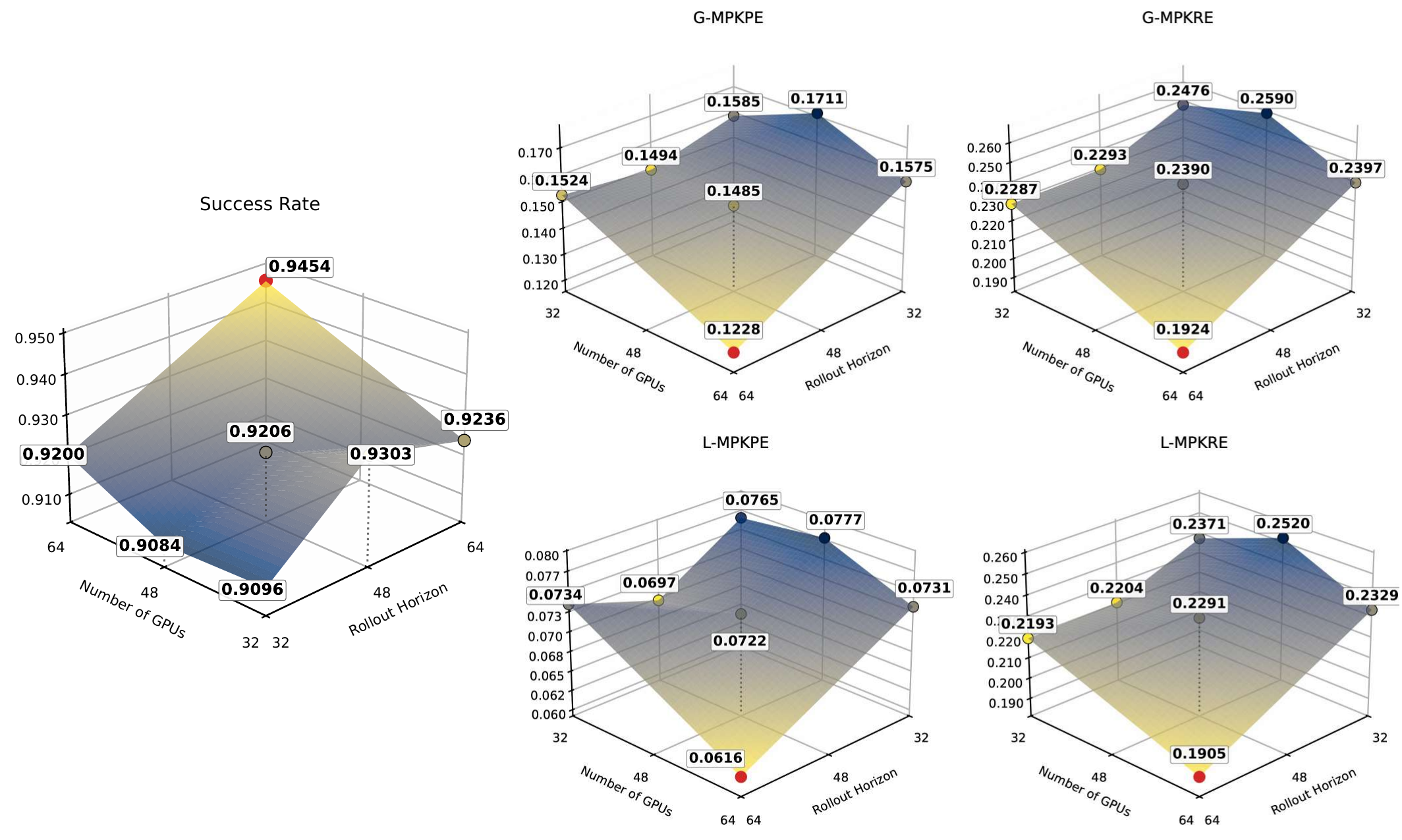}
    \caption{Results of scaling \textbf{on-policy data collection} on the \textbf{Ours Test Set} under \textbf{root-and-end-effector mode}. Best results with the highest success rate and lowest tracking errors are highlighted with red circles.}
    \label{fig:scale_on_policy_ours_five_point}
    \vspace{-4mm}
\end{figure}

\begin{figure}[h!]
    \centering
    \includegraphics[width=0.95\linewidth]{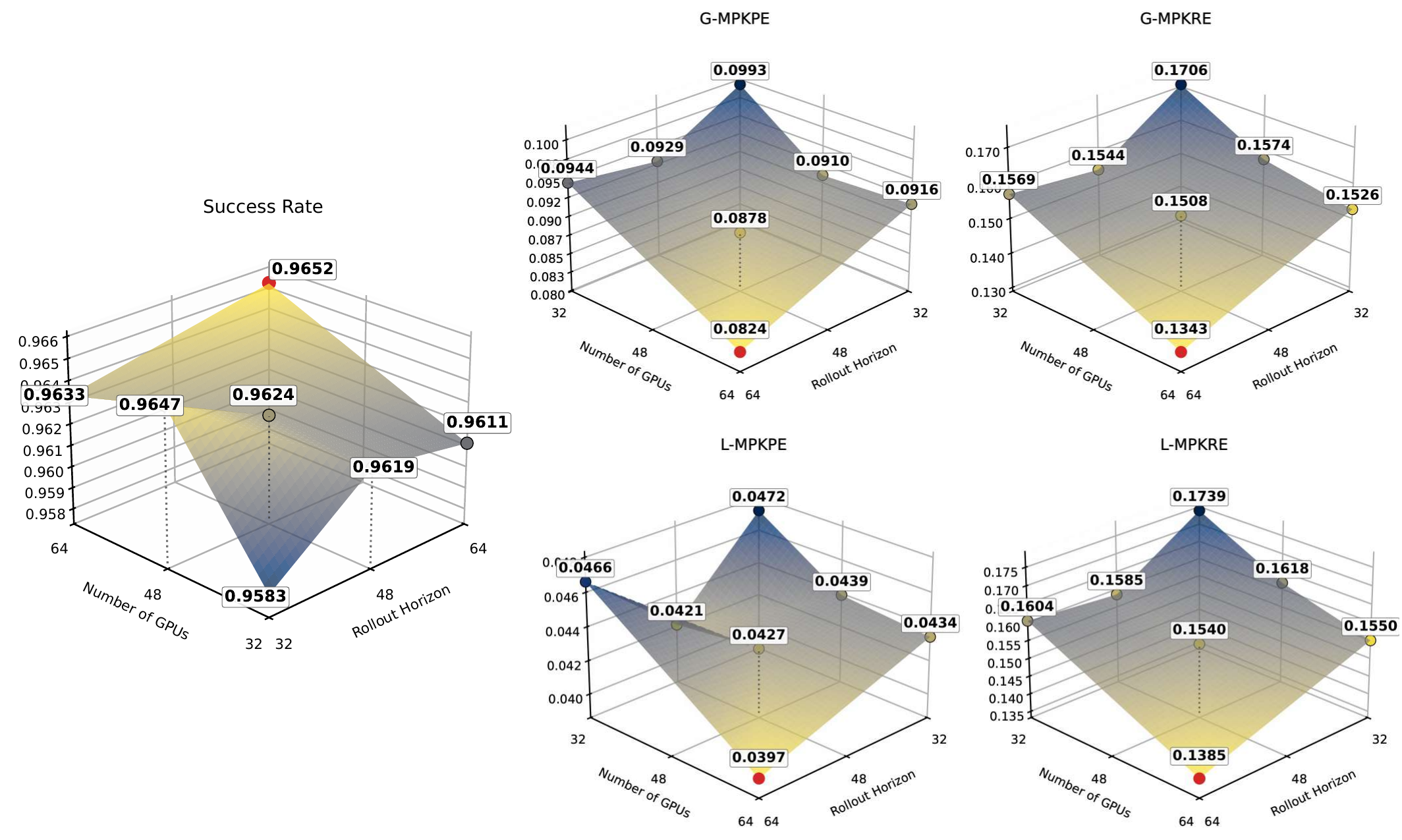}
    \caption{Results of scaling \textbf{on-policy data collection} on the \textbf{BONES Test Set} under \textbf{root-and-upper-body mode}. Best results with the highest success rate and lowest tracking errors are highlighted with red circles.}
    \label{fig:scale_on_policy_bones_seven_point}
    \vspace{-4mm}
\end{figure}

\begin{figure}[h!]
    \centering
    \includegraphics[width=0.95\linewidth]{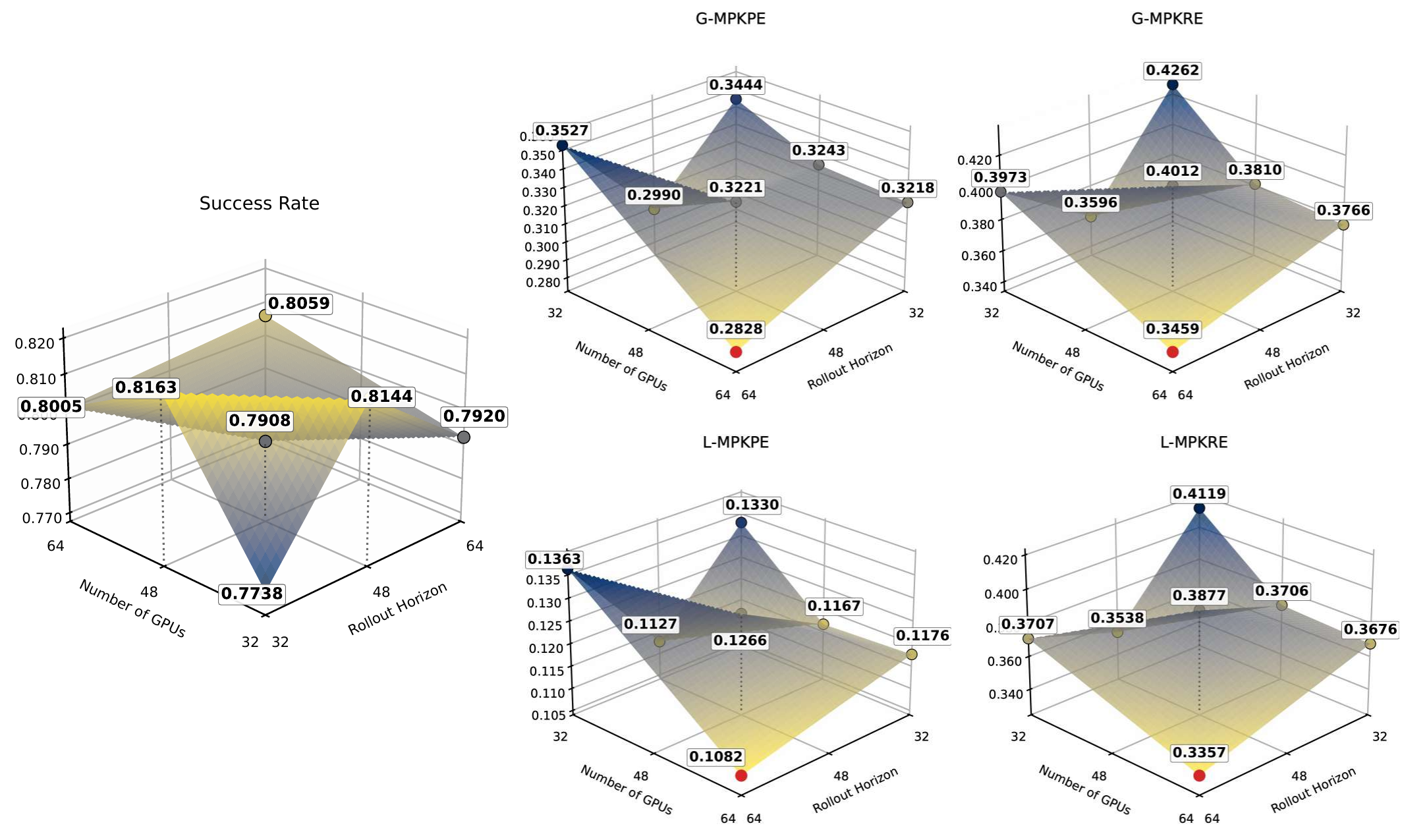}
    \caption{Results of scaling \textbf{on-policy data collection} on the \textbf{Ours Test Set} under \textbf{root-and-upper-body mode}. Best results with the highest success rate and lowest tracking errors are highlighted with red circles.}
    \label{fig:scale_on_policy_ours_seven_point}
    \vspace{-4mm}
\end{figure}

To evaluate the effect of scaling on-policy data collection, we report results of two test sets under four representative control modes: root-and-hand mode (figure~\ref{fig:scale_on_policy_bones_three_point} and~\ref{fig:scale_on_policy_ours_three_point}), root-and-end-effector mode (figure~\ref{fig:scale_on_policy_bones_five_point} and~\ref{fig:scale_on_policy_ours_five_point}), root-and-upper-body mode (figure~\ref{fig:scale_on_policy_bones_seven_point} and~\ref{fig:scale_on_policy_ours_seven_point})), and whole-body mode (figure~\ref{fig:scale_on_policy_bones_whole_body} and~\ref{fig:scale_on_policy_ours_whole_body}). Both the rollout horizon and the number of GPUs are varied across three levels (32, 48, and 64). For all configurations, we evaluate the checkpoint obtained after 20,000 training epochs.

\newpage

\newpage

\newpage

\newpage

\vspace{-5mm}
\subsection{Scaling Reference Motions}
\vspace{-3mm}
\label{app:scaling_reference_motions}

We evaluate the effect of scaling reference motions by partitioning the full dataset into five nested subsets of increasing size, such that each larger subset fully contains all smaller ones. These subsets are denoted as {XXS}, {XS}, {S}, {M}, and {L}. Specifically, {XXS}, {XS}, and {S} correspond to one-third, two-thirds, and the complete BONES-SEED dataset excluding the held-out test split, respectively. The {M} subset comprises all reference motion datasets except Embody3D, whereas {L} contains the complete collection. Details on the statistics of these subsets are presented in table~\ref{tab:ref_stats}. 

\begin{table}[h]
\renewcommand{\arraystretch}{1.15}
    \centering
    \begin{tabular}{ccc}
        \toprule 
        \textbf{Subset} & \textbf{Motion Sequences} & \textbf{Motion Frames} \\
        \midrule 
        XXS & 44,073 & 16,159,416 \\
        XS & 88,146 & 32,230,197 \\
        S & 132,220 & 48,321,234 \\
        M & 165,028 & 71,130,505 \\
        L & 184,206 & 101,979,598 \\
        \bottomrule
    \end{tabular}
    \caption{\textbf{Reference Motion Subset Statistics.} The frames are calculated based on 50 FPS.}
    \label{tab:ref_stats}
\end{table}

\begin{figure}[h!]
    \centering
    \includegraphics[width=0.95\linewidth]{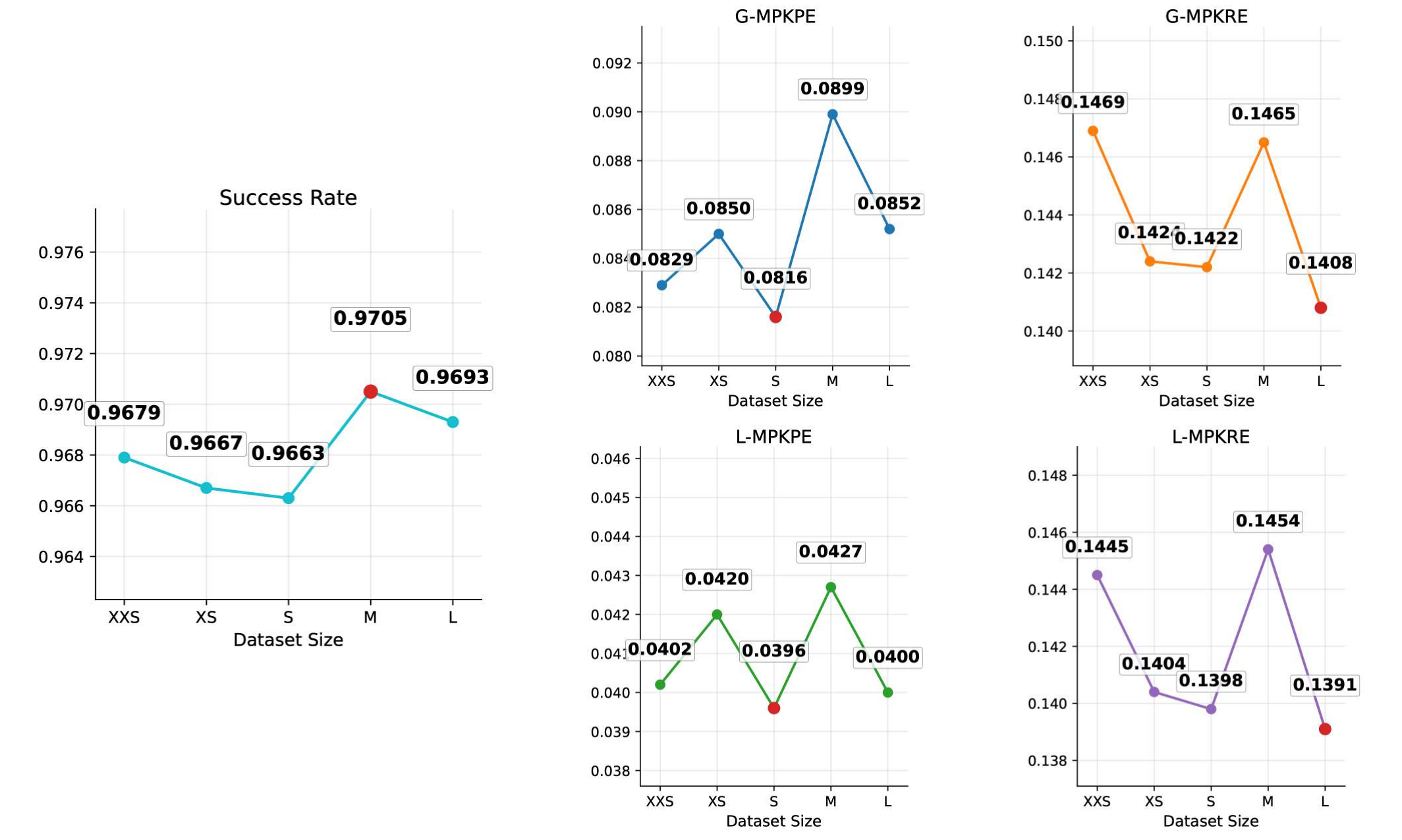}
    \caption{Results of scaling \textbf{reference motions} on \textbf{BONES Test Set} under \textbf{root-and-hand mode}. Best results corresponding to the highest success rate and lowest tracking errors are highlighted with red circles.}
    \label{fig:scale_ref_three_point_bones}
    \vspace{-4mm}
\end{figure}

\begin{figure}[h!]
    \centering
    \includegraphics[width=0.95\linewidth]{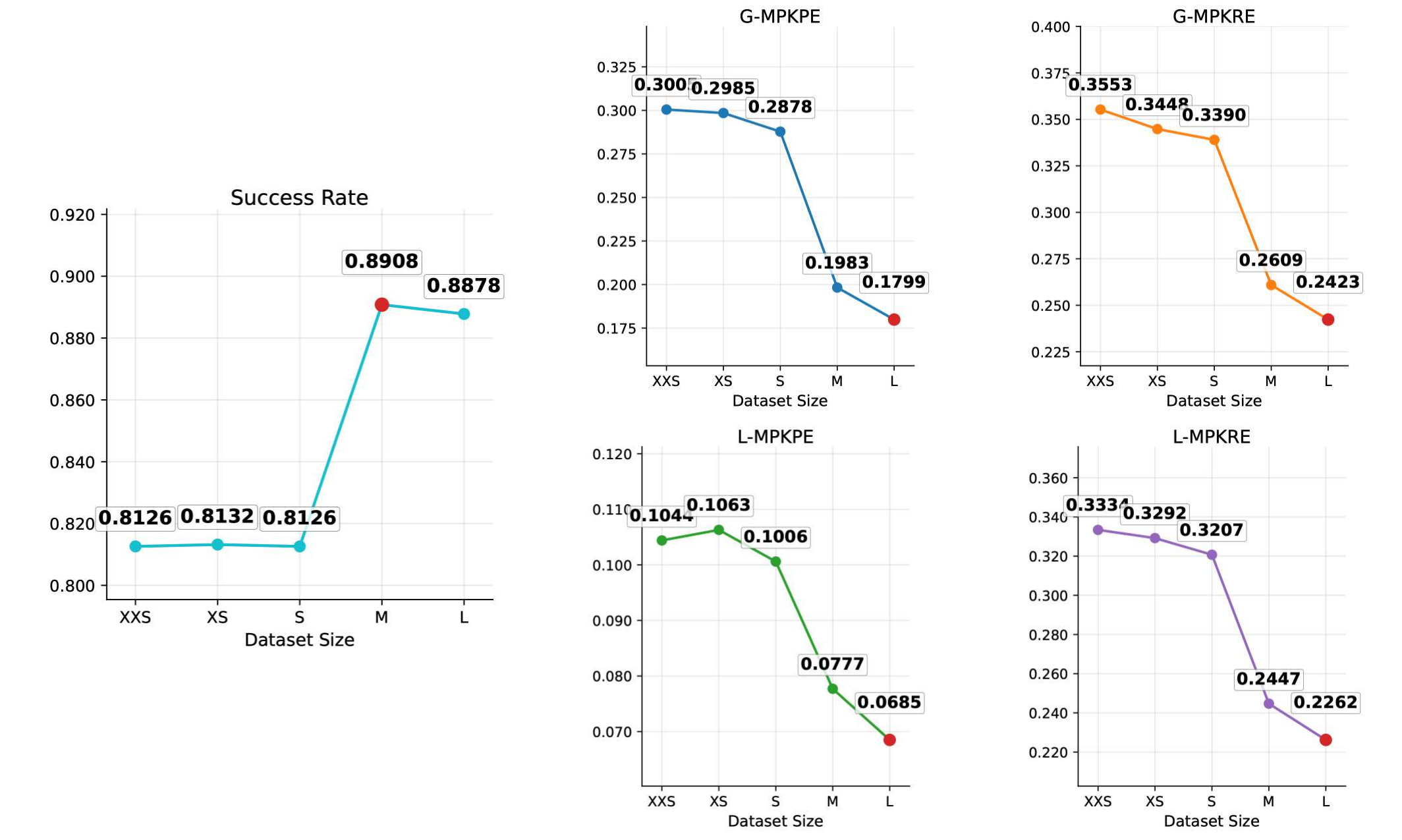}
    \caption{Results of scaling \textbf{reference motions} on \textbf{Ours Test Set} under \textbf{root-and-hand mode}. Best results corresponding to the highest success rate and lowest tracking errors are highlighted with red circles.}
    \label{fig:scale_ref_three_point_ours}
    \vspace{-4mm}
\end{figure}

\begin{figure}[h!]
    \centering
    \includegraphics[width=0.95\linewidth]{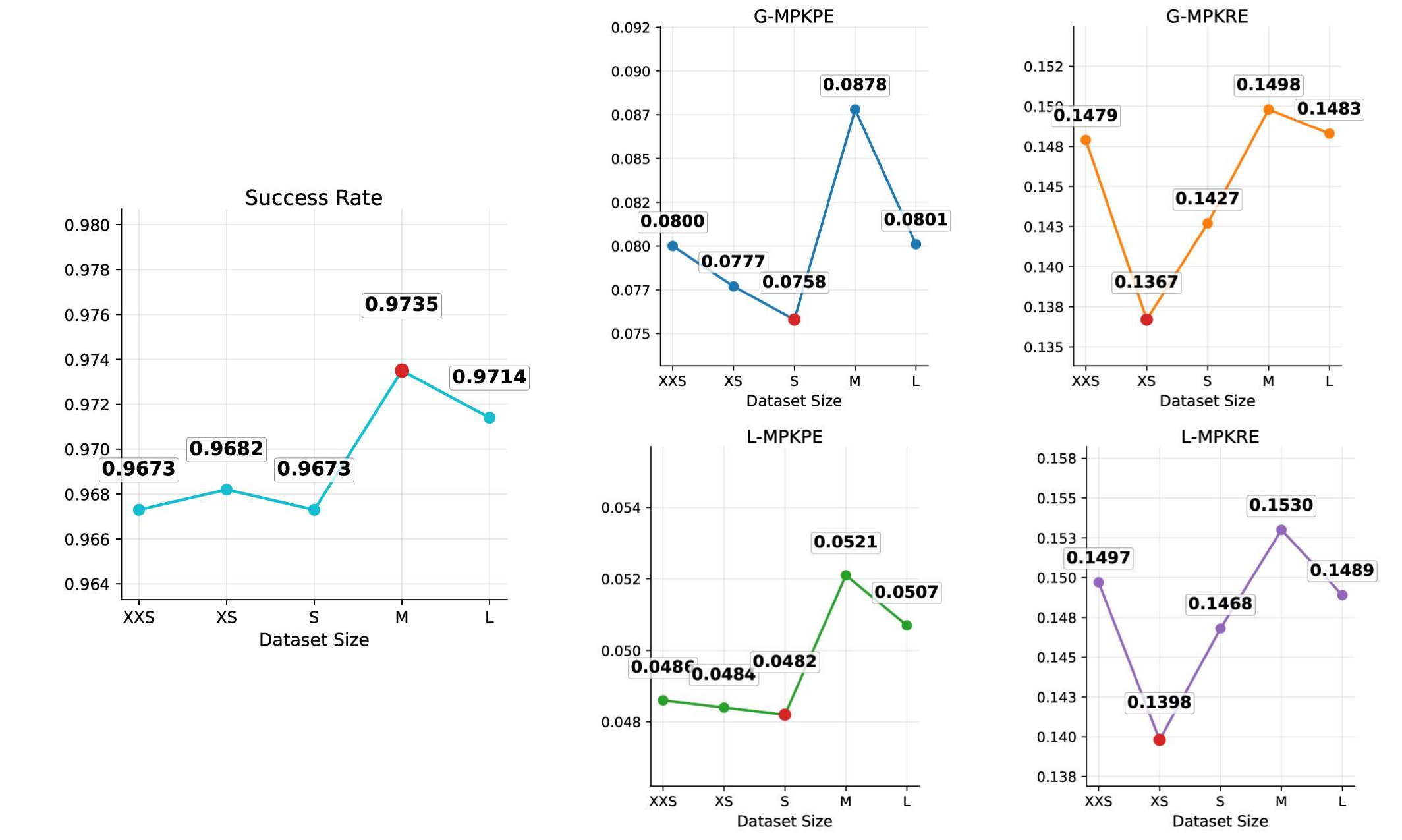}
    \caption{Results of scaling \textbf{reference motions} on \textbf{BONES Test Set} under \textbf{root-and-end-effector mode}. Best results corresponding to the highest success rate and lowest tracking errors are highlighted with red circles.}
    \label{fig:scale_ref_five_point_bones}
    \vspace{-4mm}
\end{figure}

\begin{figure}[h!]
    \centering
    \includegraphics[width=0.95\linewidth]{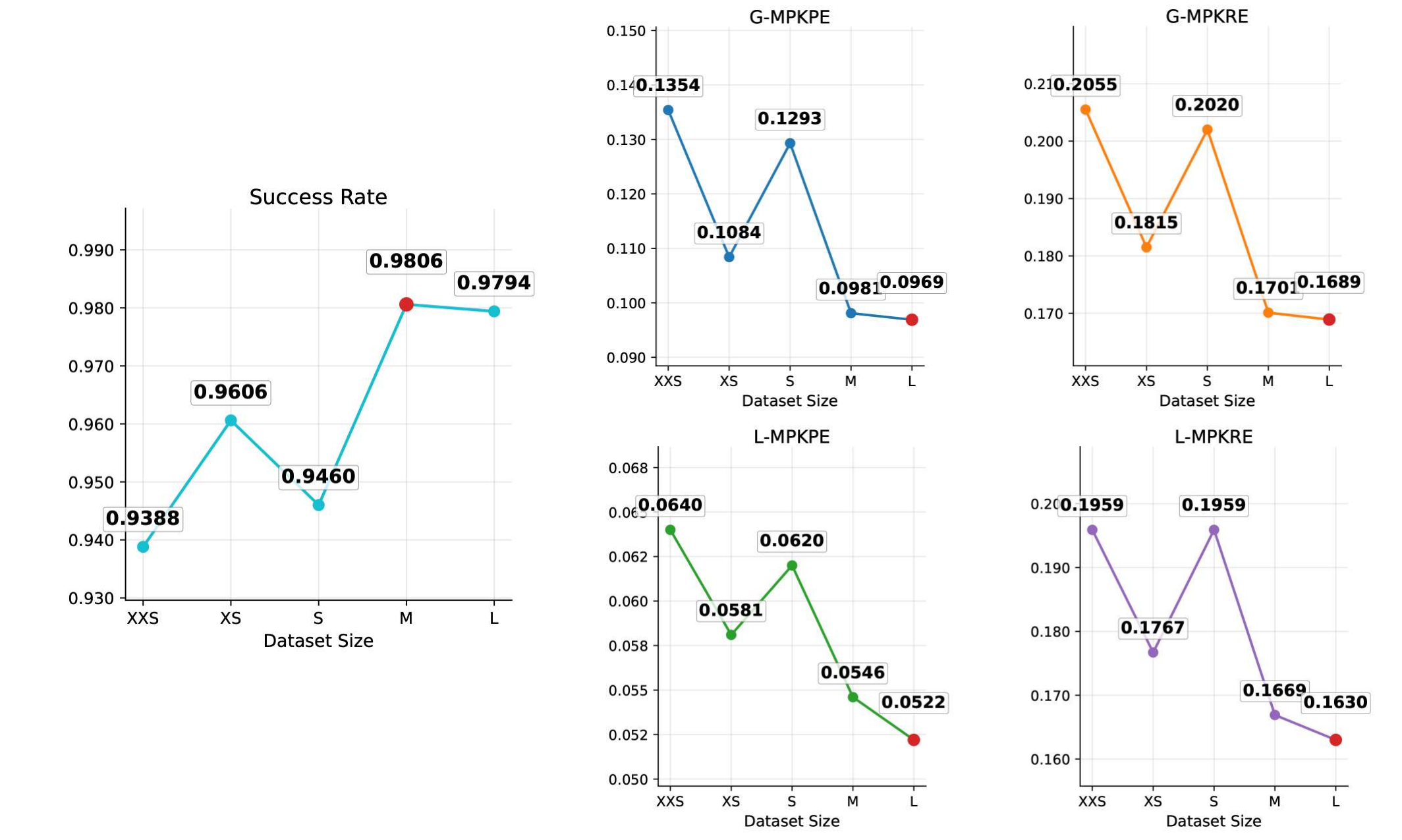}
    \caption{Results of scaling \textbf{reference motions} on \textbf{Ours Test Set} under \textbf{root-and-end-effector mode}. Best results corresponding to the highest success rate and lowest tracking errors are highlighted with red circles.}
    \label{fig:scale_ref_five_point_ours}
    \vspace{-4mm}
\end{figure}

\begin{figure}[h!]
    \centering
    \includegraphics[width=0.95\linewidth]{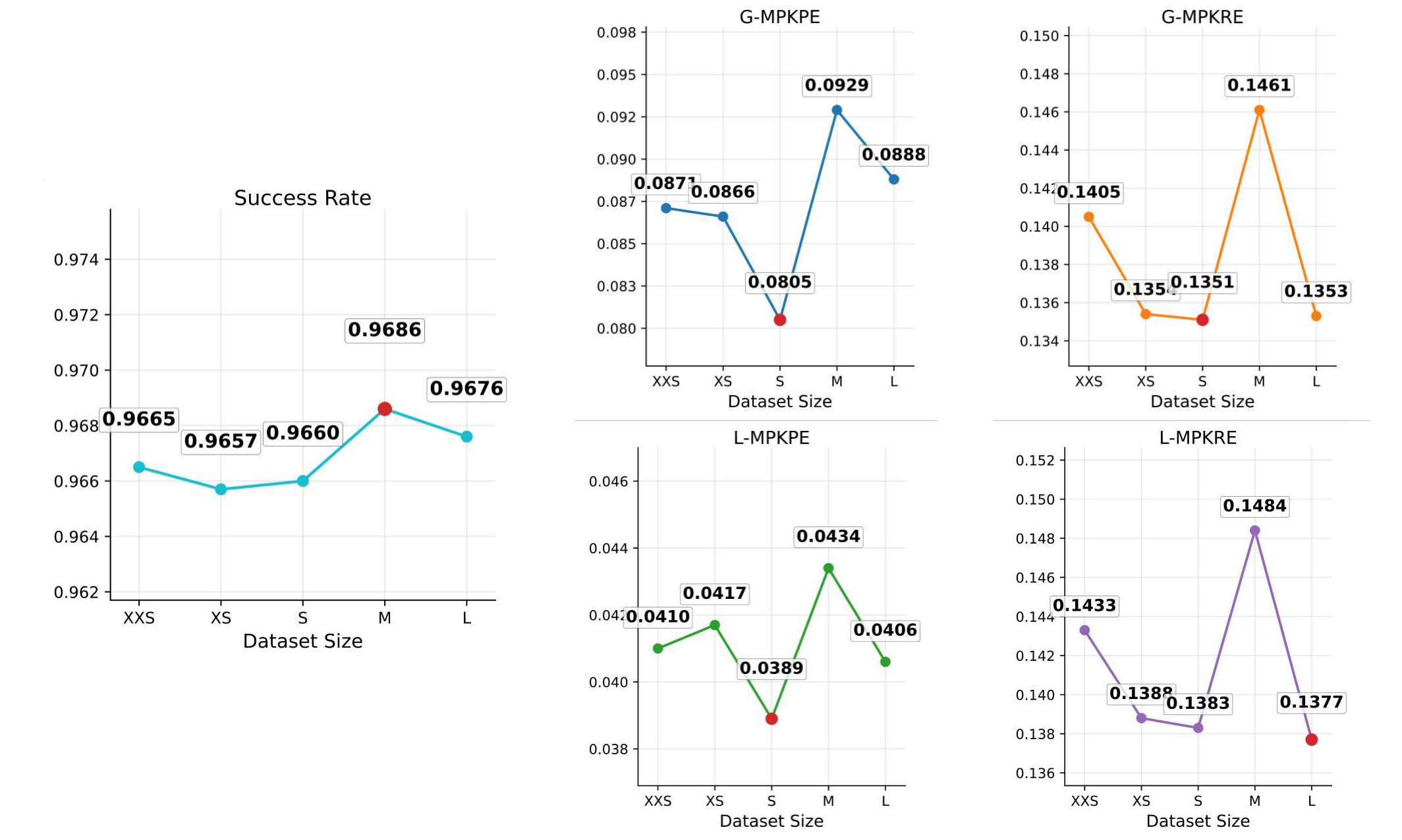}
    \caption{Results of scaling \textbf{reference motions} on \textbf{BONES Test Set} under \textbf{root-and-upper-body mode}. Best results corresponding to the highest success rate and lowest tracking errors are highlighted with red circles.}
    \label{fig:scale_ref_seven_point_bones}
    \vspace{-4mm}
\end{figure}

\begin{figure}[h!]
    \centering
    \includegraphics[width=0.95\linewidth]{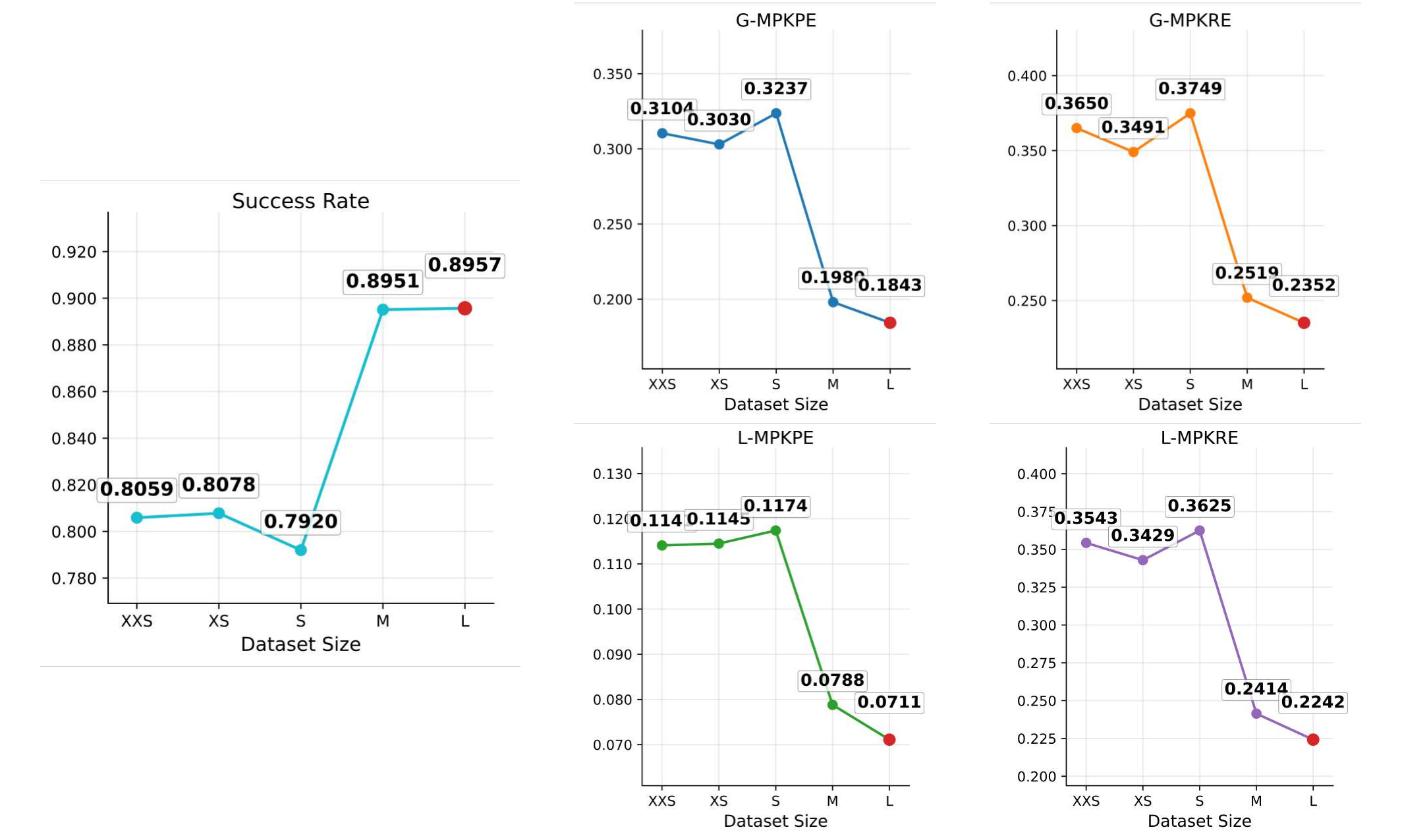}
    \caption{Results of scaling \textbf{reference motions} on \textbf{Ours Test Set} under \textbf{root-and-upper-body mode}. Best results corresponding to the highest success rate and lowest tracking errors are highlighted with red circles.}
    \label{fig:scale_ref_seven_point_ours}
    \vspace{-4mm}
\end{figure}

All training configurations and the model size are kept fixed across experiments, and the optimal on-policy data configuration (64 GPUs and a rollout horizon of 64) is adopted throughout. We still report results of two test sets under four representative control modes: root-and-hand mode (figure~\ref{fig:scale_ref_three_point_bones} and~\ref{fig:scale_ref_three_point_ours}), root-and-end-effector mode (figure~\ref{fig:scale_ref_five_point_bones} and~\ref{fig:scale_ref_five_point_ours}), root-and-upper-body mode (figure~\ref{fig:scale_ref_seven_point_bones} and~\ref{fig:scale_ref_seven_point_ours}), and whole-body mode (figure~\ref{fig:scale_ref_whole_body_bones} and~\ref{fig:scale_ref_whole_body_ours}). We evaluate the checkpoint obtained after 22,200 training epochs.

To provide context for interpreting the observed scaling behavior, we first quantitatively analyze the behavioral characteristics of the training partitions. Specifically, we define a behavioral feature as:
$$
(h,r,p,v_x,v_y,\omega_z,j_p,j_v)
$$
where $h$ denotes the root height, $r$ and $p$ are the root roll and pitch angles, $v_x$ and $v_y$ are the root linear velocities on the XY-plane, $\omega_z$ is the root angular velocity along the z-axis, $j_p$ is the joint positions and $j_v$ is the joint velocities. The root velocities are all expressed in the humanoid heading frame. Together, these components form an expressive feature vector of 64 dimensions given Unitree G1 ($1+2+2+1+29+29$). 

We extract such behavioral features from all motion frames, and fit a K-Means clustering model using 200,000 randomly sampled features from the complete L partition, yielding a shared clustering space consisting of 2,000 clusters. Subsequently, an equal number of behavioral features are randomly sampled from each partition and assigned to the nearest cluster in the shared clustering space. A cluster is considered occupied if at least one sampled feature is assigned to it. The \textbf{occupancy rate} is defined as the ratio of occupied clusters to the total number of clusters. A higher occupancy rate indicates that the partition spans a broader range of behavioral regions in the shared feature space and therefore exhibits greater behavioral coverage. 
It is worth noting that this quantitative measure is inherently dependent on the choice of reference motions used to construct the shared clustering space. In this paper, we use the complete partition and implicit assume that it provides sufficiently broad behavioral support for meaningful comparisons across partitions. Nevertheless, no finite dataset can exhaust the space of possible behaviors and rare or long-tail behavioral patterns may still be absent. Consequently, determining an appropriate support for behavioral coverage analysis remains an important direction for future work.

\newpage

\newpage

\newpage

\newpage

\vspace{-5mm}
\subsection{Scaling with Model Architectures}
\vspace{-3mm}
\label{app:scaling_models}

We next explore the effect of model architectures on the scaling behavior of BFMs. Specifically, we examine whether the proposed Humanoid Transformer outperforms the MLP architecture and whether further increasing its capacity leads to additional performance gains. Throughout this paper, the actor and critic are instantiated with identical backbone architectures and model capacities. They differ only in their input and output dimensions, reflecting the asymmetric observations and prediction objectives. 

\begin{figure}[h!]
    \centering
    \includegraphics[width=0.95\linewidth]{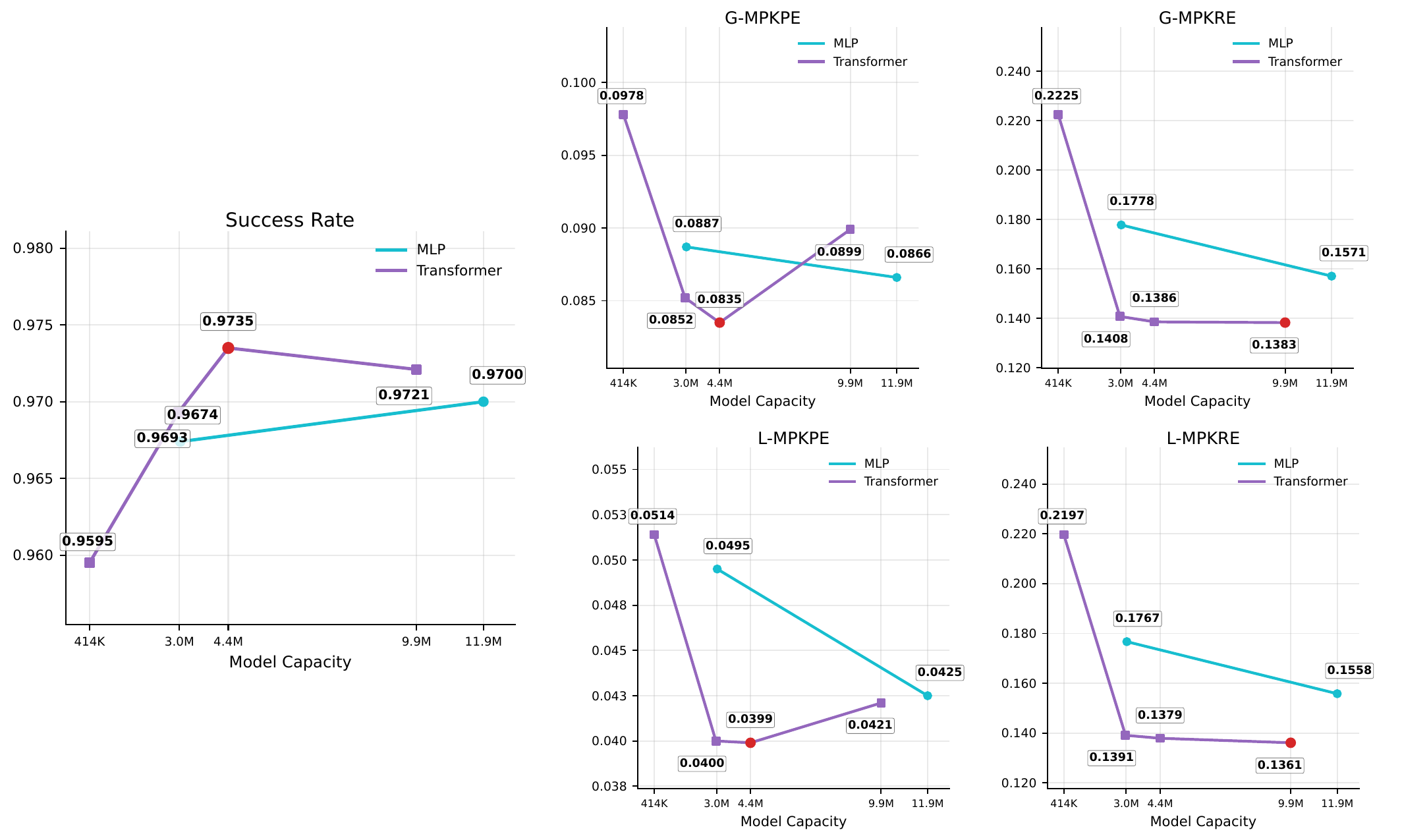}
    \caption{Results of scaling \textbf{model architectures} on \textbf{BONES Test Set} under \textbf{root-and-hand mode}. Best results corresponding to the highest success rate and lowest tracking errors are highlighted with red circles.}
    \label{fig:scale_model_three_point_bones}
    \vspace{-4mm}
\end{figure}

\begin{figure}[h!]
    \centering
    \includegraphics[width=0.95\linewidth]{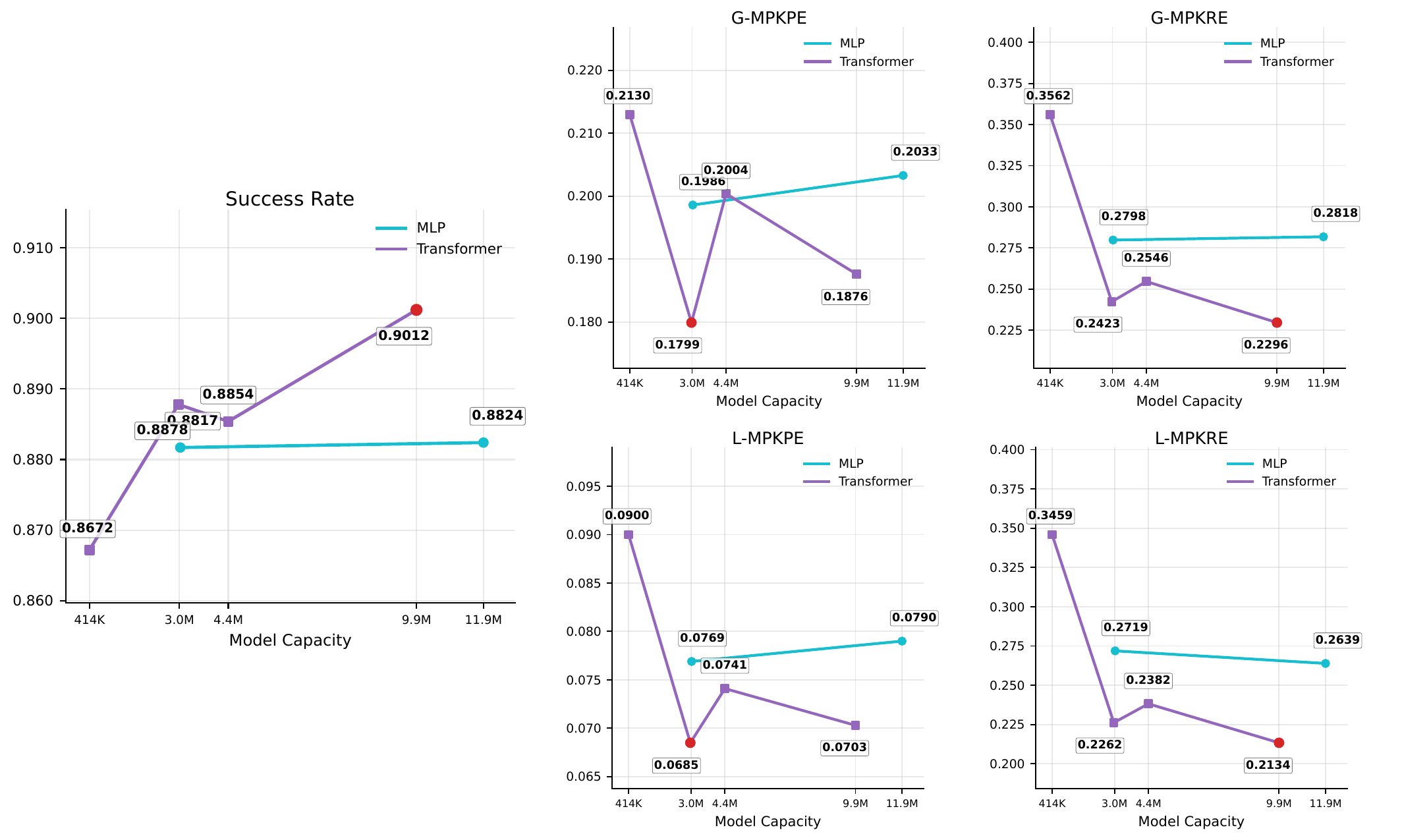}
    \caption{Results of scaling \textbf{model architectures} on \textbf{Ours Test Set} under \textbf{root-and-hand mode}. Best results corresponding to the highest success rate and lowest tracking errors are highlighted with red circles.}
    \label{fig:scale_model_three_point_ours}
    \vspace{-4mm}
\end{figure}

\begin{figure}[h!]
    \centering
    \includegraphics[width=0.95\linewidth]{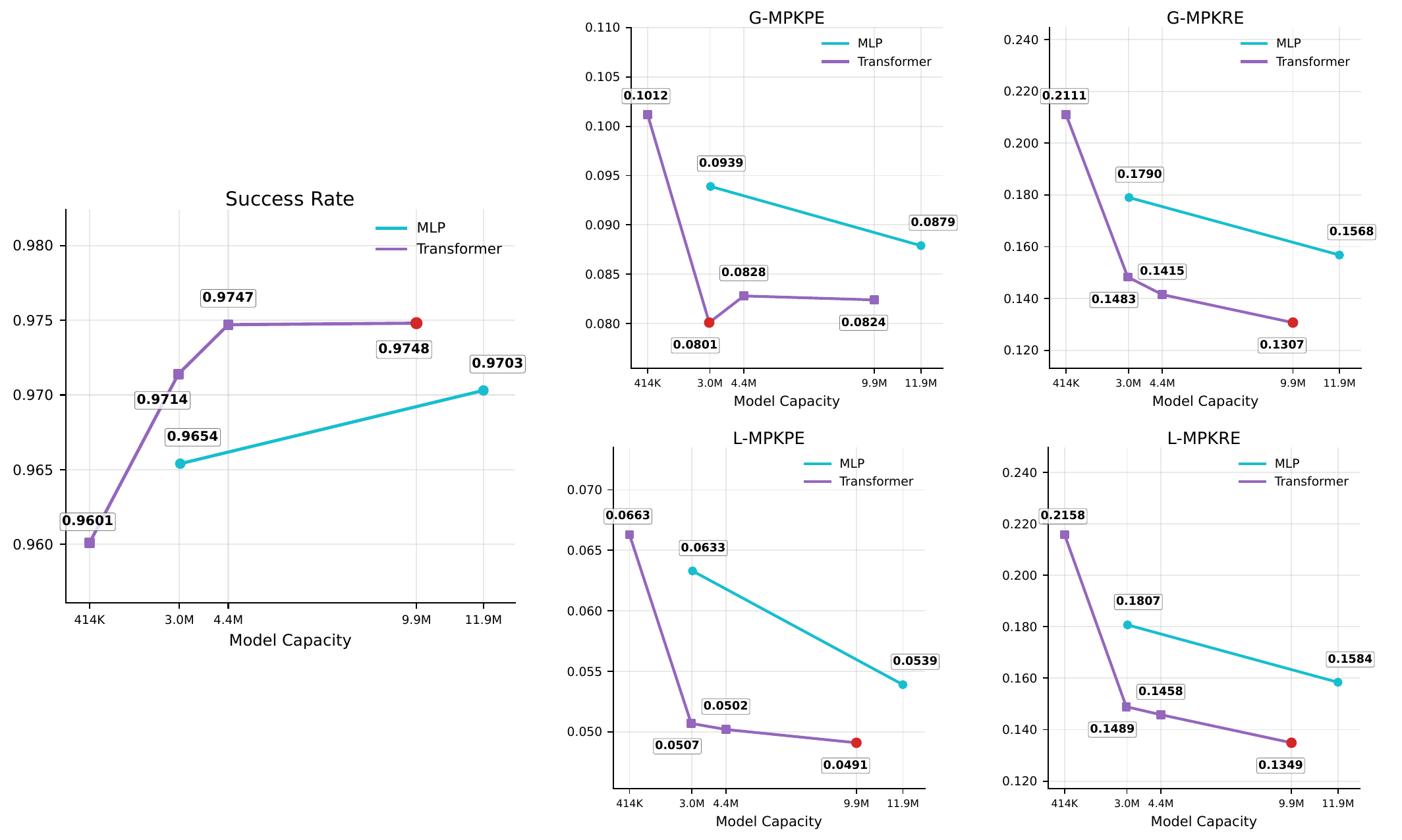}
    \caption{Results of scaling \textbf{model architectures} on \textbf{BONES Test Set} under \textbf{root-and-end-effector mode}. Best results corresponding to the highest success rate and lowest tracking errors are highlighted with red circles.}
    \label{fig:scale_model_five_point_bones}
    \vspace{-4mm}
\end{figure}

\begin{figure}[h!]
    \centering
    \includegraphics[width=0.95\linewidth]{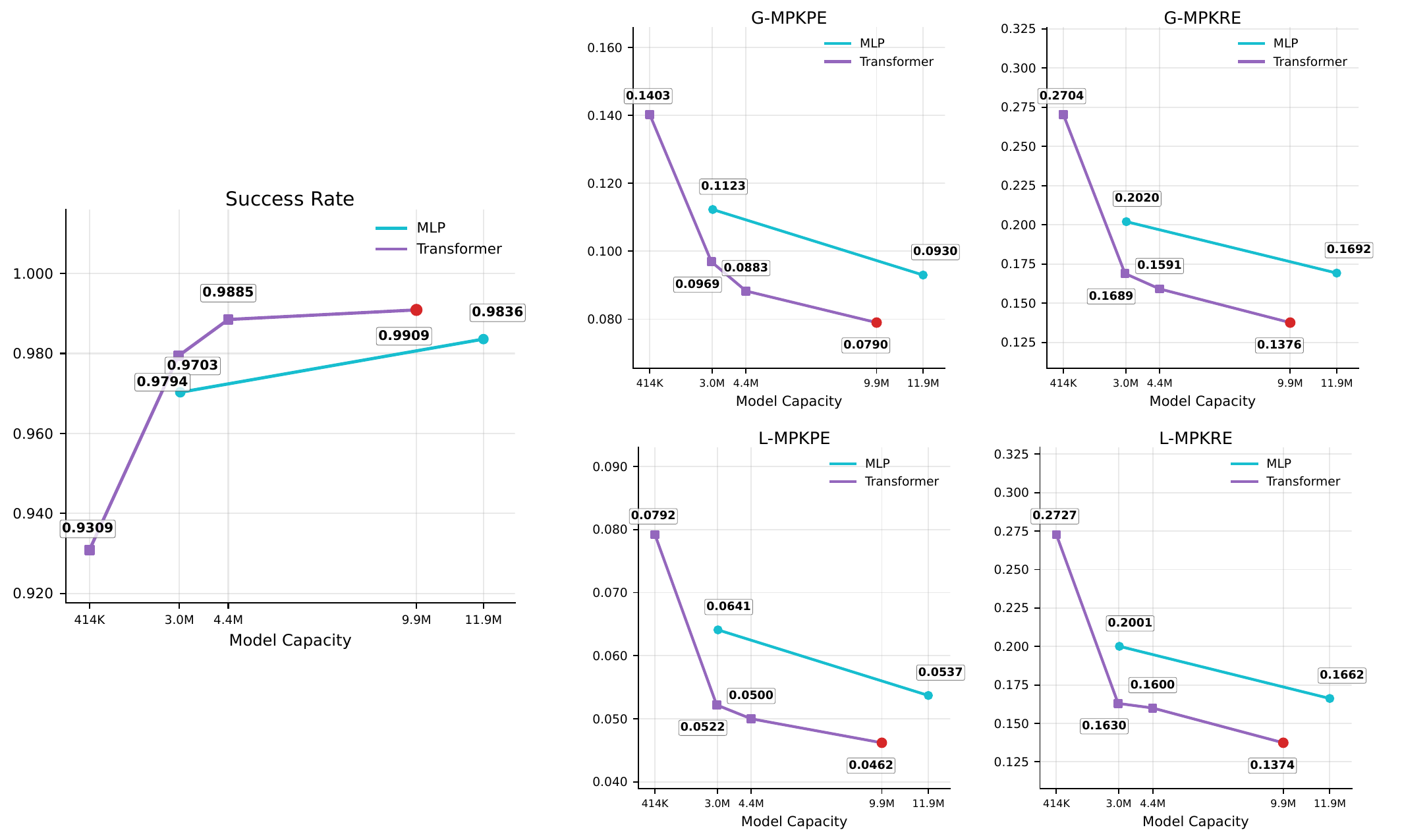}
    \caption{Results of scaling \textbf{model architectures} on \textbf{Ours Test Set} under \textbf{root-and-end-effector mode}. Best results corresponding to the highest success rate and lowest tracking errors are highlighted with red circles.}
    \label{fig:scale_model_five_point_ours}
    \vspace{-4mm}
\end{figure}

\begin{figure}[h!]
    \centering
    \includegraphics[width=0.95\linewidth]{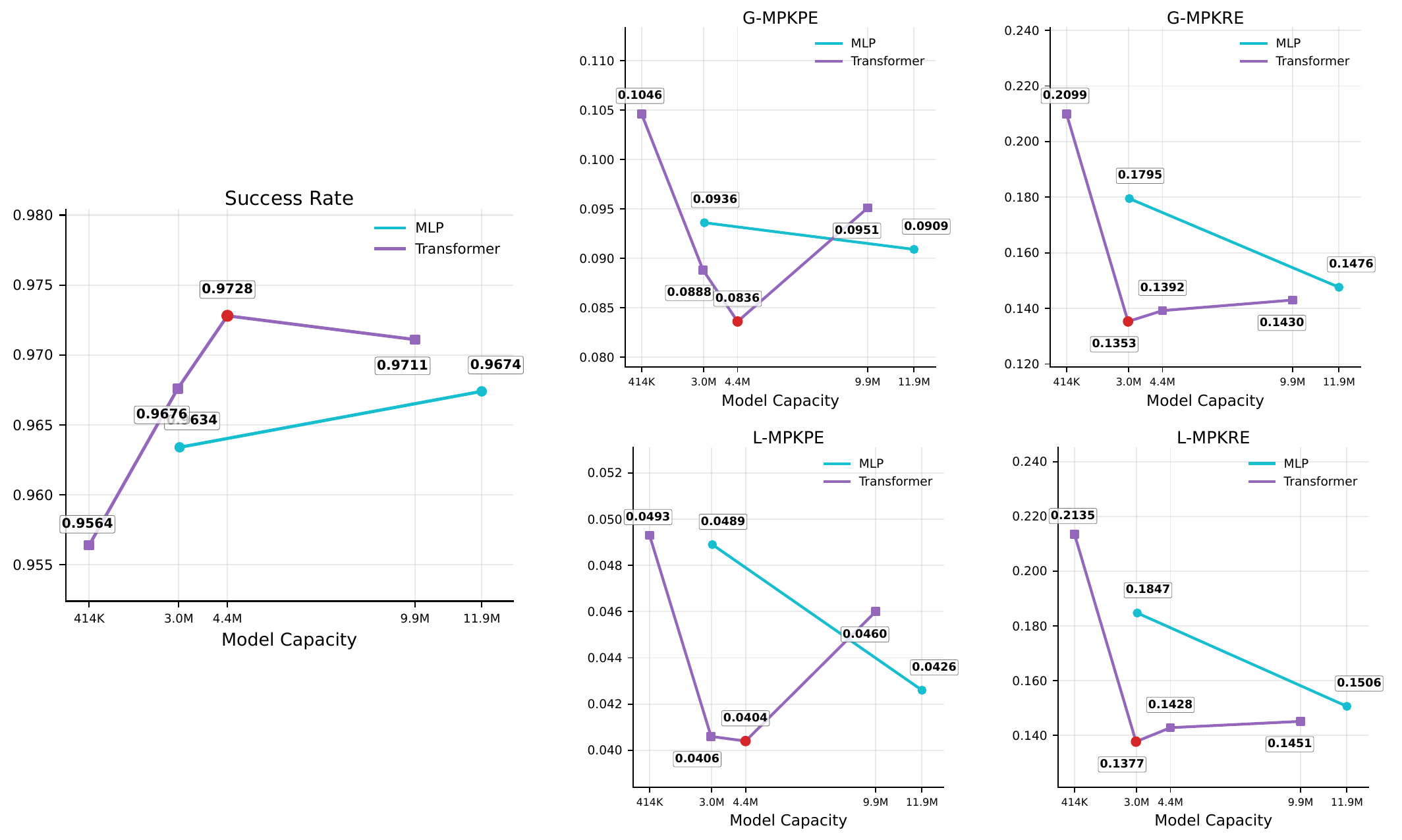}
    \caption{Results of scaling \textbf{model architectures} on \textbf{BONES Test Set} under \textbf{root-and-upper-body mode}. Best results corresponding to the highest success rate and lowest tracking errors are highlighted with red circles.}
    \label{fig:scale_model_seven_point_bones}
    \vspace{-4mm}
\end{figure}

\begin{figure}[h!]
    \centering
    \includegraphics[width=0.95\linewidth]{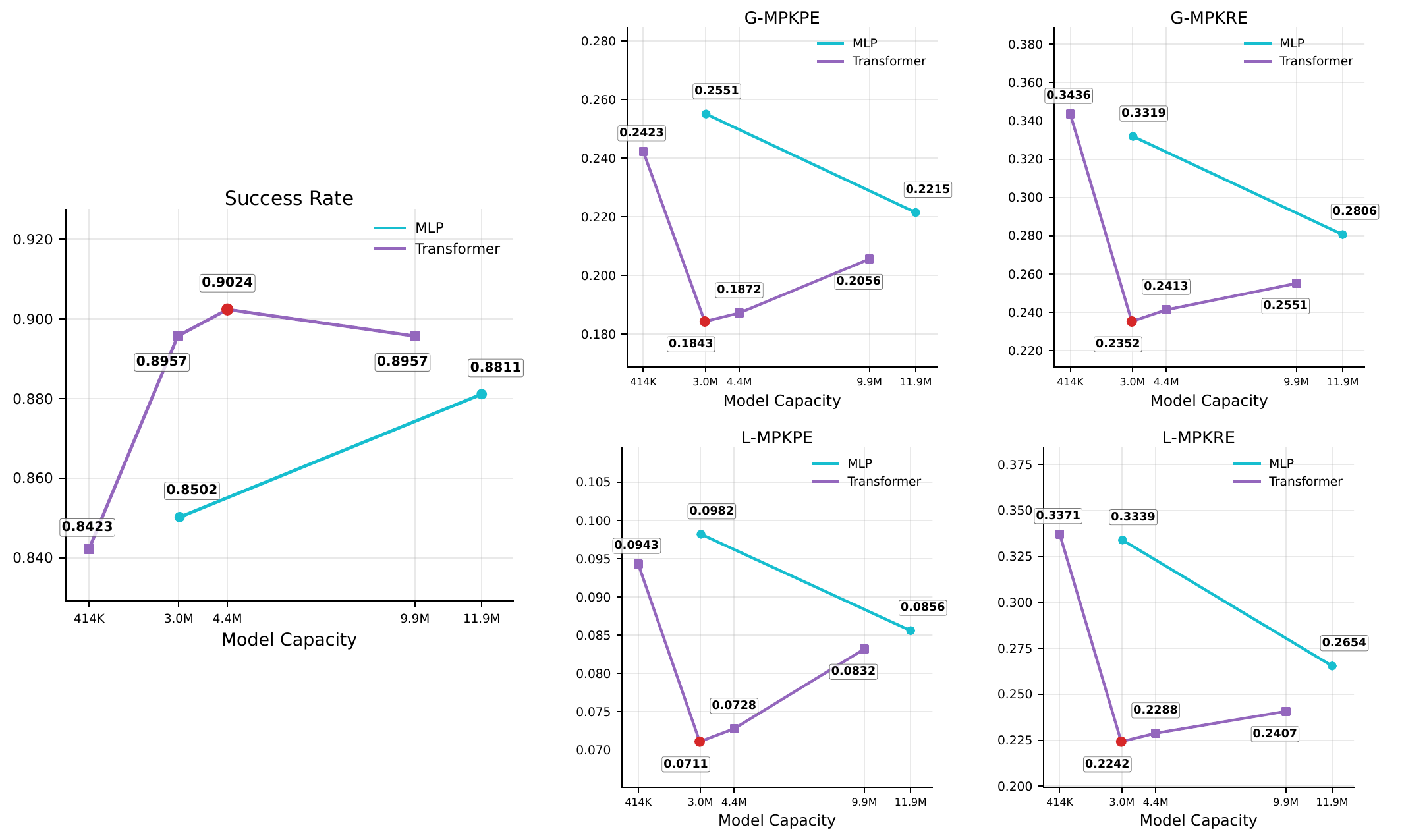}
    \caption{Results of scaling \textbf{model architectures} on \textbf{Ours Test Set} under \textbf{root-and-upper-body mode}. Best results corresponding to the highest success rate and lowest tracking errors are highlighted with red circles.}
    \label{fig:scale_model_seven_point_ours}
    \vspace{-4mm}
\end{figure}

All algorithmic and environmental configurations and kept fixed across experiments, including the PPO hyperparameters listed in table~\ref{tab:hyperparams}, the on-policy data collection configuration of 64 GPUs with a rollout horizon of 64, and the L partition of the reference motion dataset. We still report results of two test sets under four representative control modes: root-and-hand mode (figure~\ref{fig:scale_model_three_point_bones} and~\ref{fig:scale_model_three_point_ours}), root-and-end-effector mode (figure~\ref{fig:scale_model_five_point_bones} and~\ref{fig:scale_model_five_point_ours}), root-and-upper-body mode (figure~\ref{fig:scale_model_seven_point_bones} and~\ref{fig:scale_model_seven_point_ours}), and whole-body mode (figure~\ref{fig:scale_model_bones_whole_body} and~\ref{fig:scale_model_ours_whole_body}). We evaluate the checkpoint obtained after 22,200 training epochs for each experiment.

For the MLP backbone, we adopt two representative configurations following prior work. The small configuration contains 3.05M parameters, with hidden dimensions of [1024, 768, 512], while the large configuration contains 11.86M parameters, with hidden dimensions of [2048, 2048, 1024, 1024, 512, 512]. Both MLP variants employ ELU activations. No normalization is applied to the observations.

For the proposed Humanoid Transformer, we evaluate four model configurations by varying four critical components: the embedding dimension, feed-forward dimension, number of attention heads, and network depth. The S model contains 0.41M parameters, with an embedding dimension of 128, 2 attention heads, a feed-forward dimension of 128, and 2 Transformer layers. The M model contains 3.00M parameters, with an embedding dimension of 256, 4 attention heads, a feed-forward dimension of 256, and 4 layers. The L model contains 4.44M parameters and shares the same width as the M model but increases the depth to 6 layers, thereby isolating the effect of scaling network depth. Finally, the XL model contains 9.91M parameters, with an embedding dimension of 384, 6 attention heads, a feed-forward dimension of 384, and 6 layers, representing joint scaling of both network width and depth. 

It is worth noting that our scaling study remains relatively limited compared with those conducted for Large Language Models. This limitation stems from two primary factors. First, humanoid robot control still lacks the mature infrastructure required to support truly large-scale pretraining. Distributed training remains fragile due to multi-GPU communication, simulator instability, and the limited exploration of efficient training systems tailored to humanoid foundation models, not to mention the community has yet to reach a consensus on the appropriate foundation model paradigm for humanoid robots. Second, practical deployment imposes strict computational constraints. To sustain an inference frequency above 50 Hz while reserving computational budget for future integration with high-level models, the parameter count of Behavior Foundation Models cannot be increased indefinitely under current hardware constraints. Nevertheless, as onboard computing capabilities continue to improve, establishing a scalable training pipeline is of great value for future humanoid foundation models. We are actively working to address these limitations with the constraints of current hardware and software systems. Nevertheless, we believe that substantial progress will ultimately require coordinated efforts from the broader community. To facilitate such progress, we will open-source all the resources, with the hope of providing a scalable foundation upon which the community can collectively build.

\newpage

\newpage

\newpage

\end{document}